\documentclass{article}

\usepackage[a4paper, margin=1.5cm]{geometry}

\usepackage{verbatim}
\usepackage{graphicx}
\usepackage{natbib}
\usepackage{caption}
\usepackage{algorithm}
\usepackage{algorithmic}
\usepackage{newfloat}
\usepackage{listings}
\usepackage{amsmath}
\usepackage{amssymb}
\usepackage{amsfonts}
\usepackage{amsopn}
\usepackage{epsfig}
\usepackage{microtype}
\usepackage{subfigure}
\usepackage{booktabs}
\usepackage{amsfonts}
\usepackage{nicefrac}
\usepackage{microtype}
\usepackage{xcolor}
\usepackage{stmaryrd}
\usepackage{lipsum}
\usepackage{rotating}
\usepackage{wasysym}
\usepackage{phonetic}
\usepackage{halloweenmath}
\usepackage{bm}
\usepackage{amsmath,stackengine,scalerel}
\usepackage{bbm}
\usepackage{censor}
\usepackage[makeroom]{cancel}
\usepackage{mathtools}
\usepackage{amsthm}
\usepackage{url}
\def\UrlBreaks{\do\/\do-}

\newcommand{\tsp}[0]{{\rm T}}

\newcommand{\infset}[1]{{\mathbb{#1}}}
\newcommand{\finset}[1]{{\tt #1}}
\newcommand{\distrib}[1]{{\mathcal{#1}}}
\newcommand{\rv}[1]{{{\rm #1}}}
\newcommand{\boolfont}[1]{{\tt #1}}
\newcommand{\ordset}[1]{{\tt #1}}

\newcommand{\rkhs}[1]{\mathcal{H}_{#1}}
\newcommand{\subrkhs}[1]{\mathcal{H}_{#1}}
\newcommand{\subsubrkhs}[1]{\mathcal{H}_{#1}}
\newcommand{\frkhs}[1]{\mathcal{F}_{#1}}
\newcommand{\subfrkhs}[1]{\mathcal{F}_{#1}}
\newcommand{\subsubfrkhs}[1]{\mathcal{F}_{#1}}

\newcommand{\rkbs}[1]{\mathcal{B}_{#1}}
\newcommand{\subrkbs}[1]{\mathcal{B}_{#1}}
\newcommand{\subsubrkbs}[1]{\mathcal{B}_{#1}}
\newcommand{\frkbs}[1]{\mathcal{F}_{#1}}
\newcommand{\subfrkbs}[1]{\mathcal{F}_{#1}}
\newcommand{\subsubfrkbs}[1]{\mathcal{F}_{#1}}

\newcommand{\rkks}[1]{\mathcal{K}_{#1}}
\newcommand{\subrkks}[1]{\mathcal{K}_{#1}}
\newcommand{\subsubrkks}[1]{\mathcal{K}_{#1}}
\newcommand{\frkks}[1]{\mathcal{F}_{#1}}
\newcommand{\subfrkks}[1]{\mathcal{F}_{#1}}
\newcommand{\subsubfrkks}[1]{\mathcal{F}_{#1}}

\newcommand{\rkkbs}[1]{\mathfrak{K}_{#1}}
\newcommand{\subrkkbs}[1]{\mathfrak{K}_{#1}}
\newcommand{\subsubrkkbs}[1]{\mathfrak{K}_{#1}}
\newcommand{\frkkbs}[1]{\mathcal{F}_{#1}}
\newcommand{\subfrkkbs}[1]{\mathcal{F}_{#1}}
\newcommand{\subsubfrkkbs}[1]{\mathcal{F}_{#1}}

\newcommand{\hypothesisspace}[0]{\textara{f}}
\newcommand{\unitnormball}[1]{\textara{f}_{\!#1}}
\newcommand{\unitnormballkrein}[1]{\textara{f}_{\!#1}}

\newcommand{\normdist}[0]{\distrib{N}}
\newcommand{\expect}[0]{\mathbb{E}}
\newcommand{\prob}[0]{\mathbb{P}}
\newcommand{\deviate}[0]{\mathbb{D}}
\newcommand{\empire}[1]{\hat{\!#1}}

\newcommand{\gp}[0]{\mathcal{GP}}

\newcommand{\krein}[0]{Kre{\u{\i}}n}
\newcommand{\frechet}[0]{Fr{\'e}chet}
\newcommand{\gateaux}[0]{G{\^a}teaux}

\newcommand{\bg}[1]{{\bm{#1}}}

\newcommand{\origin}{{\mathcal{O}}}
\newcommand{\nextval}{{\mathcal{N}}}
\newcommand{\changein}{{\Delta}}
\newcommand{\learnrate}{{\eta}}
\newcommand{\bangrad}{{\nu}}
\newcommand{\shadx}{{\mu}}
\newcommand{\shadw}{{\omega}}
\newcommand{\psidualscale}{\widetilde{\epsilon}}
\newcommand{\phidualscale}{\widetilde{\delta}}
\newcommand{\shadwo}{{\widetilde{\omega}}}
\newcommand{\phidual}{\epsilon_\phi}
\newcommand{\phiphidual}{\epsilon_\phi}
\newcommand{\psidual}{\epsilon_\psi}
\newcommand{\psipsidual}{\epsilon_\psi}
\newcommand{\regstep}{{\bullet}}
\newcommand{\backstep}{{\boxast}}
\newcommand{\wtune}{{\delta}}
\newcommand{\degrowth}{{\varpi}}

\makeatletter
\def\widebreve{\mathpalette\wide@breve}
\def\wide@breve#1#2{\sbox\z@{$#1#2$}%
     \mathop{\vbox{\m@th\ialign{##\crcr
\kern0.08em\brevefill#1{0.8\wd\z@}\crcr\noalign{\nointerlineskip}%
                    $\hss#1#2\hss$\crcr}}}\limits}
\def\brevefill#1#2{$\m@th\sbox\tw@{$#1($}%
  \hss\resizebox{#2}{\wd\tw@}{\rotatebox[origin=c]{90}{\upshape(}}\hss$}
\makeatletter

\DeclareFontFamily{U}{mathx}{}
\DeclareFontShape{U}{mathx}{m}{n}{<-> mathx10}{}
\DeclareSymbolFont{mathx}{U}{mathx}{m}{n}
\DeclareMathAccent{\widehat}{0}{mathx}{"70}
\DeclareMathAccent{\widecheck}{0}{mathx}{"71}

\DeclareMathOperator\sgn{sgn} 
\DeclareMathOperator\saddle{saddle} 
\DeclareMathOperator\erf{erf} 
\DeclareMathOperator\argmax{argmax} 
\DeclareMathOperator\argmin{argmin}

\theoremstyle{definition}
\newtheorem{def_rkbs}{Definition}
\newtheorem{def_canonical_scaling}[def_rkbs]{Definition}
\newtheorem{def_canonical_scaling_main}[def_rkbs]{Definition}

\theoremstyle{remark}
\newtheorem{rem_convergence}{Remark}

\newtheorem{rem_nontrivalid}[rem_convergence]{Remark}

\theoremstyle{plain}
\newtheorem{th_convergephi_main}{Lemma}
\newtheorem{th_convergepsi_main}[th_convergephi_main]{Lemma}

\newtheorem{lem_densekey}{Lemma}
\newtheorem{lem_denseprekey}[lem_densekey]{Lemma}

\newtheorem{th_convergephi}[lem_densekey]{Theorem}
\newtheorem{th_convergepsi}[lem_densekey]{Theorem}
\newtheorem{th_convergephipsi}[lem_densekey]{Theorem}
\newtheorem{th_convergephipsi_spec}[lem_densekey]{Theorem}

\newtheorem{th_canexistgen}[lem_densekey]{Theorem}
\newtheorem{cor_canexistgen_simple}[lem_densekey]{Corollary}
\newtheorem{th_canexistgen_main}[lem_densekey]{Theorem}
\newtheorem{th_onerad}[lem_densekey]{Theorem}
\newtheorem{th_onerad_main}[lem_densekey]{Theorem}

\everypar{\looseness=-1}

\begin{document}

\title{Gradient Descent in Neural Networks as Sequential Learning in RKBS}
\author{Alistair Shilton, Sunil Gupta, Santu Rana, Svetha Venkatesh\\ \{alistair.shilton, sunil.gupta,santu.rana,svetha.venkatesh\}@deakin.edu.au}

\maketitle
%\thispagetitle

\begin{abstract}
The study of Neural Tangent Kernels (NTKs) has provided much needed insight 
into convergence and generalization properties of neural networks in the 
over-parametrized (wide) limit by approximating the network using a 
first-order Taylor expansion with respect to its weights in the neighborhood 
of their initialization values.  This allows neural network training to be 
analyzed from the perspective of reproducing kernel Hilbert spaces (RKHS), 
which is informative in the over-parametrized regime, but a poor approximation 
for narrower networks as the weights change more during training.  Our goal is 
to extend beyond the limits of NTK toward a more general theory.  We construct 
an exact power-series representation of the neural network in a finite 
neighborhood of the initial weights as an inner product of two feature maps, 
respectively from data and weight-step space, to feature space, allowing 
neural network training to be analyzed from the perspective of reproducing 
kernel {\em Banach} space (RKBS). We prove that, regardless of width, the 
training sequence produced by gradient descent can be exactly replicated by 
regularized sequential learning in RKBS.  Using this, we present novel bound 
on uniform convergence where the iterations count and learning rate play a 
central role, giving new theoretical insight into neural network training.
\end{abstract}

\section{Introduction} \label{sec:intro}

The remarkable progress made in neural networks in recent decades has led to an 
explosion in their adoption in a wide swathe of applications.  However this 
widespread success has also left unanswered questions, the most obvious of 
which is why non-convex, massively over-parameterised networks are able to 
perform so much better than predicted by traditional machine learning 
theory.

Neural tangent kernels represent an attempt to answer this question.  As per 
\citep{Jac2,Aro4}, during training, the evolution of an over-parameterised 
neural network follows the kernel gradient of the functional cost with respect 
to a neural tangent kernel (NTK).  It was shown that, for a sufficiently wide 
network with random weight initialisation, the NTK is effectively fixed, 
and results from machine learning in reproducing kernel Hilbert space (RKHS) 
can thus be brought to bear on the problem.  This has led to a plethora of 
results analysing the convergence \citep{Du1,All3,Du2,Zou1,Zou2} and 
generalisation \citep{Aro4,Aro6,Cao4} properties of neural networks.

Despite their successes, NTK models are not without problems.  As noted in 
\citep{Bai2}, the expressive power of the linear approximation used by NTK 
is limited to that of the corresponding, randomised feature space or RKHS, as 
evidenced by the observed gap between NTK predictions and actual performance.  
To break out of this regime, \citep{Bai2} proposed using a second or 
higher-order approximation of the network.  Moreover it is natural to ask 
how well a linear approximation of the behaviour, constructed on the 
assumption of small weight-steps, will scale to larger weight steps in 
narrower networks.

To overcome these difficulties we replace the Taylor approximation used in NTK 
with an an exact power series representation of the neural network in a finite 
neighbourhood around the initial weights.  We demonstrate that this leads to a 
representation as an inner product between two feature maps, from data and 
weight-step space respectively.  This structure underlies the construction of 
reproducing kernel Banach spaces (RKBS, \citep{Lin10}), allowing us to go on to 
show an equivalence between back-propagation and sequential learning in RKBS, 
which is similar to NTK but without the constraints of linearity, allowing us 
to derive new bounds on uniform convergence for networks of arbitrary width.

\section{Related Work} \label{sec:relwork}

There has been a significant amount of work looking at uniform convergence 
behaviour of networks of different types using variety of assumptions 
during training \citep{Ney1,Ney2,Ney3,Ney4,Har2,Bar8,Gol3,Aro3,Zey1,Dra2,Li8, 
Nag1,Nag2,Zho5}.

The study of the connection between kernel methods and neural networks has a 
long history.  \citep{Nea1} demonstrated that, in the infinite-width limit, 
iid randomly initialised single-layer networks converge to draws from a 
Gaussian process.  This was extended to multi-layered neural networks in 
\citep{Lee8,Mat5} by assuming random weights up to (but not including) the 
output layer.  Other works deriving approximate kernels by assuming random 
weights include \citep{Rah2,Bac3,Bac4,Dan1,Dan2}.

Neural tangent kernels \citep{Jac2,Aro4} are a more recent development.  The 
basis of NTK is to approximate the behaviour of neural network (for a given 
input ${\bf x}$) as the weights and biases vary about some initial values 
using a first-order Taylor approximation.  This approximation is linear in 
the change in weights, and the coefficients of this approximation are 
functions of ${\bf x}$ and may therefore be treated as a feature map, making 
the model amenable to the kernel trick and subsequent analysis in terms of 
RKHS theory.  This approach may be generalised to higher order approximations 
\citep{Bai2}, but the size of change in the weights that can be approximated 
remains limited except in the over-parametrised limit, where the variation of 
the weights becomes small.

Arc-cosine kernels \citep{Cho8} work on a similar premise.  For 
activation functions of the form $\tau (\xi) = (\xi)_+^p$, $p \in 
\infset{N}$, in the infinite-width limit, arc-cosine kernels capture the 
feature map of the network.  Depth is achieved by composition of kernels.  
However once again this approach is restricted to networks of infinite 
width, whereas our approach works for arbitrary networks.

Finally there has been some very recent work \citep{Bar9,San9,Par2,Uns1} in a 
similar vein to the current work, seeking to connect neural networks to RKBS 
theory.  However these works consider only $1$ and $2$ layer networks (we 
consider networks of arbitrary depth), and no equivalence is established 
between the weight-steps found by back-propagation and those found by 
regularised learning in RKBS.

\section{Notations}

Let $\infset{N} = \{ 0,1,2,\ldots \}$, $\infset{N}_n = \{ 0,1,\ldots, n-1 \}$. 
Vectors and matrices are denoted ${\bf a}$ and ${\bf A}$, respectively, with 
elements $a_i$, $A_{i,i'}$, and columns ${\bf A}_{:i}$ indexed by $i,i' \in 
\infset{N}$.  We define:
\[
 \begin{array}{l}
  \| {\bf x} \|_p = ( \sum_i |x_i|^p )^{1/p}, 
  \| {\bf A} \|_{p,q} = \| [ \| {\bf A}_{:i} \|_p ]_i \|_{q_{{\;}_{{{\;}_{\;}}}}} \\
  \| {\bf x} \|_\infty = \max_i \{ |x_i| \}, 
  \| {\bf x} \|_{-\infty} = \min_i \{ |x_i| \} \\
 \end{array}
\]
$\forall p,q \in [-\infty,0)\cup(0,\infty]$, which are norms if $p,q \in 
[1,\infty]$.  The Frobenius norm and inner product are $\| \cdot \|_F = \| 
\cdot \|_{2,2}$, $\left< {\bf A}, {\bf B} \right>_F = {\rm Tr} ({\bf A}^\tsp 
{\bf B})$.  The Kronecker and Hadamard product are ${\bf a} \otimes {\bf b}$, 
${\bf a} \odot {\bf b}$.  The Kronecker and Hadamard powers are ${\bf 
a}^{\otimes c} = {\bf a}\otimes{\bf a}\otimes\overset{c\;{\rm times}}{\ldots} 
\otimes {\bf a}$, ${\bf a}^{\odot c} = {\bf a} \odot {\bf a} \odot \overset{c 
\;{\rm times}}{\ldots} \odot {\bf a}$.  The elementwise absolute and sign are 
$| {\bf a} |$, $\sgn ({\bf a})$.  Finally, for vectors ${\bf a}, {\bf b}$ we 
let ${\bg{\varrho}} ({\bf a}, {\bf b}) = [ \; a_0 ({\bf b}^{\otimes 1})^\tsp, 
a_1 ({\bf b}^{\otimes 2})^\tsp, \ldots \; ]^\tsp$.  ${\rm diag} ({\bf A})$ is 
a vector containing the diagonal elements of ${\bf A}$, and conversely 
${\rm diag} ({\bf a})$ is a diagonal matrix with diagonal elements from ${\bf a}$.

We study fully connected $D$-layer neural networks ${\bf f} : (\infset{X} 
\subset \infset{R}^n) \to (\infset{Y} \subset \infset{R}^m)$ with layer widths 
$H^{[j]}$ (and $H^{[-1]} = n$) trained on a training set $\{ ({\bf x}^{\{k\}}, 
{\bf y}^{\{k\}}) \in \infset{X} \times \infset{Y} : k \in \infset{N}_N \}$. We 
use index range conventions $k \in \infset{N}_N$, $j \in \infset{N}_D$ and 
$i_{j+1} \in \infset{N}_{H^{[j]}}$, and for clarity we write:
\[
 \begin{array}{c}
  \scriptscriptstyle{{\mbox{\small Relating to: }}} \;\;\; 
  \scriptscriptstyle{\underline{\mbox{\small Training vector $k_{{\;\!}_{\;\!}}\!$}}} \;\;\; 
  \scriptscriptstyle{\underline{\mbox{\small Layer $j$}}} \;\;\;
  \scriptscriptstyle{\underline{\mbox{\small The $r^{\rm th}_{\;}$ derivative}}} \;\;\; \\
  \;\;\;\; \searrow \;\;\;\; \downarrow \;\;\;\; \swarrow \\
  X^{\{k\}[j](r)}_{\;\ldots\ldots} \\
  \!\!\!\nearrow \;\;\;\;\;\;\; \nwarrow \;\;\;\;\; \\
  \!\!\!\scriptscriptstyle{\overline{\mbox{\small Variable name}}} \;\;\;\;\;\; 
  \scriptscriptstyle{\overline{\mbox{\small Variable indices}}} \;\;\; \\
 \end{array}
\]
so for example ${\bf W}^{[j]}$ is the weight matrix for layer $j$, ${\bf 
x}^{\{k\}[j]}$ is the input (image) to layer $j$ of the network given network 
input ${\bf x}^{\{k\}}$, and $f^{(2)}(z)$ is the $2^{\rm nd}$ derivative of 
$f(z)$.  With regard to training, $X_\origin$ means ``value of variable $X$ 
before iteration'' and $X_{\changein}$ means ``change in $X$ due to iteration''. 
Finally, where 
relevant, we use a superscript $X^\backstep$ to indicate that $X$ relates to 
gradient descent (back-propagation), and $X^\regstep$ if $X$ relates to RKBS 
regularised risk minimisation.

\section{Background}

\subsection{Reproducing Kernel Banach Space}

\begin{figure*}
\begin{center}
\begin{tabular}{|l||l|l|}
 \hline 
 & Notation in present Paper & Notation used in \citep{Lin10} \\
 \hline 
 \hline 
 Data space:        & $\infset{X}         \subset \infset{R}^n$                                                                         & $\Omega_1$ (input space)  \\
 \hline 
 Weight-step space: & $\infset{W}_\origin \subset \prod_{j \in \infset{N}_D} \infset{R}^{H^{[j-1]} \times H^{[j]}} \times \infset{R}^{H^{[j]}}$ & $\Omega_2$ (weight space) \\
 \hline 
 Data Feature map:        & ${\bg{\Phi}}_\origin : \infset{X}         \to \mathcal{X}_\origin \subset \infset{R}^{\infty \times m}$ & $\Phi_1 : \Omega_1 \to \mathcal{W}_1$ \\
 \hline
 Weight-step feature map: & ${\bg{\Psi}}_\origin : \infset{W}_\origin \to \mathcal{W}_\origin \subset \infset{R}^{\infty \times m}$ & $\Phi_2 : \Omega_2 \to \mathcal{W}_2$ \\
 \hline
 Data Banach space:        & $\mathcal{X}_\origin = {{\rm span}} ({\bg{\Phi}}_{\origin} (\infset{X})        )$, $\| \cdot \|_{\mathcal{X}_\origin} = \| \cdot \|_{F}$ & $\mathcal{W}_1$ with norm $\| \cdot \|_{\mathcal{W}_1}$ \\
 \hline
 Weight-step Banach space: & $\mathcal{W}_\origin = {{\rm span}} ({\bg{\Psi}}_{\origin} (\infset{W}_\origin))$, $\| \cdot \|_{\mathcal{W}_\origin} = \| \cdot \|_{F}$ & $\mathcal{W}_2$ with norm $\| \cdot \|_{\mathcal{W}_2}$ \\
 \hline
 Bilinear form: & $\left< {{\bg{\Omega}}}, {{\bg{\Xi}}} \right>_{\mathcal{X}_\origin \times \mathcal{W}_\origin} = {\rm diag} \left( {{\bg{\Omega}}}^\tsp {{\bg{\Xi}}} \right)$ & $\left< \cdot, \cdot \right>_{{\mathcal{W}}_1 \times {\mathcal{W}}_2} : \mathcal{W}_1 \times \mathcal{W}_2 \to \infset{Y}$ \\
 \hline
\end{tabular}
\end{center}
\caption{Summary of the construction of reproducing kernel Banach space as per 
         \citep{Lin10}.}
\label{tab:banconstruct_main}
\end{figure*}

A reproducing kernel Hilbert space (RKHS) \citep{Aro1} is a Hilbert space 
$\mathcal{H}$ of functions $f : \infset{X} \to \infset{Y}$ for which the point 
evaluation functionals $\delta_{\bf x} (f) = f({\bf x})$ are continuous.  
Thus, applying the Riesz representor theorem, there exists a kernel $K$ such 
that:
\[
 \begin{array}{l}
  f \left( {\bf x} \right) = \left< f \left( \cdot \right), K \left( {\bf x}, \cdot \right) \right>_\mathcal{H} \;\;\;\;\forall f \in \mathcal{H}
 \end{array}
\]
Subsequently $K({\bf x}, {\bf x}') = \left< K({\bf x},\cdot), K({\bf x}', 
\cdot) \right>$ and, by the Moore-Aronszajn theorem, $K$ is uniquely 
defined by $\mathcal{H}$ and vice-versa.  $K$ is called the reproducing 
kernel, and the corresponding RKHS is denoted $\mathcal{H}_K$.  RKHS based 
approaches have gained popularity as they are well 
suited to many aspects of machine learning \citep{Ste3,Sha3}.  The inner 
product structure enables the kernel trick, 
and the kernel 
is readily understood as a similarity measure.  Furthermore 
the structure of RKHSs has led to a rich framework of 
complexity analysis and generalisation bounds \citep{Ste3,Sha3}.
More recently neural tangent 
kernels were introduced \citep{Jac2}, allowing RKHS theory to be applied to the 
neural network training in the over-parametrised regime.

In an effort to introduce a richer set of geometrical structures into RKHS 
theory, reproducing kernel Banach spaces (RKBSs) generalise RKHSs by starting 
with a Banach space of functions 
%\citep{Der1,Lin10,Zha11,Zha14,Son1,Sri3,Xu4}.  Precisely:
\citep{Der1,Zha11,Son1,Xu4,Lin10} etc.  Precisely:
\begin{def_rkbs}[Reproducing kernel Banach space (RKBS - \citep{Lin10})] A 
 reproducing kernel Banach space $\mathcal{B}$ on a set $\infset{X}$ is a 
 Banach space of functions $f : \infset{X} \to \infset{Y}$ such that every 
 point evaluation $\delta_{\bf x} : \mathcal{B} \to \infset{Y}$, ${\bf x} \in 
 \infset{X}$, on $\mathcal{B}$ is continuous (so $\forall {\bf x} \in 
 \infset{X}$ $\exists C_{\bf x} \in \infset{R}_+$ such that $|\delta_{\bf 
 x}(f)| = |f({\bf x})| \leq C_{\bf x} \|f\|_{\mathcal{B}}$ $\forall f \in 
 \mathcal{B}$).
 \label{def:rkbs_main}
\end{def_rkbs}

There are several distinct approaches to RKBS construction.  In the present 
context however we find the approach of \citep[Theorem 2.1]{Lin10} most 
convenient.  Given the components outlined in Figure 
\ref{tab:banconstruct_main}, and assuming that ${\bg{\Phi}}_\origin 
(\infset{X})$ is dense in $\mathcal{X}_\origin$  and that ${\bg{\Psi}}_\origin 
(\infset{W}_\origin)$ is dense in $\mathcal{W}_\origin$, we define the 
reproducing kernel Banach space ${{\mathcal{B}}}_\origin$ on $\infset{X}$ as:
\begin{equation}
 \begin{array}{l}
  {{\mathcal{B}}}_\origin = \left\{ \left. \left< {{\bg{\Phi}}}_\origin \left( \cdot \right), {\bg{\Omega}} \right>_{\mathcal{X}_\origin \times \mathcal{W}_\origin} \right| {\bg{\Omega}} \in {\mathcal{W}_\origin} \right\} \\
  \mbox{where: } 
  \left\| \left< {{\bg{\Phi}}}_\origin \left( \cdot \right), {\bg{\Omega}} \right>_{\mathcal{X}_\origin \times \mathcal{W}_\origin} \right\|_{{{\mathcal{B}}}_\origin} = \left\| {\bg{\Omega}} \right\|_{\mathcal{W}_\origin} \\
 \end{array}
\label{eq:banach_rkbs}
\end{equation}
with reproducing Banach kernel:
\begin{equation}
 \begin{array}{rl}
  K_\origin \left( {\bf x}, {\bf W}_\changein \right) = \left< {\bg{\Phi}}_\origin \left( {\bf x} \right), {\bg{\Psi}}_\origin \left( {\bf W}_\changein \right) \right>_{\mathcal{X}_\origin \times \mathcal{W}_\origin}
 \end{array}
\label{eq:banach_repro}
\end{equation}

\section{Setup and Assumptions} \label{sec:setup}

We assume a fully-connected, $D$-layer feedforward neural network ${\bf f} : 
(\infset{X} \subseteq \infset{R}^n) \to (\infset{Y} \subseteq \infset{R}^m)$ 
with layers of widths $H^{[0]}, H^{[1]}, \ldots, H^{[D-1]}$, where $H^{[D-1]} 
= m$ and we define $H^{[-1]} = n$.  We assume layer $j \in \infset{N}_D$ ($j 
\in \infset{N}_D$ throughout) contains only neurons with activation function 
$\tau^{[j]} : \infset{R} \to \infset{R}$.  The network is defined recursively 
$\forall j \in \infset{N}_D$:
\begin{equation}
 \begin{array}{rl}
  {\bf f} \left( {\bf x} \right) 
  &\!\!\!\!=
  {\bf x}^{[D]} \in \infset{R}^{H^{[D-1]}} \\
  {\bf x}^{[j+1]} 
  &\!\!\!\!= 
  \tau^{[j]} ( \tilde{\bf x}^{[j]} ) \in \infset{R}^{H^{[j]}} \\
  \tilde{\bf x}^{[j]} 
  &\!\!\!\!= 
  \frac{1}{\sqrt{H^{[j]}}} {\bf W}^{[j]\tsp} {\bf x}^{[j]} + \alpha^{[j]} {\bf b}^{[j]} \in \infset{R}^{H^{[j]}} \\
  {\bf x}^{[0]} 
  &\!\!\!\!= 
  {\bf x} \in \infset{X} \subset \infset{R}^{H^{[-1]}} \mbox{ ($H^{[-1]} = n$)} \\
 \end{array}
 \label{eq:yall_main}
\end{equation}
where ${\bf W}^{[j]} \in \infset{R}^{H^{[j-1]} \times H^{[j]}}$ and ${\bf 
b}^{[j]} \in \infset{R}^{H^{[j]}}$ are weights and biases, which we summarise 
as ${\bf W} \in \infset{W}$, and $\alpha^{[j]} \in \infset{R}_+$ are fixed.  
The set of functions of this form is denoted $\mathcal{F}$.

We assume the goal of training is to take a training set and find weights 
and biases to minimise the empirical risk:
\begin{equation}
 \begin{array}{l}
  {\bf W}^\star = \mathop{\rm argmin}\limits_{{\bf W} \in \infset{W}} R_E \left( {\bf W}, \infset{D} \right) \\
  R_E \left( {\bf W}, \infset{D} \right) = \sum_k E \left( {\bf x}^{\{k\}}, {\bf y}^{\{k\}}, {\bf f}_{\bf W} \left( {\bf x}^{\{k\}} \right) \right) \\
 \end{array}
 \label{eq:traingoalW_main}
\end{equation}
where ${\bf f}_{\bf W}$ is a network of the form (\ref{eq:yall_main}) with weights 
and biases ${\bf W}$, $\infset{D} = \{ ({\bf x}^{\{k\}},{\bf y}^{\{k\}}) \in 
\infset{X} \times \infset{Y} : k \in \infset{N}_N \}$ is a training set ($k 
\in \infset{N}_N$ throughout), and $E: \infset{X} \times \infset{Y} \times 
\infset{R}^m \to \infset{R}$ is an error function defining the purpose of the 
network.

We make the following technical assumptions:
\begin{enumerate}
 \item Input space: $\infset{X} = [-1,1]^n$.
 \item Error function: $E: \infset{X} \times \infset{Y} \times \infset{R}^m 
       \to \infset{R}$ is $\mathcal{C}^1$ and $L_E$-Lipschitz in its third 
       argument.
 \item Activation functions: for all $j \in \infset{N}_D$, $\tau^{[j]} : 
       \infset{R} \to [-1,1]^{H^{[j]}}$ is bounded, 
       $\mathcal{C}^\infty$, and has a power-series representations with 
       region of convergence (ROC) at least $\rho^{[j]} \in \infset{R}_+$ 
       around all $z \in \infset{R}$.
 \item Weight non-triviality: for all $j \in \infset{N}_D$, ${\bf W}^{[j]} \ne 
       {\bf 0}$ at all times during training.\footnote{Networks that do not 
       meet this requirement have a constant output independent of input 
       ${\bf x}$.  We do not consider this a restrictive assumption as it is 
       highly unlikely that a randomly initialised network trained with a 
       typical training set will ever reach this state.}
 \item Weight initialisation: we assume LeCun initialisation, so for all $j 
       \in \infset{N}_D$, $W_{i_j,i_{j+1}}^{[j]}, b_{i_{j+1}}^{[j]} \sim 
       \mathcal{N} (0,1)$.
 \item Training: we assume training is gradient descent (back-propagation) 
       with learning rate $\learnrate \in \infset{R}_+$. 
\end{enumerate}

\subsection{Back-Propogation Training} 

As stated above, we assume the network is trained using back-propagation 
(gradient descent) \citep{Goo1}.  
This is an iterative approach.  An iteration starts with initial weights and 
biases ${\bf W}_\origin \in \infset{W}$.  A weight-step:
\[
 \begin{array}{rl}
  {\bf W}_\changein^\backstep
  &\!\!\!\!=
  -\learnrate
  \left. \frac{\partial}{\partial {\bf W}} \sum_k E \left( {\bf x}^{\{k\}}, {\bf y}^{\{k\}}, {\bf f}_{\bf W} \left( {\bf x}^{\{k\}} \right) \right) \right|_{{\bf W} = {\bf W}_\origin}
 \end{array}
\]
is calculated, and weights and biases are updated as ${\bf W} = {\bf 
W}_\origin + {\bf W}_\changein^\backstep$.  Our notational convention for 
activations before an iteration, and the subsequent change due to a weight 
step, are given in figure \ref{fig:netstep}.  
The weight-step is \citep{Goo1} (see appendix \ref{append:dualform} for a derivation):
\begin{equation}
 \begin{array}{rl}
  {\bf W}_{\changein :i_{j+1}}^{[j]\backstep}
  &\!\!\!\!= 
  -\frac{\learnrate}{\sqrt{H^{[D-1]} H^{[D-2]} \ldots H^{[j+1]}}} \mathop{\sum}\limits_{k} 
  \gamma_{\origin i_{j+1}}^{\{k\}[j]} 
  \frac{{\bf x}_{\origin}^{\{k\}[j]}}{\sqrt{H^{[j]}}} \\ 

  b_{\changein :i_{j+1}}^{[j]\backstep}
  &\!\!\!\!= 
  -\frac{\learnrate}{\sqrt{H^{[D-1]} H^{[D-2]} \ldots H^{[j+1]}}} \mathop{\sum}\limits_{k} 
  \gamma_{\origin i_{j+1}}^{\{k\}[j]} 
  \alpha^{[j]} \\
 \end{array}
\label{eq:backproprep_main}
\end{equation}
for all $j \in \infset{N}_D, i_{j+1}$ where, recursively $\forall j \in \infset{N}_{D-1}$:
\[
 \begin{array}{rl}
  \gamma_{\origin i_{j}}^{\{k\}[j-1]}
  &\!\!\!\!= 
  \mathop{\sum}\limits_{i_{j+1}} \gamma_{\origin i_{j+1}}^{\{k\}[j]} W_{\origin i_{j},i^{\prime}_{j+1}}^{[j]} \tau^{[j-1](1)} \left( \tilde{x}_{\origin i_{j}}^{\{k\}[j-1]} \right) \\

  \gamma_{\origin i_D}^{\{k\}[D-1]}
  &\!\!\!\!=
   \nabla_{i_D} E \left( \ldots^{\{k\}} \right) \tau^{[D-1](1)} \left( \tilde{x}_{\origin i_D}^{\{k\}[D-1]} \right) \\
 \end{array}
\]
Note that the change in bias ${\bf b}_\changein^{[j]}$ is proportional to 
$\alpha^{[j]}$.

\section{Analysis of a Single Iteration}

\begin{figure*}
\begin{center}
\[
 \begin{array}{c}
 \begin{array}{ccccc}

 \left(\!\!\!
 \begin{array}{c}
 \mbox{{\bf Before Iteration}} \\
 \begin{array}{l}
  {\bf f}_\origin \left( {\bf x} \right) 
  =
  {\bf x}_\origin^{[D]} \in \infset{Y} \\
  {\bf x}_\origin^{[j+1]} 
  = 
  \tau^{[j]} \left( \tilde{\bf x}_\origin^{[j]} \right) \\
  \tilde{\bf x}_\origin^{[j]} 
  = 
  \frac{1}{\sqrt{H^{[j]}}} {\bf W}_\origin^{[j]\tsp} {\bf x}_\origin^{[j]} + \alpha^{[j]} {\bf b}_\origin^{[j]} \\
  {\bf x}_\origin^{[0]} 
  = 
  {\bf x} \in \infset{X} \\
 \end{array} \\
 \end{array} 
 \!\!\!\right)

 &\!\!\! + \!\!\!&

 \left(\!\!\!
 \begin{array}{c}
 \mbox{{\bf Weight-step Change}} \\
 \begin{array}{l}
  {\bf f}^{{\;}^{{{\;}^{\;}}}}_{\changein_{{\;}_{{{\;}_{\;}}}}} \!\!\!\!\!\left( {\bf x} \right) 
  = 
  {\bf f} \left( {\bf x} \right) - {\bf f}_\origin \left( {\bf x} \right) \\
  {\bf x}_{\changein_{{\;}_{{{\;}_{\;}}}}}^{[j+1]} 
  = 
  {\bf x}^{[j+1]} - {\bf x}_\origin^{[j+1]} \\
  \tilde{\bf x}_{\changein_{{\;}_{{{\;}_{\;}}}}}^{[j]}\!\!\!\! 
  = 
  \tilde{\bf x}^{[j]} - \tilde{\bf x}_\origin^{[j]} \\
  {\bf x}_\changein^{[0]} 
  = 
  {\bf x} \in {\bf 0}_n \\
 \end{array} \\
 \end{array}
 \!\!\!\right)

 &\!\!\! = \!\!\!&

 \left(\!\!\!
 \begin{array}{c}
 \mbox{{\bf After Iteration}} \\
 \begin{array}{l}
  {\bf f}^{{\;}^{{{\;}^{\;}}}}_{{\;}_{{{\;}_{\;}}}}\!\!\!\!\!\! \left( {\bf x} \right) 
  =
  {\bf x}^{[D]} \in \infset{Y} \\
  {\bf x}_{{\;}_{{\;}_{{\;}_{\;}}}}^{[j+1]}\! 
  = 
  \tau^{[j]} \left( \tilde{\bf x}_{{\;}_{{\;}_{\;}}}^{[j]} \!\right) \\
  \tilde{\bf x}_{{\;}_{{\;}_{{\;}_{\;}}}}^{[j]}\! 
  = 
  \frac{1}{\sqrt{H^{[j]}}} {\bf W}^{[j]\tsp} {\bf x}^{[j]} + \alpha^{[j]} {\bf b}^{[j]} \\
  {\bf x}_{{\;}_{{\;}_{{\;}_{\;}}}}^{[0]}\! 
  = 
  {\bf x} \in \infset{X} \\
 \end{array} \\
 \end{array}
 \!\!\!\right)

 \end{array} 
 \end{array}
\]
\end{center}
\caption{Definition of terms for neural network before and after an iteration.}
\label{fig:netstep}
\end{figure*}

In the first phase of our analysis we consider the change in neural network 
behaviour resulting from a small in weights and biases (a weight-step).  
The overall training 
sequence is readily extrapolated from this as per section \ref{sec:radcomp}.  
Our first goal is to rewrite the neural network after a training iteration as:
\begin{equation}
 \begin{array}{rl}
  {\bf f} \left( {\bf x} \right) 
  &\!\!\!\!= {\bf f}_\origin \left( {\bf x} \right) + {\bf f}_{\changein} \left( {\bf x} \right) \\
 \end{array}
\end{equation}
where ${\bf f}_\origin = {\bf f}_{{\bf W}_\origin} : ( \infset{X} \subset 
\infset{R}^n ) \to ( \infset{Y} \subset \infset{R}^m )$ is the neural network 
before the iteration and ${\bf f}_\changein:(\infset{X} \subset \infset{R}^n) 
\to \infset{R}^m$ is the change in network behaviour due to the change ${\bf 
W} = {\bf W}_\origin \to {\bf W} = {\bf W}_\origin + {\bf W}_\changein$ in 
weights and biases for this iteration, as detailed in Figure \ref{fig:netstep}, 
so that:
\begin{equation}
 \begin{array}{c}
  {\bf f}_{\changein} \left( {\bf x} \right) 
  = \left< 
             {\bg{\Phi}}_{\origin} \left( {\bf x}             \right),
             {\bg{\Psi}}_{\origin} \left( {\bf W}_{\changein} \right)
             \right>_{\mathcal{X}_\origin \times \mathcal{W}_{\origin_{{\;}_{\;}}}} \\

  \begin{array}{rl}
  \mbox{where:}
  & {\bg{\Phi}}_{\origin} : \infset{X}         \to \mathcal{X}_\origin = {\rm span} \left( {{\bg{\Phi}}}_\origin \left( \infset{X}         \right) \right) \subset \infset{R}^{\infty \times m} \\
  & {\bg{\Psi}}_{\origin} : \infset{W}_\origin \to \mathcal{W}_\origin = {\rm span} \left( {{\bg{\Psi}}}_\origin \left( \infset{W}_\origin \right) \right) \subset \infset{R}^{\infty \times m} \\
 \end{array}
 \end{array}
\label{eq:fstep_main}
\end{equation}
are feature maps determined entirely by the structure of the network (number 
and width of layers, activation functions) and the initial weights and biases 
${\bf W}_\origin$; and $\left< {\bg{\Xi}}, {\bg{\Omega}} 
\right>_{\mathcal{X}_\origin \times \mathcal{W}_\origin} = {\rm diag} \left( 
{\bg{\Xi}}^\tsp {\bg{\Omega}} \right)$ is a bilinear form.  Subsequently our 
second goal is to derive kernels and norms from these feature maps to allow us 
to study their convergence properties, and we finish by proving an equivalence 
between the weight-step ${\bf W}_{\changein}^{\backstep}$ due to a single step 
of back-propagation and the analogous weight-step ${\bf W}_\changein^\regstep$ 
that minimises the (RKBS) regularised risk.

\subsection{Contribution 1: Feature-Map Expansion}

In this section we derive appropriate feature maps to express the change in 
neural network behaviour for a finite weight-step.  Our approach is simple 
in principle but technical, so details are reserved for appendix 
\ref{append:dualform}.  Roughly speaking however, we begin by noting that,
for a smooth activation function $\tau^{[j]} 
: \infset{R} \to \infset{R}$, $z \in \infset{R}$ and finite-dimensional 
vectors ${\bf c}, {\bf c}'$ whose inner product lies in the radius of 
convergence $\rho^{[j]}$ (so that $|\left<{\bf c}, {\bf 
c}\right>| < \rho^{[j]}$), the power-series 
representation of $\tau^{[j]}$ about $z$ can be written 
$\tau^{[j]} ( z + \left< {\bf c}, {\bf c} \right>) = \tau^{[j]} (z) + \left< 
{\bg{\varrho}} ({\bf g}^{[j]}(z), {\bf c}), {\bg{\varrho}} ({\bf 1}_\infty, {\bf c}') \right>$, 
where:
\[
 {{
 \begin{array}{rl}
  {\bg{\varrho}} \left( {\bf a}, {\bf d} \right) 
  &\!\!\!\!= 
  \left[ \begin{array}{c} 
   a_0 {\bf d}^{\otimes 1\tsp} \; 
   a_1 {\bf d}^{\otimes 2\tsp} \; 
   a_2 {\bf d}^{\otimes 3\tsp} \; 
   \ldots \\ 
  \end{array} \right]^\tsp
  \\
  {\bf g}^{[j]} \left( z \right) 
  &\!\!\!\!= 
  \left[ \begin{array}{c}
   \frac{1}{1!} \tau^{[j](1)} \left( z \right) \;
   \frac{1}{2!} \tau^{[j](2)} \left( z \right) \;
   \frac{1}{3!} \tau^{[j](3)} \left( z \right) \;
   \ldots \\
  \end{array} \right]^\tsp
 \end{array}
 }}
\]
Given an input ${\bf x}$, starting at layer $0$ and working forward, 
and with reference to Figure \ref{fig:netstep}, we can write the change 
${\bf x}_\changein^{[1]}$ in the output of layer $0$ due to the weight-step 
${\bf W}_\changein$ as:
\begin{equation}
 \begin{array}{c}
 \begin{array}{rl}
  {\bf x}_{\changein}^{[1]}
  = \left[ \left< 
             {\bg{\Phi}}_{\origin :i_1}^{[0]} \left( {\bf x} \right),
             {\bg{\Psi}}_{\origin :i_1}^{[0]} \left( {\bf W}_\changein \right)
            \right> \right]_{i_1} \\
 \end{array} \\
 \begin{array}{rl}
  {\mbox{where: }}
  {\bg{\Phi}}_{\origin:i_{1}}^{[0]} \left( {\bf x} \right)
   &\!\!\!\!= 
             {\bg{\varrho}} \left( {\bf g}^{[0]} \left( \tilde{x}_{\origin i_1}^{[0]} \right),
             \shadx_{i_1}^{[0]} \left[ \begin{array}{c} 
                    \frac{1}{\sqrt{2       }} \alpha^{[0]}    \\ 
                    \frac{1}{\sqrt{2H^{[0]}}} {\bf x}_\origin \\ 
             \end{array} \right] 
             \right) \\

  {\bg{\Psi}}_{\origin:i_{1}}^{[0]} \left( {\bf W}_{\changein} \right)
  &\!\!\!\!= 
             {\bg{\varrho}} \left( {\bf 1}_\infty,
             \frac{1}{\shadx_{i_1}^{[0]}} \left[ \begin{array}{c}
                    \sqrt{2} {    b}_{\changein  i_1}^{[0]} \\
                    \sqrt{2} {\bf W}_{\changein :i_1}^{[0]} \\
             \end{array} \right] 
             \right) \\
 \end{array} \\
 \end{array}
\label{eq:featmapfirst}
\end{equation}
where we note that both feature maps have a finite radius of convergence.  
The feature maps are parameterised by the scale factors $\shadx_{i_1}^{[0]} 
\in \infset{R}_+$ whose role is mainly technical, insofar as they will allow 
us to show equivalence between RKBS regularised risk minimisation and 
back-propogation.\footnote{See appendix for more discussion.}  
Their exact value (beyond existence) is unimportant here.

The process is repeated for subsequent layers (see appendix \ref{append:dualform} for 
details).  After working through all layers:
\begin{equation}
 \begin{array}{rl}
  {\bf f}_{\changein} \left( {\bf x} \right) 
  &\!\!\!\!= \left< 
             {\bg{\Phi}}_{\origin} \left( {\bf x} \right),
             {\bg{\Psi}}_{\origin} \left( {\bf W}_\changein \right)
             \right>_{\mathcal{X}_\origin \times \mathcal{W}_\origin}
 \end{array}
 \label{eq:yfrombase_main}
\end{equation}
where ${\bg{\Phi}}_{\origin} = {\bg{\Phi}}_{\origin}^{[D-1]}$, 
${\bg{\Psi}}_{\origin} = {\bg{\Psi}}_{\origin}^{[D-1]}$, and 
$\forall j \in \infset{N}_D \backslash \{0\}$:
\begin{equation}
 \begin{array}{c}
  \begin{array}{l}
   {\bg{\Phi}}_{\origin:i_{j+1}}^{[j]} \left( {\bf x} \right)
   = 
              {\bg{\varrho}} \left( {\bf g}^{[j]} \left( \tilde{x}_{\origin i_{j+1}}^{[j]} \right),
              {\shadx}_{i_{j+1}}^{[j]} \left[ \begin{array}{c}
                     \left[ \begin{array}{c} \frac{1}{\sqrt{2}} \alpha^{[j]} \\ \frac{1}{\sqrt{H^{[j]}}} {\bf x}_{\origin}^{[j]} \\ \end{array} \right] \\
                     \left[ \begin{array}{c} \frac{{\shadw}_{i_j}^{[j]} W_{\origin i_j, i_{j+1}}^{[j]}}{{\shadwo}_{i_ji_{j+1}}^{[j]} \sqrt{H^{[j]}}} {\bg{\Phi}}_{\origin:i_j}^{[j-1]} \left( {\bf x} \right) \\ \end{array} \right]_{i_j} \\
                     \left[ \begin{array}{c} \frac{{\shadw}_{i_j}^{[j]}                               }{\sqrt{H^{[j]}}} {\bg{\Phi}}_{\origin:i_j}^{[j-1]} \left( {\bf x} \right) \\ \end{array} \right]_{i_j}  \\
              \end{array} \right]
              \right) \\

   {\bg{\Psi}}_{\origin:i_{j+1}}^{[j]} \left( {\bf W}_{\changein} \right)
   = 
              {\bg{\varrho}} \left( {\bf 1}_\infty,
              \frac{1}{{\shadx}_{i_{j+1}}^{[j]}} \left[ \begin{array}{c}
                     \left[ \begin{array}{c} \sqrt{2} b_{\changein i_{j+1}}^{[j]} \\ {\bf W}_{\changein :i_{j+1}}^{[j]} \\ \end{array} \right]  \\
                     \left[ \begin{array}{c} \frac{{\shadwo}_{i_j,i_{j+1}}^{[j]}}{{\shadw}_{i_j}^{[j]}} {\bg{\Psi}}_{\origin :i_j}^{[j-1]} \left( {\bf W}_\changein \right) \\ \end{array} \right]_{i_j} \\
                     \left[ \begin{array}{c} \frac{W_{\changein i_j, i_{j+1}}^{[j]}}{{\shadw}_{i_j}^{[j]}} {\bg{\Psi}}_{\origin :i_j}^{[j-1]} \left( {\bf W}_\changein \right) \\ \end{array} \right]_{i_j} \\
              \end{array} \right]
              \right) \\
  \end{array} \\
 \end{array}
\label{eq:featmaprest}
\end{equation}
recursively $\forall j \in \infset{N}_D \backslash \{0\}$ 
which are parameterised by scale factors $\shadx_{i_{j+1}}^{[j]} \in \infset{R}_+$ 
and shadow weights $\shadw_{i_{j}}^{[j]}, \shadwo_{i_j,i_{j+1}}^{[j]} \in 
\infset{R}_+$, which play a role in the equivalence between RKBS regularised 
risk minimisation and back-propagation (otherwise their exact values are 
unimportant).

\subsection{Contribution 2: Induced Kernels and Norms}

In the previous section we established that, as a result of a single 
weight-step ${\bf W}_\changein$, we can write:
\[
 \begin{array}{rl}
  {\bf f} \left( {\bf x} \right) 
  &\!\!\!\!= {\bf f}_\origin \left( {\bf x} \right) + 
             {\bf f}_{\changein} \left( {\bf x} \right) \\
  {\bf f}_{\changein} \left( {\bf x} \right) 
  &\!\!\!\!= \left< 
             {\bg{\Phi}}_{\origin} \left( {\bf x} \right),
             {\bg{\Psi}}_{\origin} \left( {\bf W}_\changein \right)
             \right>_{\mathcal{X}_\origin \times \mathcal{W}_\origin}
 \end{array}
\]
where ${\bf f}_\origin : \infset{X} \to \infset{Y}$ is the neural network 
pre-iteration and the feature maps ${\bg{\Phi}}_{\origin}$, 
${\bg{\Psi}}_{\origin}$ are feature maps (\ref{eq:featmapfirst}-\ref{eq:featmaprest}).  
Using these, we induce kernels on $\infset{X}$, $\infset{W}_\origin$ using the kernel 
trick:
\[
 \begin{array}{rl}
  {{\bf K}}_{\mathcal{X}_\origin} \left( {\bf x}, {\bf x}' \right) 
  &\!\!\!\!= 
  \left< {{\bg{\Phi}}}_{\origin} \left( {\bf x}           \right), {{\bg{\Phi}}}_{\origin} \left( {\bf x}'           \right) \right>_{\mathcal{X}_\origin \times \mathcal{X}_\origin} \\
  &\!\!\!\!= 
  {{\bg{\Phi}}}_{\origin}^\tsp \left( {\bf x}           \right) {{\bg{\Phi}}}_{\origin} \left( {\bf x}'           \right) \\

  {{\bf K}}_{\mathcal{W}_\origin} \left( {\bf W}_\changein, {\bf W}'_\changein \right) 
  &\!\!\!\!= 
  \left< {{\bg{\Psi}}}_{\origin} \left( {\bf W}_\changein \right), {{\bg{\Psi}}}_{\origin} \left( {\bf W}'_\changein \right) \right>_{\mathcal{W}_\origin \times \mathcal{W}_\origin} \\
  &\!\!\!\!= 
  {{\bg{\Psi}}}_{\origin}^\tsp \left( {\bf W}_\changein \right) {{\bg{\Psi}}}_{\origin} \left( {\bf W}'_\changein \right) \\
 \end{array}
\]
We call these kernels {\em neural 
neighbourhood kernels} (NNK) as they describe the similarity structure in 
the finite neighbourhood of the ${\bf W}_\origin$ (c/f NTK, which is the 
behaviour tangent to, or in the infinitessimal neighbourhood of, ${\bf 
W}_\origin$).  These matrix-valued kernels are symmetric and positive 
definite by construction, and could potentially be used (transfered) in 
support vector machines (SVMs) or similar kernel-based methods, measuring 
similarity on $\infset{X}$ and $\infset{W}_\origin$, respectively.  
Similarly, we induce a Banach kernel:
\begin{equation}
 \begin{array}{rl}
  {{\bf K}}_{\origin} \left( {\bf x}, {\bf W}'_\changein \right) 
  &\!\!\!\!= 
  \left< {{\bg{\Phi}}}_{\origin} \left( {\bf x} \right), {{\bg{\Psi}}}_{\origin} \left( {\bf W}'_\changein \right) \right>_{\mathcal{X}_\origin \times \mathcal{W}_\origin} \\
  &\!\!\!\!= 
  {\rm diag} \left( {\bf f}_{\changein} \left( {\bf x} \right) \right) \\
 \end{array}
\label{eq:inducedbanachkernel}
\end{equation}
which is trivially the change in the network output ${\bf 
f}_\changein ({\bf x})$ under weight-step ${\bf W}_\changein$ for input vector 
${\bf x}$, diagonalised.

The precise form of the neural neighbourhood kernels is rather 
complicated (the derivation is straightforward but resulting recursive 
equation is very long).  
The NNK is the (un-approximated) analog of the NTK.  However while neural 
networks evolve - to first order - in the RKHS defined by the NTK, the same is 
not true of the NNK.  Indeed, it is not difficult to see that {\em RKHS 
regularisation using the NNK will always result in a weight vector that 
is {\em not} the image of a weight-step under ${\bg{\Psi}}_\origin$ 
(i.e. RKHS theory is insufficient, and we need RKBS 
theory to proceed).}  Thus, as the NNK is not the main focus of our paper will 
not reproduce them in the body of the paper - the interested reader can find 
them in appendix \ref{sec:inducedkernelsetc}. 
Using these 
induced kernels we may obtain expressions for the norms of the images of 
$\infset{X}$ and $\infset{W}_\origin$ in feature space:
\begin{equation}
 \begin{array}{rl}
  \left\| {{\bg{\Phi}}}_{\origin} \left( {\bf x} \right) \right\|_{\mathcal{X}_\origin}^2
  &\!\!\!\!=
  \left\| {{\bg{\Phi}}}_{\origin}^{[D-1]} \left( {\bf x} \right) \right\|_F^2 \\
  &\!\!\!\!=
  {\rm Tr} \left( {\bf K}_{\mathcal{X}_\origin} \left( {\bf x}, {\bf x} \right) \right) \\

  \left\| {{\bg{\Psi}}}_{\origin} \left( {\bf W}_\changein \right) \right\|_{\mathcal{W}_\origin}^2
  &\!\!\!\!=
  \left\| {{\bg{\Psi}}}_{\origin}^{[D-1]} \left( {\bf W}_\changein \right) \right\|_F^2 \\
  &\!\!\!\!=
  {\rm Tr} \left( {\bf K}_{\mathcal{W}_\origin} \left( {\bf W}_\changein, {\bf W}_\changein \right) \right) \\
 \end{array}
\end{equation}
Once again, the precise form of these expressions is complicated (the 
derivation is straightforward but the answer is very long - details can 
be found in appendix \ref{sec:inducedkernelsetc}).  The importance of these 
norms lies in deriving conditions on convergence of the feature maps.  
Defining the helper function:
\begin{equation}
 \begin{array}{rl}
  \overline{\sigma}^{[j]} \left( \zeta \right)
  &\!\!\!\!=
  \mathop{\max}\limits_{z \in \infset{R}_+ \cup \{0\}} \left\{ \mathop{\sum}\limits_{l=1}^\infty \left( \frac{1}{l!} \right)^2 \tau^{[j](l)} \left( z \right) \tau^{[j](l)} \left( z \right) \zeta^l \right\} \\
 \end{array}
\label{eq:euceuc_lambda_main}
\end{equation}
and the constants:
\begin{equation}
 {{
 \begin{array}{rl}
  s^{[j]2}
  &\!\!\!\!=
  \left\{ \begin{array}{ll}
   \frac{1}{2} \alpha^{[j]2} + \frac{1}{2} \frac{H^{ [-1]}}{H^{[0]}} & \mbox{if } j = 0 \\
   \frac{1}{2} \alpha^{[j]2} +             \frac{H^{[j-1]}}{H^{[j]}} & \mbox{otherwise} \\
  \end{array} \right. \\

  t_{\changein i_{j+1}}^{[j]2} 
  &\!\!\!\!= 
  \left\{ \begin{array}{ll}
   \left[ 2b_{\changein i_{  1}}^{[0]2} + 2 \left\| {\bf W}_{\changein :i_{  1}}^{[0]} \right\|_2^2 \right]_{i_{  1}} & \mbox{if } j = 0 \\
   \left[ 2b_{\changein i_{j+1}}^{[j]2} +   \left\| {\bf W}_{\changein :i_{j+1}}^{[j]} \right\|_2^2 \right]_{i_{j+1}} & \mbox{otherwise} \\
  \end{array} \right. \\
 \end{array}
 }}
\label{eq:st_define}
\end{equation}
which are, loosely speaking, surrogates for, respectively, 
the expansion/contraction (fanout) in width from layer $j$ to layer 
$j-1$ and the size of the weight-step at layer $j$; 
in appendix \ref{append:convergecond} we derive the following Lemmas 
(see corresponding proofs of Theorems \ref{th:convergephi} and \ref{th:convergepsi} 
in the appendix):
\begin{th_convergephi_main}
 Let $\phidual^{[j]} \in (0,1)$ $\forall j \in \infset{N}_D$.  For a given 
 neural network and initial weights ${\bf W}_\origin$ define 
 $\frac{\phi^{[j]}}{H^{[j]}} = \overline{\sigma}^{[j]} ((1-\phidual^{[j]}) 
 \sqrt{{\rho}^{[j]}})$ $\forall j \in \infset{N}_D$.  If the scale factors 
 satisfy:
 \[
  \begin{array}{rll}
                     {\shadx}_{i_{j+1}}^{[j]2}
                     &\leq \left\{ \begin{array}{ll}
                     \frac{1}{\frac{s^{[j]2}}{\left( 1-\phidual^{[j]} \right) {{{\rho}}^{[j]\frac{1}{2}}}}
                     +
                     \frac{1}{H^{[j]}} \mathop{\sum}\limits_{i_j} \shadw_{i_j}^{[j]2} \left( \frac{W_{\origin i_j i_{j+1}}^{[j]2}}{\shadwo_{i_j,i_{j+1}}^{[j]2}} + 1 \right) 
                     \frac{\frac{\phi^{[j-1]}}{H^{[j-1]}}}{\left( 1-\phidual^{[j]} \right) {{{\rho}}^{[j]\frac{1}{2}}}}} 
                     & \mbox{if } j > 0 \\ \\

                     \frac{1}{\left( \frac{s^{[0]2}}{\left( 1-\phidual^{[0]} \right) \sqrt{{{\rho}}^{[0]}}} \right)}
                     & \mbox{if } j = 0 \\
                     \end{array} \right.
  \end{array}
 \]
 $\forall j \in \infset{N}_D, i_{j+1}$ then $\left\| {\bg{\Phi}}_\origin^{[j]} 
 ({\bf x}) \right\|_F^2 \leq \phi^{[j]}$ $\forall {\bf x} \in \infset{X}$.
 \label{th:convergephi_main}
\end{th_convergephi_main}
\begin{th_convergepsi_main}
 Let $\psidual^{[j]} \in (0,1)$ $\forall j \in \infset{N}_D$.  For a given 
 neural network and initial weights ${\bf W}_\origin$ 
 and weight-step ${\bf W}_\changein$, if:
 \[
  \begin{array}{l}
   \begin{array}{r}
   t_{\changein i_{1}}^{[0]2}
   \leq
   \left( 1 - \psidual^{[0]} \right) {\shadx}_{i_{1}}^{[0]2}
   \end{array} \\

   \begin{array}{r}
   t_{\changein i_{j+1}}^{[j]2} + \mathop{\sum}\limits_{i_j} \frac{\left\| {{\bg{\Psi}}}_{\origin:i_{j}}^{[j-1]} \left( {\bf W}_\changein \right) \right\|_2^2}{\shadw_{i_j}^{[j]2}} 
   \left( {\shadwo}_{i_j,i_{j+1}}^{[j]2} + W_{\changein i_j, i_{j+1}}^{[j]2} \right)
   \leq
   \left( 1 - \psidual^{[j]} \right) {\shadx}_{i_{j+1}}^{[j]2}
   \end{array} \\
  \end{array}
 \]
 $\forall j \in \infset{N}_D, i_{j'+1}$ then $\left\| {{\bg{\Psi}}}_\origin^{[j]} ({\bf W}_\changein) 
 \right\|_F^2 \leq H^{[j]} \frac{1-\psidual^{[j]}}{\psidual^{[j]}}$.
 \label{th:convergepsi_main}
\end{th_convergepsi_main}
Lemma \ref{th:convergephi_main} allows us to place bounds on the scale factors 
to ensure that the feature map ${\bg{\Phi}}_\origin^{[j]} : \infset{X} \to 
\mathcal{X}_\origin$ is finite (well defined) for all ${\bf x} \in \infset{X}$.  
Lemma \ref{th:convergepsi_main} is similar, but rather than bounding the scale 
factors it takes these as given (by Lemma \ref{th:convergephi_main}) and places 
bounds on the size of the weight-step ${\bf W}_\changein$ for which the feature 
map is ${\bg{\Psi}}_\origin^{[j]} : \infset{W} \to \mathcal{W}_\origin$ is finite 
(well defined).  Taken together, therefore, they give some bound on the size of 
weight-step that can be modelled for a give neural network structure and initial 
weights ${\bf W}_\origin$.  However they say nothing directly about the shadow 
weights.  In the next section we use these Lemmas to establish a link between 
the weight-step generated by gradient descent and learning in RKBS, as well as 
clarifying the size of weight-step which we can model using our construct.

\subsection{Contribution 3: Equivalence of Gradient Descent and regularised Risk Minimisation in RKBS}

In our previous contributions we showed that the change ${\bf f}_\changein$ in 
neural network behaviour can be represented as by the form 
(\ref{eq:yfrombase_main}) with feature maps (\ref{eq:featmapfirst},\ref{eq:featmaprest}), 
derived kernels from these feature maps and gave 
bounds on the size of the weight-step for which this representation is valid.  
In this section we use these results to establish a link between gradient 
descent learning in neural networks and regularised risk minimisation in 
reproducing kernel Banach space (RKBS).

To begin, it is not difficult to see that the feature maps 
${\bg{\Phi}}_\origin : \infset{X} \to \mathcal{X}_\origin$ and 
${\bg{\Psi}}_\origin : \infset{W}_\origin \to \mathcal{W}_\origin$ define a 
RKBS $\mathcal{B}_\origin$ imbued with $\| \cdot \|_{\mathcal{B}_\origin} = 
\| \cdot \|_F$ using (\ref{eq:banach_rkbs}) with the reproducing Banach kernel 
(\ref{eq:inducedbanachkernel}) deriving from (\ref{eq:banach_repro}):\footnote{In 
the appendix we prove that these maps satisfy the relevant density requirements.}
\[
 \begin{array}{rl}
  {{\bf K}}_{\origin} \left( {\bf x}, {\bf W}'_\changein \right) 
  &\!\!\!\!= 
  \left< {{\bg{\Phi}}}_{\origin} \left( {\bf x} \right), {{\bg{\Psi}}}_{\origin} \left( {\bf W}'_\changein \right) \right>_{\mathcal{X}_\origin \times \mathcal{W}_\origin} \\
  &\!\!\!\!= 
  {\rm diag} \left( {\bf f}_{\changein} \left( {\bf x} \right) \right) \\
 \end{array}
\]
in terms of which the change ${\bf f}_\changein^\backstep$ in the network's 
behaviour due to a back-progation iteration may be written:
\[
 \begin{array}{rl}
  {\bf f}_\changein^\backstep \left( \cdot \right)
  = 
  {\bf K}_\origin \left( \cdot, {\bf W}_\changein^\backstep \right) {\bf 1}
 \end{array}
\]

For a given neural network with initial weights ${\bf W}_\origin$, we assume 
that the weight-step is chosen using gradient descent.\footnote{Note that the training 
set $\infset{D}$ here is for this iteration only and may be a random subset 
of a larger training set.}  An alternative approach might be to select a 
weight-step to minimise the regularised risk in RKBS, specifically:
\begin{equation}
 \begin{array}{l}
  {\bf W}_\changein^\regstep = \mathop{\rm argmin}\limits_{{\bf W}_\changein \in \infset{W}_\origin} R_\lambda \left( {\bf W}_\changein \right) \\
  R_\lambda \left( {\bf W}_\changein \right) 
  =
  \lambda \left\| {\bg{\Psi}}_\origin \left( {\bf W}_\changein \right) \right\|_{\mathcal{W}_\origin}
  + 
  R_E \left( {\bf W}_\changein + {\bf W}_\origin, \infset{D} \right) \\
 \end{array}
\label{eq:Wchangein_main}
\end{equation}
where we call $\lambda$ the trade-off coefficient.  Larger trade-off 
coefficients favour smaller weight-steps, and vice-versa.  The advantage of 
this form over the back-propagation derived weight-step is that we can 
directly apply complexity bounds etc. from RKBS theory, and then extend to the 
complete training process.  This motivates us to ask:
\begin{quote}
 {\it For a given neural network with initial weights and biases ${\bf 
 W}_\origin$, let ${\bf W}_\changein^\backstep$ be the back-propagation 
 weight-step (gradient descent with learning rate $\learnrate$) defined by 
 (\ref{eq:backproprep_main}), and let ${\bf W}_\changein^\regstep$ be a 
 weight-step solving the regularised risk minimisation problem 
 (\ref{eq:Wchangein_main}).  Given the gradient-descent derived weight-step 
 ${\bf W}_\changein^\backstep$, can we select scale factors, shadow weights 
 and trade-off parameter $\lambda$ (as a function of ${\bf 
 W}_\changein^\backstep$) that would guarantee that ${\bf 
 W}_\changein^\regstep = {\bf W}_\changein^\backstep$?}
\end{quote}
If the answer is yes (which we demonstrate) then we can gain understanding of 
back-propagation by analysing (\ref{eq:Wchangein_main}).  Now, the solution to 
(\ref{eq:Wchangein_main}) must satisfy first-order optimality conditions (assuming 
differentiability for simplicity), so:
\[
{{
 \begin{array}{r}
  \frac{\partial}{\partial {\bf W}_\changein} 
  \left\| {\bg{\Psi}}_\origin \left( {\bf W}_\changein^\regstep \right) \right\|_{\mathcal{W}_\origin} 
  =
  \frac{-1}{\lambda}
  \frac{\partial}{\partial {\bf W}_\changein} R_E \left( {\bf W}_\changein^\regstep + {\bf W}_\origin, \infset{D} \right)
 \end{array}
}}
\]
Note that {\em if} the gradient of the regularisation term satisfies:
\[
 \begin{array}{rl}
  \frac{\partial}{\partial {\bf W}_\changein} \left\| {\bg{\Psi}}_\origin \left( {\bf W}_\changein^\backstep \right) \right\|_{\mathcal{W}_\origin} = \bangrad {\bf W}_\changein^\backstep
 \end{array}
\]
for some $\bangrad \in \infset{R}_+$, and $\frac{1}{\lambda} = {\learnrate 
\bangrad}$, then ${\bf W}_\changein^\regstep = {\bf 
W}_\changein^\backstep$.  Thus the question of whether there exists scaling 
factors, shadow weights and $\lambda$ such that the regularised risk 
minimisation weight-step corresponds to the gradient-descent weight-step for a 
specified learning rate $\learnrate$ can be answered in the affirmative by 
proving the existence of canonical scalings, which we define as follows:
\begin{def_canonical_scaling_main}[Canonical Scaling]
 For a given neural network, initial weights ${\bf W}_\origin$ and weight step 
 ${\bf W}_{\changein}^\backstep$ generated by back-propagation, we define a 
 {\em canonical scaling} to be a set of scaling factors and shadow weights for 
 which 
   $\frac{\partial}{\partial {\bf W}_\changein} \left\| {\bg{\Psi}}_\origin \left( {\bf W}_\changein^\backstep \right) \right\|_{\mathcal{W}_\origin} = \bangrad {\bf W}_\changein^\backstep$
 for $\bangrad \in \infset{R}_+$, and $\left\| {\bg{\Psi}}_\origin ( 
 {\bf W}_\changein^\backstep) \right\|_{\mathcal{W}_\origin}, \left\| 
 {\bg{\Phi}}_\origin ({\bf x}) \right\|_{\mathcal{X}_\origin} < \infty$ 
 $\forall {\bf x} \in \infset{X}$.
\end{def_canonical_scaling_main}

To prove the existence of canonical scaling and thus the key connection 
between gradient descent (back-propogation) and learning in RKBS, using 
(\ref{eq:st_define}), defining:
\[
 \begin{array}{rl}
  {t}_{\changein i_{j+1}}^{[j]\backstep 2} 
  &\!\!\!\!= 
  \left\{ \begin{array}{ll}
   2b_{\changein i_{  1}}^{[0]\backstep 2} + 2 \left\| {\bf W}_{\changein :i_{  1}}^{[0]\backstep} \right\|_2^2  
   & \mbox{if } j = 0 \\
   2b_{\changein i_{j+1}}^{[j]\backstep 2} + 2 \left\| {\bf W}_{\changein :i_{j+1}}^{[j]\backstep} \right\|_2^2
   & \mbox{otherwise} \\
  \end{array} \right. \\
 \end{array}
\]
$\forall j \in \infset{N}_D, i_{j+1}$, in the appendix we prove the following 
key result based on Lemmas \ref{th:convergephi_main} and \ref{th:convergepsi_main}:
\begin{th_canexistgen_main}
 Let $\epsilon, \chi \in (0,1)$.  For a given neural network 
 with initial weights ${\bf W}_\origin$, let ${\bf W}_\changein^\backstep$ be 
 the weight-step for this derived from back-propagation, assuming wlog that 
 $\alpha^{[j]} \in \infset{R}_+$ is chosen such that $\forall j \in 
 \infset{N}_{D-1}$:\footnote{Note that $b_{\changein i_{j+1}}^{[j]\backstep}$ 
 is proportional to $\alpha^{[j]}$, so we can always increase 
 ${t}_{\changein i_{j+1}}^{[j]\backstep 2}$ to ensure the 
 condition holds by adjusting $\alpha^{[j]}$.}
 \[
  \begin{array}{rl}
   \left\| {\bf W}_{\changein}^{[j+1]\backstep} \right\|_F^2
   =
   \chi \left\| {\bf t}_{\changein}^{[j]\backstep} \right\|_{\infty}^2 \\
  \end{array}
 \]
 Let $\psidual = 1-\frac{1}{1-\chi}\frac{s^{[D-1]2}}{(1-\epsilon) \sqrt{{{\rho}}^{[D-1]}}} 
 \| {\bf t}_{\changein}^{[D-1]\backstep} \|_\infty^2$ and:\footnote{The positivity 
 of $\psidual$ is due to the constraints on $\| {\bf t}_{\changein}^{[D-1]\backstep} 
 \|_\infty^2$.}
 \[
  \begin{array}{l}
   \phidual^{[j]}
   =  1 - \frac{1}{\sqrt{\rho^{[j]}}} \overline{\sigma}^{[j]-1} \left( 
      \frac{{\frac{1}{1-\chi} \frac{\chi^{D-1-j} \psidual \left( 1-\phidual^{[j+1]} \right) \sqrt{\rho^{[j+1]}}}{\left\| {\bf t}_{\changein}^{[j+1]\backstep} \right\|_\infty^2}}
            -
            \frac{\chi^{D-1-j} \psidual}{1-\chi^{D-1-j} \psidual} s^{[j+1]2} }
           {\mathop{\max}\limits_{i_{j+2}} \left\{ \frac{\frac{1}{H^{[j+1]}} \left\| {\bf W}_{\origin :i_{j+2}}^{[j+1]} \right\|_2^2}{
            \left\| {\bf W}^{[j+1]\backstep}_{\changein :i_{j+2}} \right\|_2^2 - \left\| {\bf W}^{[j+1]\backstep}_{\changein :i_{j+2}} \right\|_\infty^2
            } + \frac{H^{[j]}}{H^{[j+1]}} \right\} }
       \right)
  \end{array}
 \]
 $\forall j \in \infset{N}_{D-1}$, where $\phidual^{[D-1]} = \epsilon$.  If the weight-step satisfies:
 \[
  \begin{array}{rl}
   \left\| {\bf t}_{\changein}^{[j]\backstep} \right\|_{\infty}^2
   &< B^{[j]2} = 
   \left\{ \begin{array}{ll}
   \frac{\left( 1-\chi \right)^2 \left( 1 - \chi^{D-j-1} \psidual \right)}{\left( \frac{s^{[j]2}}{\left( 1-\phidual^{[j]} \right) \sqrt{{{\rho}}^{[j]}}} \right)}
   & \mbox{if } j < D-1 \\
   \frac{\left( 1-\chi \right)^2}{\left( \frac{s^{[D-1]2}}{\left( 1-\epsilon \right) \sqrt{{{\rho}}^{[D-1]}}} \right)}
   & \mbox{otherwise} \\
   \end{array} \right.
  \end{array}
 \]
 $\forall j \in \infset{N}_D$ then there exists of a canonical scaling:
 \[
  \begin{array}{l}
   \left. \frac{\partial}{\partial {\bf W}_\changein} \left\| {\bg{\Psi}}_\origin \left( {\bf W}_\changein \right) \right\|_{\mathcal{W}_\origin} \right|_{{\bf W}_\changein = {\bf W}_\changein^\backstep} 
   = 
   \bangrad {\bf W}_\changein^\backstep \\

   \bangrad
   =
   \frac{4}{\left\| {\bf t}_{\changein}^{[D-1]\backstep} \right\|_\infty^2}
   \frac{{\left( 1-\psidual \right)}{\left( 1-\chi \right)}}{\left( {\psidual-\chi} \right)^2} < \infty \\

  \end{array}
 \]
 where $\| {\bg{\Phi}}_\origin ({\bf x}) \|_F^2 \leq H^{[D-1]} \overline{\sigma}^{[D-1]} 
 ( ( 1-\epsilon ) \sqrt{{\rho}^{[D-1]}} )$ $\forall {\bf x} \in \mathbb{X}$ and $\| {{\bg{\Psi}}}_\origin 
 ({\bf W}_\changein) \|_F^2 \leq H^{[D-1]} \frac{1-\psidual}{\psidual}$.  
 \label{th:canexistgen_main}
\end{th_canexistgen_main}

\begin{figure}
\begin{centering}
\includegraphics[width=0.7\columnwidth]{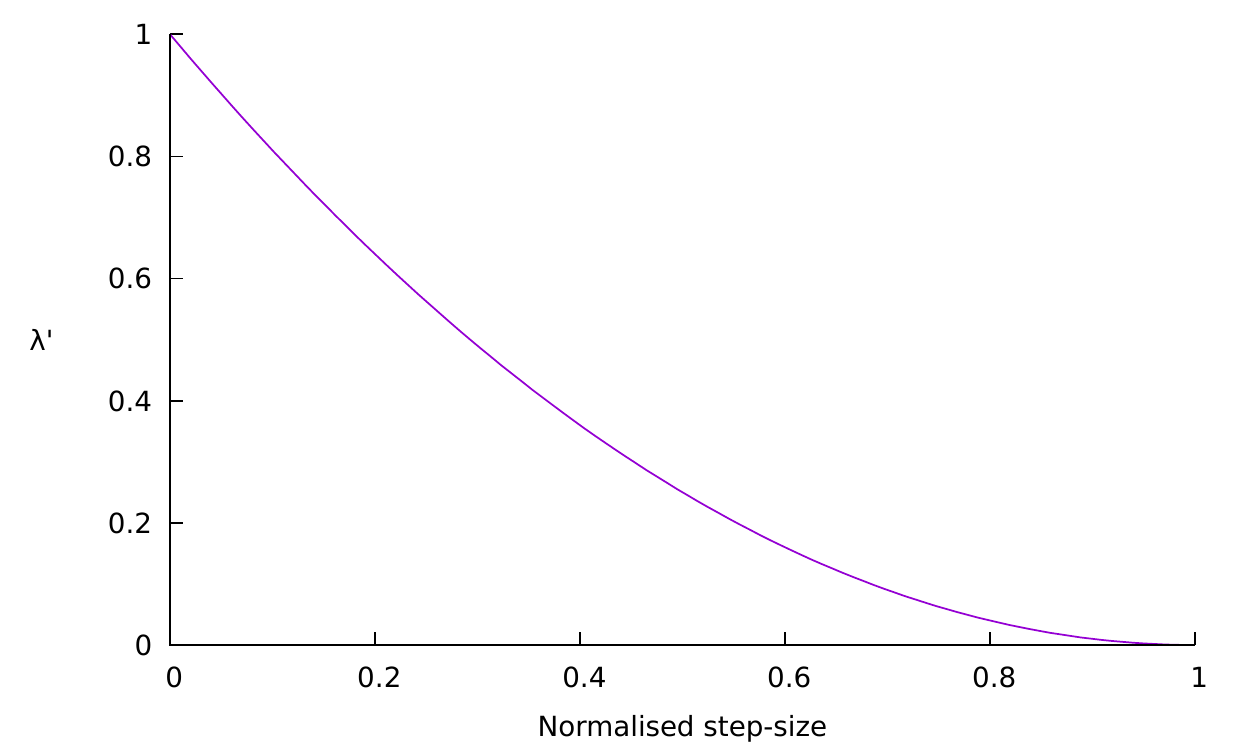} \\
\end{centering}
\caption{Normalised Canonical trade-off coefficient $\lambda' = 
         \frac{4\learnrate}{B^{[D-1]2}} \lambda$ vs 
         normalised gradient-descent 
         step-size $\frac{\| {\bf t}_{\changein}^{[j]\backstep} 
         \|_{\infty}}{B^{[D-1]}}$.}
\label{fig:regterm}
\end{figure}

This is proven as corollary \ref{cor:canexistgen_simple} in the appendix.  This 
theorem tells us that, for any set of initial weights ${\bf W}_\origin$, for a 
sufficiently small weight-step ${\bf W}_\changein^{\backstep}$ generated by back 
propogation, there exists a canonical scaling - i.e. a set of scaling factors and 
shadow weights such that the gradient-descent weight-step is exactly equivalent 
to the step generated by regularised RKBS learning using an appropriate trade-off 
parameter $\lambda$.  Note that:
\begin{itemize}
 \item The maximum step-size $B^{[j]}$ in layer $j$ (up to near-identity scaling terms) is 
       determined by the inverse fanout $1/s^{[j]}$ (which scales roughly as 
       $H^{[j]}/H^{[j-1]}$) and the scaled radius of convergence $(1-\phidual^{[j]}) 
       \sqrt{\rho^{[j]}}$ (which scales roughly inverse the the weight-step in 
       subsequent layers).  This bound will tend to get smaller as we move from the 
       output layer back toward the input, but so too will the weight-steps in many 
       cases due to the problem of vanishing gradients. 
 \item The trade-off coefficient (degree of regularisation) required by this 
       canonical scaling is:
       \[
        \begin{array}{l}
         \lambda
         =
         \frac{1}{\learnrate \bangrad}
         =
         \frac{B^{[D-1]2}}{4\learnrate} \left( 1 - \left( {\frac{\left\| {\bf t}_{\changein}^{[j]\backstep} \right\|_{\infty}}{B^{[D-1]}}} \right)^2 \right)^2 \\
        \end{array}
       \]
       as shown in figure \ref{fig:regterm}.  Note that (a) the degree of 
       regularisation required to generate an equivalent RKBS weight-step 
       is inversely proportional to the learning rate used to generate the 
       original back-propogation weight-step and (b) larger gradient-descent 
       weight-steps are equivalent to less regularised RKBS weight-steps, as 
       might be expected.  
\end{itemize}

\section{Application - Rademacher Complexity}  \label{sec:radcomp}

Having established an exact representation of neural networks in finite 
neighbourhoods of weights and biases, established the link with RKBS theory 
and demonstrated that a gradient descent step is equivalent to a regularised 
step in RKBS by appropriate, a-posterior selection of scale factors and 
shadow weights (canonical scaling), we now consider an application of this 
framework to uniform convergence analysis using Rademacher complexity.  The 
Rademacher complexity of a set $\mathcal{G}$ of real-valued functions is a 
measure of its capacity.  Assuming training vectors ${\bf x}_i \sim \nu$ 
and Rademacher random variables $\epsilon_i \in \{ -1,1 \}$, the Rademacher 
complexity of $\mathcal{G}$ is \citep{Men2}:
\[
 \begin{array}{rl}
  \mathcal{R}_{N} \left( \mathcal{G} \right) 
  &\!\!\!\!= \mathbb{E}_{\nu,\epsilon} \left[ {\rm sup}_{f \in \mathcal{G}} \left| \frac{1}{N} {\sum}_i \epsilon_i f \left( {\bf x}_i \right) \right|  \right] \\
 \end{array}
\]
This may be used in uniform convergence analysis to bound how quickly the 
empirical risk 
converges to the expected risk, typically of the form:
\[
 \begin{array}{l}
  \left| \overline{R} \left( g \right) - R_E \left( g \right) \right| \leq c \mathcal{R}_N \left( \mathcal{G} \right) + \mbox{excess risk}
 \end{array}
\]
The following theorem demonstrates how our framework may be used to bound 
Rademacher complexity for a scalar-output neural network:
\begin{th_onerad_main}
 Let $\epsilon, \chi \in (0,1)$ and for a given neural 
 network with initial weights ${\bf W}_\origin$, and let ${\bf 
 W}_\changein^\backstep$ be the weight-step for this derived from 
 back-propagation satisfying the conditions set out in corollary 
 \ref{cor:canexistgen_simple}.  Then $f_\changein^\backstep \in \mathcal{F}^\regstep$, where the 
 Rademacher complexity of $\mathcal{F}^\regstep$ is bounded as:
\[
 \begin{array}{l}
  \frac{\mathcal{R}_N \left( \mathcal{F}^\regstep \right)}{H^{[D-1]}}
  \leq
  \sqrt{\frac{\bar{\sigma}^{[D-1]} \left( \left( 1-\epsilon \right) \sqrt{\rho^{[D-1]}} \right)}{N} {\frac{\left\| {\bf t}_{\changein}^{[D-1]\backstep} \right\|_\infty^2}
   {\frac{1}{1-\chi} B^{[D-1]2} - \left\| {\bf t}_{\changein}^{[D-1]\backstep} \right\|_\infty^2} }} 
 \end{array}
\]
 \label{th:onerad_main}
\end{th_onerad_main}
The proof of this theorem is a straightforward application of theorem 
\ref{th:canexistgen_main} (see proof of theorem \ref{th:onerad} in the 
appendix for details).  Apart from the usual $\frac{1}{N}$ scaling this 
theorem is very different from typical bounds on Rademacher complexity, 
as it directly bounds the complexity using the size of the weight-step 
for a single iteration of gradient descent.  Note:
\begin{itemize}
 \item By the properties of Rademacher complexity \citep[Theorem 12]{Bar1} 
       the complexity of the trained neural network after $T$ iterations 
       may be bounded by a simple summation of the bounds on each step.  
       This supports the practice of using early-stopping to prevent 
       overfitting in neural networks.\footnote{We suspect that a more 
       careful analysis accounting for the overlap between the RKBSs for 
       each step may give sublinear dependence on $T$, but this is beyond 
       the scope of the present work.}
 \item The size of the back-propogation weight-step scales propotionally 
       with the learning rate, so for sufficiently small learning rates 
       we find:
       \[
        \begin{array}{l}
         \frac{\mathcal{R}_N \left( \mathcal{F}^\regstep \right)}{H^{[D-1]}}
         \lessapprox
         \sqrt{\frac{\bar{\sigma}^{[D-1]} \left( \left( 1-\epsilon \right) \sqrt{\rho^{[D-1]}} \right)}{N}
          } \left\| {\bf t}_{\changein}^{[D-1]\backstep} \right\|_\infty
        \end{array}
       \]
       However as the learning rate increases the bound will increase 
       at an accelerating rate.\footnote{Eventually the bound will become 
       meaningless as the step-size exceeds the size we can model using 
       an RKBS.}  This supports the lower learning rates act as a form of 
       regularisation in neural network training.
\end{itemize}

To finish we consider a concrete example.  Assume a $2$-layer network with a 
scalar output ($D = 2$, $n = H^{[1]} = 1$) with activation functions 
$\tau^{[0]} = \tau^{[1]} ={\rm tanh}$ (so $\rho^{[0]} = \rho^{[1]} = 
\frac{\pi}{2}$ and $\tau^{(1)[0]} (\zeta), \tau^{(1)[1]} (\zeta) \in [-1,1]$.  
The form of $\sigma_{z,z'}^{[j]}$ is non-trivial (see appendix 
\ref{sec:tanhcase} for details), but it is not difficult to see that 
$\overline{\sigma}^{[j]} =\sigma_{0,0}^{[j]}$, so, using the power-series 
about $0$:
\[
 \begin{array}{rl}
  \overline{\sigma}^{[0]} \left( \zeta \right) = \overline{\sigma}^{[1]} \left( \zeta \right) = \mathop{\sum}\limits_{m=1}^\infty \left( \frac{2^{2m} \left( 2^{2m} - 1 \right) B_{2m}}{(2m)!} \right)^2 \zeta^{2m-1} \\
 \end{array}
\]
and moreover $s^{[0]2} = \alpha^{[0]2}/2 + {m}/2{H^{[0]}}$ 
and $s^{[1]2} = \alpha^{[1]2}/2 + H^{[0]}$.  Using our assumptions 
and the back-propagation equations, it is not difficult to see that:
\[
 \begin{array}{rl}
  \left\| {\bf t}_{\changein}^{[0]\backstep} \right\|_\infty^2
  &\!\!\!\!\leq
  2 \learnrate^2 N^2 L_E^2 \left\| {\bf W}_{\origin}^{[1]} \right\|_F^2 \left( 1 + \alpha^{[0]2} \right) \\

  \left\| {\bf t}_{\changein}^{[1]\backstep} \right\|_\infty^2
  &\!\!\!\!\leq 
  2 \learnrate^2 N^2 L_E^2 \left( 1 + \alpha^{[1]2} \right) \\
 \end{array}
\]
and 
  $B^{[1]} = \frac{\left( 1-\chi \right)^2 \left( 1-\epsilon \right) \sqrt{\frac{\pi}{2}}}{\frac{1}{2} \alpha^{[1]2} + H^{[0]}}$.  
Assuming a learning rate:
\[
 \begin{array}{rl}
  \learnrate
  = s
  {\frac{\left( 1-\chi \right)}{N L_E \sqrt{2 \left( 1 + \alpha^{[1]2} \right)}} 
  \sqrt{\frac{\left( 1-\epsilon \right) \sqrt{\frac{\pi}{2}}}{\frac{1}{2} \alpha^{[1]2} + H^{[0]}}}}
 \end{array}
\]
where $0 < s \ll 1$ - which follows the practice of scaling the learning 
rate inversely with the batch size and the network width - we obtain the 
following Rademacher complexity bound for the neural network trained for 
$T$ iterations:
\[
 \begin{array}{l}
  \frac{\mathcal{R}_N \left( \mathcal{F}^\regstep \right)}{H^{[D-1]}}
   \lessapprox
  sT \sqrt{2 \frac{\bar{\sigma}^{[D-1]} \left( \left( 1-\epsilon \right) \sqrt{\frac{\pi}{2}} \right)}{N} }
  \frac{\left( 1-\chi \right)^2 \left( 1-\epsilon \right) \sqrt{\frac{\pi}{2}}}{\frac{1}{2} \alpha^{[1]2} + H^{[0]}}
  \end{array}
 \]
which we note scales inversely with the both the number of training vectors 
$N$ {\em and} the width of the network $H^{[0]}$.

\section{Conclusions and Future Directions}

In this paper we have established a connection between neural network training 
using gradient descent and regularised learning in reproducing kernel Banach 
space.  We have introduced an exact representation of the behaviour of neural 
networks as the weights and biases are varied in a finite neighbourhood of 
some initial weights and biases in terms of an inner product of two feature 
maps, one from data space to feature space, the other from weight-step space 
to feature space.  Using this, we showed that the change in neural network 
behaviour due to a single iteration of back-propagation lies in a reproducing 
kernel Banach space, and moreover that the weight-step found by 
back-propagation can be exactly replicated through regularised risk 
minimisation in RKBS.  Subsequently we presented an upper bound on the 
Rademacher complexity of neural networks applicable to both the over- and 
under-parametrised regimes, and discussed how this bound depends on learning 
rate, dataset size, network width and the number of training iterations used.

With regard to future work we foresee a number of useful directions.  First, 
the analysis should be extended to non-smooth activation functions such as 
ReLU, presumably by modifying the feature map using a representation other 
than a power series expansion.  Second, the precise influence of the learning 
rate needs to be further explicated, along with other details of Theorem 
\ref{th:canexistgen_main}.  With regard to the neural neighbourhood 
kernels themselves, it would be helpful if these could be reduced to closed 
form to allow them to be used in practice.\footnote{A Taylor approximation 
the NNKs is readily obtained, but it is difficult to say how accurate 
this may be without further analysis.}  Finally, more work is needed to understand 
the impact of the depth of the network on this theory.

\bibliography{universal}

\begin{thebibliography}{62}
\providecommand{\natexlab}[1]{#1}
\providecommand{\url}[1]{\texttt{#1}}
\expandafter\ifx\csname urlstyle\endcsname\relax
  \providecommand{\doi}[1]{doi: #1}\else
  \providecommand{\doi}{doi: \begingroup \urlstyle{rm}\Url}\fi

\bibitem[Allen-Zhu et~al.(2018)Allen-Zhu, Li, and Liang]{Zey1}
Z.~Allen-Zhu, Y.~Li, and Y.~Liang.
\newblock Learning and generalization in overparameterized neural networks,
  going beyond two layers.
\newblock \emph{{arXiv preprint arXiv:1811.04918}}, 2018.

\bibitem[Allen-Zhu et~al.(2019)Allen-Zhu, Li, and Song]{All3}
Z.~Allen-Zhu, Y.~Li, and Z.~Song.
\newblock A convergence theory for deep learning via over-parameterization.
\newblock In \emph{International Conference on Machine Learning}, pages
  242--252. PMLR, 2019.

\bibitem[Aronszajn(1950)]{Aro1}
N.~Aronszajn.
\newblock Theory of reproducing kernels.
\newblock \emph{Transactions of the American Mathematical Society},
  68:\penalty0 337--404, Jan--Jun 1950.

\bibitem[Arora et~al.(2018)Arora, Ge, Neyshabur, and Zhang]{Aro3}
S.~Arora, R.~Ge, B.~Neyshabur, and Y.~Zhang.
\newblock Stronger generalization bounds for deep nets via a compression
  approach.
\newblock In \emph{Proceedings of {ICML}}, 2018.

\bibitem[Arora et~al.(2019{\natexlab{a}})Arora, Du, Hu, Li, and Wang]{Aro6}
S.~Arora, S.~Du, W.~Hu, Z.~Li, and R.~Wang.
\newblock Fine-grained analysis of optimization and generalization for
  overparameterized two-layer neural networks.
\newblock In \emph{International Conference on Machine Learning}, pages
  322--332. PMLR, 2019{\natexlab{a}}.

\bibitem[Arora et~al.(2019{\natexlab{b}})Arora, Du, Hu, Li, Salakhutdinov, and
  Wang]{Aro4}
S.~Arora, S.~S. Du, W.~Hu, Z.~Li, R.~R. Salakhutdinov, and R.~Wang.
\newblock On exact computation with an infinitely wide neural net.
\newblock In \emph{Advances in Neural Information Processing Systems}, pages
  8139--8148, 2019{\natexlab{b}}.

\bibitem[Bach(2017)]{Bac4}
F.~Bach.
\newblock On the equivalence between kernel quadrature rules and random feature
  expansions.
\newblock \emph{The Journal of Machine Learning Research}, 18\penalty0
  (1):\penalty0 714--751, 2017.

\bibitem[Bach(2014)]{Bac3}
F.~R. Bach.
\newblock Breaking the curse of dimensionality with convex neural networks.
\newblock \emph{CoRR}, abs/1412.8690, 2014.
\newblock URL \url{http://arxiv.org/abs/1412.8690}.

\bibitem[Bai and Lee(2019)]{Bai2}
Y.~Bai and J.~D. Lee.
\newblock Beyond linearization: On quadratic and higher-order approximation of
  wide neural networks.
\newblock \emph{arXiv preprint arXiv:1910.01619}, 2019.

\bibitem[Bartlett and Mendelson(2002)]{Bar1}
P.~L. Bartlett and S.~Mendelson.
\newblock Rademacher and gaussian complexities: Risk bounds and structural
  results.
\newblock \emph{Journal of Machine Learning Research}, 3:\penalty0 463--482,
  2002.

\bibitem[Bartlett et~al.(2017)Bartlett, Foster, and Telgarsky]{Bar8}
P.~L. Bartlett, D.~J. Foster, and M.~J. Telgarsky.
\newblock Spectrally-normalized margin bounds for neural networks.
\newblock In \emph{Advances in Neural Information Processing Systems}, pages
  6240--6249, 2017.

\bibitem[Bartolucci et~al.(2021)Bartolucci, De~Vito, Rosasco, and
  Vigogna]{Bar9}
F.~Bartolucci, E.~De~Vito, L.~Rosasco, and S.~Vigogna.
\newblock Understanding neural networks with reproducing kernel banach spaces.
\newblock \emph{arXiv preprint arXiv:2109.09710}, 2021.

\bibitem[Cao and Gu(2019)]{Cao4}
Y.~Cao and Q.~Gu.
\newblock Generalization bounds of stochastic gradient descent for wide and
  deep neural networks.
\newblock In \emph{Advances in neural information processing systems},
  volume~32, 2019.

\bibitem[Cho and Saul(2009)]{Cho8}
Y.~Cho and L.~K. Saul.
\newblock Kernel methods for deep learning.
\newblock In B.~Y., S.~D., L.~J. D., C.~K.~I. Williams, and A.~Culotta,
  editors, \emph{Advances in Neural Information Processing Systems 22}, pages
  342--350. Curran Associates, Inc., 2009.
\newblock URL
  \url{http://papers.nips.cc/paper/3628-kernel-methods-for-deep-learning.pdf}.

\bibitem[Chowdhury and Gopalan(2017)]{Cho7}
S.~R. Chowdhury and A.~Gopalan.
\newblock On kernelized multi-armed bandits.
\newblock In D.~Precup and Y.~W. Teh, editors, \emph{Proceedings of the 34th
  International Conference on Machine Learning}, volume~70 of \emph{Proceedings
  of Machine Learning Research}, pages 844--853, International Convention
  Centre, Sydney, Australia, Aug 2017. PMLR.

\bibitem[Cortes and Vapnik(1995)]{Cor1}
C.~Cortes and V.~Vapnik.
\newblock Support vector networks.
\newblock \emph{Machine Learning}, 20\penalty0 (3):\penalty0 273--297, 1995.

\bibitem[Cristianini and Shawe-Taylor(2005)]{Cri4}
N.~Cristianini and J.~Shawe-Taylor.
\newblock \emph{An Introductino to Support Vector Machines and other
  Kernel-Based Learning Methods}.
\newblock Cambridge University Press, Cambridge, UK, 2005.

\bibitem[Daniely(2017)]{Dan2}
A.~Daniely.
\newblock Sgd learns the conjugate kernel class of the network.
\newblock In I.~Guyon, U.~V. Luxburg, S.~Bengio, H.~Wallach, R.~Fergus,
  S.~Vishwanathan, and R.~Garnett, editors, \emph{Advances in Neural
  Information Processing Systems 30}, pages 2422--2430. Curran Associates,
  Inc., 2017.
\newblock URL
  \url{http://papers.nips.cc/paper/6836-sgd-learns-the-conjugate-kernel-newline
  class-of-the-network.pdf}.

\bibitem[Daniely et~al.(2016)Daniely, Frostig, and Singer]{Dan1}
A.~Daniely, R.~Frostig, and Y.~Singer.
\newblock Toward deeper understanding of neural networks: The power of
  initialization and a dual view on expressivity.
\newblock In D.~D. Lee, M.~Sugiyama, U.~V. Luxburg, I.~Guyon, and R.~Garnett,
  editors, \emph{Advances in Neural Information Processing Systems 29}, pages
  2253--2261. Curran Associates, Inc., 2016.
\newblock URL
  \url{http://papers.nips.cc/paper/6427-toward-deeper-understanding-of-neural-networks-the-power-of-initialization-and-a-dual-view-on-expressivity.pdf}.

\bibitem[D`Aurizio(2014)]{TanhExpand}
J.~D`Aurizio.
\newblock Taylor series expansion of tanh x.
\newblock Mathematics Stack Exchange
  \url{https://math.stackexchange.com/q/1052926} (version: 2014-12-05), 2014.

\bibitem[Der and Lee(2007)]{Der1}
R.~Der and D.~Lee.
\newblock Large-margin classification in banach spaces.
\newblock In \emph{Proceedings of the {JMLR} Workshop and Conference 2:
  {AISTATS2007}}, pages 91--98, 2007.

\bibitem[Dr{\"a}xler et~al.(2018)Dr{\"a}xler, Veschgini, Salmhofer, and
  Hamprecht]{Dra2}
F.~Dr{\"a}xler, K.~Veschgini, M.~Salmhofer, and F.~A. Hamprecht.
\newblock Essentially no barriers in neural network energy landscape.
\newblock In \emph{Proceedings of the 35th International Conference on Machine
  Learning, {ICML 2018}}, 2018.

\bibitem[Du et~al.(2019{\natexlab{a}})Du, Lee, Li, Wang, and Zhai]{Du2}
S.~Du, J.~Lee, H.~Li, L.~Wang, and X.~Zhai.
\newblock Gradient descent finds global minima of deep neural networks.
\newblock In \emph{International conference on machine learning}, pages
  1675--1685. PMLR, 2019{\natexlab{a}}.

\bibitem[Du et~al.(2019{\natexlab{b}})Du, Zhai, Poczos, and Singh]{Du1}
S.~S. Du, X.~Zhai, B.~Poczos, and A.~Singh.
\newblock Gradient descent provably optimizes over-parameterized neural
  networks.
\newblock In \emph{Conference on Learning Representations}, 2019{\natexlab{b}}.

\bibitem[Genton(2001)]{Gen2}
M.~G. Genton.
\newblock Classes of kernels for machine learning: A statistics perspective.
\newblock \emph{Journal of Machine Learning Research}, 2:\penalty0 299--312,
  2001.

\bibitem[Golowich et~al.(2018)Golowich, Rakhlin, and Shamir]{Gol3}
N.~Golowich, A.~Rakhlin, and O.~Shamir.
\newblock Size-independent sample complexity of neural networks.
\newblock In \emph{{COLT}}, 2018.

\bibitem[G{\"o}nen and Alpaydin(2011)]{Gon7}
M.~G{\"o}nen and E.~Alpaydin.
\newblock Multiple kernel learning algorithms.
\newblock \emph{Journal of Machine Learning Research}, 12:\penalty0 2211--2268,
  2011.

\bibitem[Goodfellow et~al.(2016)Goodfellow, Bengio, and Courville]{Goo1}
I.~Goodfellow, Y.~Bengio, and A.~Courville.
\newblock \emph{Deep Learning}.
\newblock {MIT} Press, 2016.
\newblock \url{http://www.deeplearningbook.org}.

\bibitem[Gradshteyn and Ryzhik(2000)]{Gra1}
I.~S. Gradshteyn and I.~M. Ryzhik.
\newblock \emph{Table of Integrals, Series, and Products}.
\newblock Academic Press, London, 2000.

\bibitem[Harvey et~al.(2017)Harvey, Liaw, and Mehrabian]{Har2}
N.~Harvey, C.~Liaw, and A.~Mehrabian.
\newblock Nearly-tight {VC}-dimension bounds for piecewise linear neural
  networks.
\newblock In \emph{Proceedings of the {30th} Conference on Learning Theory,
  {COLT 2017}}, 2017.

\bibitem[Herbrich(2002)]{Her2}
R.~Herbrich.
\newblock \emph{Learning Kernel Classifiers: Theory and Algorithms}.
\newblock {MIT} Press, 2002.

\bibitem[Jacot et~al.(2018)Jacot, Gabriel, and Hongler]{Jac2}
A.~Jacot, F.~Gabriel, and C.~Hongler.
\newblock Neural tangent kernel: Convergence and generalization in neural
  networks.
\newblock In \emph{Advances in neural information processing systems}, pages
  8571--8580, 2018.

\bibitem[Knapp(2009)]{Kna1}
M.~P. Knapp.
\newblock Sines and cosines of angles in arithmetic progression.
\newblock \emph{Mathematics magazine}, 82\penalty0 (5):\penalty0 371, 2009.

\bibitem[Lee et~al.(2018)Lee, Sohl-dickstein, Pennington, Novak, Schoenholz,
  and Bahri]{Lee8}
J.~Lee, J.~Sohl-dickstein, J.~Pennington, R.~Novak, S.~Schoenholz, and
  Y.~Bahri.
\newblock Deep neural networks as gaussian processes.
\newblock In \emph{In International Conference on Learning Representations},
  2018.

\bibitem[Li et~al.(2017)Li, Venturi, and Xu]{Li7}
C.~Li, L.~Venturi, and R.~Xu.
\newblock Learning the kernel for classification and regression.
\newblock \emph{{arXiv preprint arXiv:1712.08597}}, 2017.

\bibitem[Li and Liang(2018)]{Li8}
Y.~Li and Y.~Liang.
\newblock Learning overparameterized neural networks via stochastic gradient
  descent on structured data.
\newblock In \emph{Advances in Neural Information Processing Systems 31: Annual
  Conference on Neural Information Processing Systems}, 2018.

\bibitem[Lin et~al.(2022)Lin, Zhang, and Zhang]{Lin10}
R.~Lin, H.~Zhang, and J.~Zhang.
\newblock On reproducing kernel banach spaces: Generic definitions and unified
  framework of constructions.
\newblock \emph{Acta Mathematica Sinica, English Series}, 2022.

\bibitem[Matthews et~al.(2018)Matthews, Rowland, Hron, Turner, and
  Ghahramani]{Mat5}
A.~G.~d.~G. Matthews, M.~Rowland, J.~Hron, R.~E. Turner, and Z.~Ghahramani.
\newblock Gaussian process behaviour in wide deep neural networks.
\newblock \emph{{arXiv e-prints}}, 2018.

\bibitem[Mendelson(2003)]{Men2}
S.~Mendelson.
\newblock A few notes on statistical learning theory.
\newblock In S.~Mendelson and A.~J. Smola, editors, \emph{Advanced Lectures on
  Machine Learning: Machine Learning Summer School 2002 Canberra, Australia,
  February 11--22, 2002 Revised Lectures}, pages 1--40. Springer Berlin
  Heidelberg, Berlin, Heidelberg, 2003.

\bibitem[M{\"u}ller et~al.(2001)M{\"u}ller, Mika, R{\"a}tsch, Tsuda, and
  Sch{\"o}lkopf]{Mul2}
K.-R. M{\"u}ller, S.~Mika, G.~R{\"a}tsch, K.~Tsuda, and B.~Sch{\"o}lkopf.
\newblock An introduction to kernel-based learning algorithms.
\newblock \emph{{IEEE} Transactions on Neural Networks}, 12\penalty0
  (2):\penalty0 181--198, March 2001.

\bibitem[Nagarajan and Kolter(2019{\natexlab{a}})]{Nag1}
V.~Nagarajan and J.~Z. Kolter.
\newblock Uniform convergence may be unable to explain generalization in deep
  learning.
\newblock In H.~Wallach, H.~Larochelle, A.~Beygelzimer, F.~d\'{e} Buc, E.~Fox,
  and R.~Garnett, editors, \emph{Advances in Neural Information Processing
  Systems 32}, pages 11615--11626. Curran Associates, Inc., 2019{\natexlab{a}}.
\newblock URL
  \url{http://papers.nips.cc/paper/9336-uniform-convergence-may-be-unable-to-explain-generalization-in-deep-learning.pdf}.

\bibitem[Nagarajan and Kolter(2019{\natexlab{b}})]{Nag2}
V.~Nagarajan and Z.~Kolter.
\newblock Deterministic {PAC-Bayesian} generalization bounds for deep networks
  via generalizing noise-resilience.
\newblock In \emph{International Conference on Learning Representations
  {(ICLR)}}, 2019{\natexlab{b}}.

\bibitem[Neal(1996)]{Nea1}
R.~M. Neal.
\newblock \emph{Priors for infinite networks}, pages 29--53.
\newblock Springer, 1996.

\bibitem[Neyshabur et~al.(2015)Neyshabur, Tomioka, and Srebro]{Ney1}
B.~Neyshabur, R.~Tomioka, and N.~Srebro.
\newblock Norm-based capacity control in neural networks.
\newblock In \emph{Proceedings of Conference on Learning Theory}, pages
  1376--1401, 2015.

\bibitem[Neyshabur et~al.(2017)Neyshabur, Bhojanapalli, McAllester, and
  Srebro]{Ney4}
B.~Neyshabur, S.~Bhojanapalli, D.~McAllester, and N.~Srebro.
\newblock Exploring generalization in deep learning.
\newblock In \emph{Proceedings of the 31st International Conference on Neural
  Information Processing Systems}, pages 5949--5958, 2017.

\bibitem[Neyshabur et~al.(2018)Neyshabur, Bhojanapalli, and Srebro]{Ney2}
B.~Neyshabur, S.~Bhojanapalli, and N.~Srebro.
\newblock A {PAC}-bayesian approach to spectrally-normalized margin bounds for
  neural networks.
\newblock In \emph{Proceedings of {ICLR}}, 2018.

\bibitem[Neyshabur et~al.(2019)Neyshabur, Li, Bhojanapalli, LeCun, and
  Srebro]{Ney3}
B.~Neyshabur, Z.~Li, S.~Bhojanapalli, Y.~LeCun, and N.~Srebro.
\newblock The role of over-parametrization in generalization of neural
  networks.
\newblock In \emph{Proceedings of {ICLR}}, 2019.

\bibitem[Parhi and Nowak(2021)]{Par2}
R.~Parhi and R.~D. Nowak.
\newblock Banach space representer theorems for neural networks and ridge
  splines.
\newblock \emph{J. Mach. Learn. Res.}, 22\penalty0 (43):\penalty0 1--40, 2021.

\bibitem[Rahimi and Benjamin(2009)]{Rah2}
A.~Rahimi and R.~Benjamin.
\newblock Weighted sums of random kitchen sinks: Replacing minimization with
  randomization in learning.
\newblock In D.~Koller, D.~Schuurmans, Y.~Bengio, and L.~Bottou, editors,
  \emph{Advances in Neural Information Processing Systems 21}, pages
  1313--1320. Curran Associates, Inc., 2009.

\bibitem[Sanders(2020)]{San9}
K.~Sanders.
\newblock Neural networks as functions parameterized by measures: Representer
  theorems and approximation benefits.
\newblock Master's thesis, Eindhoven University of Technology, 2020.

\bibitem[Shawe-Taylor and Cristianini(2004)]{Sha3}
J.~Shawe-Taylor and N.~Cristianini.
\newblock \emph{Kernel Methods for Pattern Analysis}.
\newblock Cambridge University Press, 2004.

\bibitem[Smola and Sch{\"o}lkopf(1998)]{Smo12}
A.~J. Smola and B.~Sch{\"o}lkopf.
\newblock On a kernel-based method for pattern recognition, regression,
  approximation and operator inversion.
\newblock \emph{Algorithmica}, 22:\penalty0 211--231, 1998.
\newblock Technical Report 1064, {GMD} First, April 1997.

\bibitem[Song et~al.(2013)Song, Zhang, and Hickernell]{Son1}
G.~Song, H.~Zhang, and F.~J. Hickernell.
\newblock Reproducing kernel banach spaces with the $\ell^1$ norm.
\newblock \emph{Applied and Computational Harmonic Analysis}, 34\penalty0
  (1):\penalty0 96--116, Jan 2013.

\bibitem[Sriperumbudur et~al.(2011)Sriperumbudur, Fukumizu, and
  Lanckriet]{Sri3}
B.~K. Sriperumbudur, K.~Fukumizu, and G.~R. Lanckriet.
\newblock Learning in hilbert vs. banach spaces: A measure embedding viewpoint.
\newblock In \emph{Advances in Neural Information Processing Systems}, pages
  1773--1781, 2011.

\bibitem[Steinwart and Christman(2008)]{Ste3}
I.~Steinwart and A.~Christman.
\newblock \emph{Support Vector Machines}.
\newblock Springer, 2008.

\bibitem[Unser(2021)]{Uns1}
M.~Unser.
\newblock A unifying representer theorem for inverse problems and machine
  learning.
\newblock \emph{Foundations of Computational Mathematics}, 21\penalty0
  (4):\penalty0 941--960, 2021.

\bibitem[Xu and Ye(2014)]{Xu4}
Y.~Xu and Q.~Ye.
\newblock Generalized mercer kernels and reproducing kernel banach spaces.
\newblock \emph{{arXiv preprint arXiv:1412.8663}}, 2014.

\bibitem[Zhang and Zhang(2012)]{Zha14}
H.~Zhang and J.~Zhang.
\newblock Regularized learning in banach spaces as an optimization problem:
  representer theorems.
\newblock \emph{Journal of Global Optimization}, 54\penalty0 (2):\penalty0
  235--250, Oct 2012.

\bibitem[Zhang et~al.(2009)Zhang, Xu, and Zhang]{Zha11}
H.~Zhang, Y.~Xu, and J.~Zhang.
\newblock Reproducing kernel banach spaces for machine learning.
\newblock \emph{Journal of Machine Learning Research}, 10:\penalty0 2741--2775,
  2009.

\bibitem[Zhou et~al.(2019)Zhou, Veitch, Austern, Adams, and Orbanz]{Zho5}
W.~Zhou, V.~Veitch, M.~Austern, R.~P. Adams, and P.~Orbanz.
\newblock Nonvacuous generalization bounds at the imagenet scale: a
  {PAC-Bayesian} compression approach.
\newblock In \emph{International Conference on Learning Representations
  {(ICLR)}}, 2019.

\bibitem[Zou and Gu(2019)]{Zou2}
D.~Zou and Q.~Gu.
\newblock An improved analysis of training over-parameterized deep neural
  networks.
\newblock In \emph{Advances in neural information processing systems},
  volume~32, 2019.

\bibitem[Zou et~al.(2020)Zou, Cao, Zhou, and Gu]{Zou1}
D.~Zou, Y.~Cao, D.~Zhou, and Q.~Gu.
\newblock Gradient descent optimizes over-parameterized deep relu networks.
\newblock \emph{Machine learning}, 109\penalty0 (3):\penalty0 467--492, 2020.

\end{thebibliography}
\bibliographystyle{abbrvnat}

%%%%%%%%%%%%%%%%%%%%%%%%%%%%%%%%%%%%%%%%%%%%%%%%%%%%%%%%%%%%%%%%%%%%%%%%%%%%%%%
%%%%%%%%%%%%%%%%%%%%%%%%%%%%%%%%%%%%%%%%%%%%%%%%%%%%%%%%%%%%%%%%%%%%%%%%%%%%%%%
% APPENDIX
%%%%%%%%%%%%%%%%%%%%%%%%%%%%%%%%%%%%%%%%%%%%%%%%%%%%%%%%%%%%%%%%%%%%%%%%%%%%%%%
%%%%%%%%%%%%%%%%%%%%%%%%%%%%%%%%%%%%%%%%%%%%%%%%%%%%%%%%%%%%%%%%%%%%%%%%%%%%%%%
\appendix
\newpage
\section{Derivations and Proofs.}

Our goal in this supplement is to present all derivations and proofs relevant 
to our paper, and also any additional material and description that may be 
useful.

We assume a fully-connected, $D$-layer feedforward neural network ${\bf f} : 
(\infset{X} \subseteq \infset{R}^n) \to (\infset{Y} \subseteq \infset{R}^m)$ 
with layers of widths $H^{[0]}, H^{[1]}, \ldots, H^{[D-1]}$, where $H^{[D-1]} 
= m$ and we define $H^{[-1]} = n$.  We assume layer $j \in \infset{N}_D$ (we 
use the convention $j \in \infset{N}_D$ throughout) is made up of neurons with 
the same activation function $\tau^{[j]} : \infset{R} \to \infset{R}$.  The 
network is defined recursively:
\begin{equation}
 \begin{array}{ll}
 \begin{array}{rl}
  {\bf f} \left( {\bf x} \right) 
  &\!\!\!\!=
  {\bf x}^{[D]} \in \infset{R}^{H^{[D-1]}} \\
 \end{array} & \\
 \begin{array}{rl}
  {\bf x}^{[j+1]} 
  &\!\!\!\!= \tau^{[j]} \left( \tilde{\bf x}^{[j]} \right) \in \infset{R}^{H^{[j]}} \\
  \tilde{\bf x}^{[j]} 
  &\!\!\!\!= \frac{1}{\sqrt{H^{[j]}}} {\bf W}^{[j]\tsp} {\bf x}^{[j]} + \alpha^{[j]} {\bf b}^{[j]} \in \infset{R}^{H^{[j]}} \\
 \end{array} & \;\;\; \forall j \in \infset{N}_D \\
 \begin{array}{rl}
  {\bf x}^{[0]} 
  &\!\!\!\!= {\bf x} \in \infset{X} \subset \infset{R}^{H^{[-1]}} \mbox{ ($H^{[-1]} = n$)} \\
 \end{array} & \\
 \end{array}
 \label{eq:yall}
\end{equation}
where ${\bf W}^{[j]} \in \infset{R}^{H^{[j-1]} \times H^{[j]}}$ and ${\bf 
b}^{[j]} \in \infset{R}^{H^{[j]}}$ are weights and biases, and $\alpha^{[j]} 
\in \infset{R}_+$ is a constant we will use later.  We define the set of 
neural networks taking the form (\ref{eq:yall}) as:
\begin{equation}
 \begin{array}{rl}
  \mathcal{F} = \left\{ \left. {\bf f} : \infset{R}^n \to \infset{R}^m \right| {\bf f} \mbox{ has form (\ref{eq:yall}) given } {\bf W} \in \infset{W} \right\}
 \end{array}
\end{equation}
where ${\bf W}$ summarises the weights and biases in (\ref{eq:yall}):
\[
 \begin{array}{rl}
  {\bf W} = \left( \left. \left( {\bf W}^{[j]}, {\bf b}^{[j]} \right) \right| j \in \infset{N}_D \right) \in \infset{W} = \mathop{\prod}\limits_{j \in \infset{N}_D} \left( \infset{R}^{H^{[j-1]} \times H^{[j]}} \times \infset{R}^{H^{[j]}} \right) \\
 \end{array}
\]

Typically, the goal in neural network training is to take a training set and 
find weights and biases to minimise some measure of mismatch between the true 
training labels and the network's predictions.  For simplicity let us assume 
we wish to minimise:
\begin{equation}
 \begin{array}{l}
  {\bf f}^\star = \mathop{\rm argmin}\limits_{{\bf f} \in \mathcal{F}} \sum_k E \left( {\bf x}^{\{k\}}, {\bf y}^{\{k\}}, {\bf f} \left( {\bf x}^{\{k\}} \right) \right)
 \end{array}
\label{eq:traingoal}
\end{equation}
or, equivalently:
\begin{equation}
 \begin{array}{l}
  {\bf W}^\star = \mathop{\rm argmin}\limits_{{\bf W} \in \infset{W}} \sum_k E \left( {\bf x}^{\{k\}}, {\bf y}^{\{k\}}, {\bf f}_{\bf W} \left( {\bf x}^{\{k\}} \right) \right)
 \end{array}
 \label{eq:traingoalW}
\end{equation}
where ${\bf f}_{\bf W}$ is a network of the form (\ref{eq:yall}) with weights 
and biases ${\bf W}$, $\infset{D} = \{ ({\bf x}^{\{k\}},{\bf y}^{\{k\}}) \in 
\infset{X} \times \infset{Y} : k \in \infset{N}_N \}$ is a training set (we 
use the convention $k \in \infset{N}_N$ throughout), and $E: \infset{X} \times 
\infset{Y} \times \infset{R}^m \to \infset{R}$ is an error function defining 
the purpose of the network.

We make the following assumptions:
\begin{enumerate}
 \item Input space: we assume $\infset{X} = [-{M}^{[-1]}, {M}^{[-1]}]^n$.
 \item Error function: we assume the error function $E: \infset{X} \times 
\infset{Y} \times \infset{R}^m \to \infset{R}$ is $\mathcal{C}^1$ and 
       $L_E$-Lipschitz in its third argument.
 \item Activation functions: we assume the activation functions $\tau^{[j]} : 
       \infset{R} \to [-{M}^{[j]}, {M}^{[j]}]^{H^{[j]}}$ are bounded and 
       $\mathcal{C}^\infty$, and that $\tau^{[j]}$ has a power-series 
       representations with region of convergence (ROC) at least $\rho^{[j]} 
       \in \infset{R}_+$ around $z$ for all $z \in \infset{R}$, $j \in 
       \infset{N}_D$.
 \item Weight non-triviality: we assume ${\bf W}^{[j]} \ne {\bf 0}$ for all $j 
       \in \infset{N}_D$ at all times during training.\footnote{Note that 
       networks that do not meet this requirement have a constant output 
       independent of input ${\bf x}$.  We do not consider this a restrictive 
       assumption as it is highly unlikely that a randomly initialised network 
       trained with a typical training set will ever reach this state.}
 \item Weight initialisation: we assume that initially $W_{i_j,i_{j+1}}^{[j]}, 
       b_{i_{j+1}}^{[j]} \sim \mathcal{N} (0,1)$ (LeCun initialisation).
 \item Training: we assume the network is trained using the back-propagation 
       with learning rate $\learnrate \in \infset{R}_+$.
\end{enumerate}

\subsection{Review of Back-Propogation} \label{sec:backprop}

We now give a brief review of back-propagation training, which is a systematic 
implementation of gradient descent on the weights and biases ${\bf W}$ to 
solve (\ref{eq:traingoalW}).  Let us consider a single training iteration, 
where we start with initial weights and biases ${\bf W}_\origin\in \infset{W}$ 
and calculate a weight-step ${\bf W}_\changein^\backstep$ so that, after this 
iteration, ${\bf W} = {\bf W}_\origin + {\bf W}_\changein^\backstep$, where:
\[
 \begin{array}{rl}
  {\bf W}_\changein^\backstep
  &\!\!\!\!=
  -\learnrate
  \left. \frac{\partial}{\partial {\bf W}} \sum_k E \left( {\bf x}^{\{k\}}, {\bf y}^{\{k\}}, {\bf f}_{\bf W} \left( {\bf x}^{\{k\}} \right) \right) \right|_{{\bf W} = {\bf W}_\origin}
 \end{array}
\]
Denote the network activation prior to the iteration given input ${\bf x}$ as:
\begin{equation}
 \begin{array}{ll}
 \begin{array}{rl}
  {\bf f}_\origin \left( {\bf x} \right) 
  &\!\!\!\!=
  {\bf x}_\origin^{[D]} \\
 \end{array} & \\
 \begin{array}{rl}
  {\bf x}_\origin^{[j+1]} 
  &\!\!\!\!= \tau^{[j]} \left( \tilde{\bf x}_\origin^{[j]} \right)  \\
  \tilde{\bf x}_\origin^{[j]} 
  &\!\!\!\!= \frac{1}{\sqrt{H^{[j]}}} {\bf W}_\origin^{[j]\tsp} {\bf x}_\origin^{[j]} + \alpha^{[j]} {\bf b}_\origin^{[j]} \\
 \end{array} & \\ 
 \begin{array}{rl}
  {\bf x}_\origin^{[0]} 
  &\!\!\!\!= {\bf x} \in \infset{X} \\
 \end{array} & \\
 \end{array}
 \label{eq:yallorigin}
\end{equation}
for all layers $j \in \infset{N}_D$.  The back-propagation iteration for layer 
$D-1$ is a gradient descent step with learning rate $\learnrate$, namely:
\[
 \begin{array}{rl}
  W_{\changein i_{D-1},i_{D}}^{[D-1]\backstep} 
  &\!\!\!\!= \left.
  -\learnrate \frac{\partial}{\partial W_{i_{D-1},i_{D}}^{[D-1]}} \sum_k E \left( {\bf x}^{\{k\}}, {\bf y}^{\{k\}}, {\bf f}_{\bf W} \left( {\bf x}^{\{k\}} \right) \right) 
  \right| {\bf W} = {\bf W}_\origin \\

  &\!\!\!\!= \left.
  -\learnrate \sum_{k,i^\prime_D} \nabla_{i^\prime_D} E \left( {\bf x}^{\{k\}}, {\bf y}^{\{k\}}, {\bf f}_{\bf W} \left( {\bf x}^{\{k\}} \right) \right) 
  \frac{\partial}{\partial W_{i_{D-1},i_{D}}^{[D-1]}} f_{i^\prime_D} \left( {\bf x}^{\{k\}} \right) 
  \right| {\bf W} = {\bf W}_\origin \\

  &\!\!\!\!= \left.
  -\learnrate \sum_{k,i^\prime_D} \nabla_{i^\prime_D} E \left( {\bf x}^{\{k\}}, {\bf y}^{\{k\}}, {\bf f}_{\bf W} \left( {\bf x}^{\{k\}} \right) \right) 
  \frac{\partial}{\partial W_{i_{D-1},i_{D}}^{[D-1]}} x_{i'_D}^{\{k\}[D]} 
  \right| {\bf W} = {\bf W}_\origin \\

  &\!\!\!\!= \left.
  -\learnrate \sum_{k,i^\prime_D} \nabla_{i^\prime_D} E \left( {\bf x}^{\{k\}}, {\bf y}^{\{k\}}, {\bf f}_{\bf W} \left( {\bf x}^{\{k\}} \right) \right) 
  \frac{\partial}{\partial W_{i_{D-1},i_{D}}^{[D-1]}} \tau^{[D-1]} \left( {\tilde{x}}_{i'_D}^{\{k\}[D-1]} \right) 
  \right| {\bf W} = {\bf W}_\origin \\

  &\!\!\!\!= \left.
  -\learnrate \sum_{k,i^\prime_D} \nabla_{i^\prime_D} E \left( {\bf x}^{\{k\}}, {\bf y}^{\{k\}}, {\bf f}_{\bf W} \left( {\bf x}^{\{k\}} \right) \right) 
  \tau^{[D-1](1)} \left( {\tilde{x}}_{i'_D}^{\{k\}[D-1]} \right)
  \frac{\partial}{\partial W_{i_{D-1},i_{D}}^{[D-1]}} {\tilde{x}}_{i'_D}^{\{k\}[D-1]}
  \right| {\bf W} = {\bf W}_\origin \\

  &\!\!\!\!= 
  -\learnrate \frac{1}{\sqrt{H^{[D-1]}}} \sum_{k} \gamma_{\origin i_D}^{\{k\}[D-1]} x_{\origin i_{D-1}}^{\{k\}[D-1]} \\
 \end{array}
\]
where $\tau^{(r)}(z) = \frac{\partial^r}{\partial^r z} \tau(z)$ and:
\[
 \begin{array}{rl}
  \gamma_{\origin i_D}^{\{k\}[D-1]}
  &\!\!\!\!=
   \nabla_{i_D} E \left( {\bf x}^{\{k\}}, {\bf y}^{\{k\}}, {\bf f}_{\bf W} \left( {\bf x}^{\{k\}} \right) \right) \tau^{[D-1](1)} \left( \tilde{x}_{\origin i_D}^{\{k\}[D-1]} \right) \\
 \end{array}
\]
Subsequently:
\[
 \begin{array}{rl}
  W_{\changein i_{D-2},i_{D-1}}^{[D-2]\backstep} 
  &\!\!\!\!= \left.
  -\learnrate \frac{\partial}{\partial W_{i_{D-2},i_{D-1}}^{[D-2]}} \sum_k E \left( {\bf x}^{\{k\}}, {\bf y}^{\{k\}}, {\bf f} \left( {\bf x}^{\{k\}} \right) \right) 
  \right| {\bf W} = {\bf W}_\origin \\

  &\!\!\!\!= \left.
  -\learnrate \sum_{k,i^\prime_D} \gamma_{\origin i^\prime_D}^{\{k\}[D-1]} \frac{\partial}{\partial W_{i_{D-2},i_{D-1}}^{[D-2]}} \tilde{x}_{i^\prime_D}^{\{k\}[D-1]} 
  \right| {\bf W} = {\bf W}_\origin \\

  &\!\!\!\!= \left.
  -\learnrate \frac{1}{\sqrt{H^{[D-1]}}} \sum_{k,i^\prime_D,i^\prime_{D-1}} \gamma_{\origin i^\prime_D}^{\{k\}[D-1]} W_{\origin i^\prime_{D-1},i^{\prime}_D}^{[D-1]} 
  \frac{\partial}{\partial W_{i_{D-2},i_{D-1}}^{[D-2]}} x_{i^\prime_{D-1}}^{\{k\}[D-1]} 
  \right| {\bf W} = {\bf W}_\origin \\

  &\!\!\!\!= \left.
  -\learnrate \frac{1}{\sqrt{H^{[D-1]}}} \sum_{k,i^\prime_D,i^\prime_{D-1}} \gamma_{\origin i^\prime_D}^{\{k\}[D-1]} W_{\origin i^\prime_{D-1},i^{\prime}_D}^{[D-1]} 
  \frac{\partial}{\partial W_{i_{D-2},i_{D-1}}^{[D-2]}} \tau^{[D-2]} \left( \tilde{x}_{i^\prime_{D-1}}^{\{k\}[D-2]} \right)
  \right| {\bf W} = {\bf W}_\origin \\

  &\!\!\!\!= \left.
  -\learnrate \frac{1}{\sqrt{H^{[D-1]}}} \sum_{k,i^\prime_D,i^\prime_{D-1}} \gamma_{\origin i^\prime_D}^{\{k\}[D-1]} W_{\origin i^\prime_{D-1},i^{\prime}_D}^{[D-1]} 
  \tau^{[D-2](1)} \left( \tilde{x}_{i^\prime_{D-1}}^{\{k\}[D-2]} \right) \frac{\partial}{\partial W_{i_{D-2},i_{D-1}}^{[D-2]}} \tilde{x}_{i^\prime_{D-1}}^{\{k\}[D-2]}
  \right| {\bf W} = {\bf W}_\origin \\

  &\!\!\!\!= \left.
  -\learnrate \frac{1}{\sqrt{H^{[D-1]}}} \sum_{k,i^\prime_{D-1}} \gamma_{\origin i^\prime_{D-1}}^{\{k\}[D-2]} 
  \frac{\partial}{\partial W_{i_{D-2},i_{D-1}}^{[D-2]}} \tilde{x}_{i^\prime_{D-1}}^{\{k\}[D-2]}
  \right| {\bf W} = {\bf W}_\origin \\

  &\!\!\!\!= 
  -\learnrate \frac{1}{\sqrt{H^{[D-1]}H^{[D-2]}}} \sum_{k} \gamma_{\origin i_{D-1}}^{\{k\}[D-2]} x_{\origin i_{D-2}}^{\{k\}[D-2]} \\
 \end{array}
\]
where:
\[
 \begin{array}{rl}
  \gamma_{\origin i_{D-1}}^{\{k\}[D-2]}
  &\!\!\!\!=
  \sum_{i_D} \gamma_{\origin i_D}^{\{k\}[D-1]} W_{\origin i_{D-1},i_D}^{\{k\}[D-1]} \tau^{[D-2](1)} \left( \tilde{x}_{\origin i_{D-1}}^{\{k\}[D-2]} \right) 
 \end{array}
\]
and so on through all layers.  Summarising, for all $j \in \infset{N}_D$ (and 
using the MATLAB notation ${\bf A}_{:i}$ for column $i$ of matrix ${\bf A}$):
\begin{equation}
 \begin{array}{rl}
   {\bf W}_{\changein :i_{j+1}}^{[j]\backstep}
  &\!\!\!\!= 
  -\learnrate \frac{1}{\sqrt{H^{[D-1]} H^{[D-2]} \ldots H^{[j+1]}}} \sum_{k} \gamma_{\origin i_{j+1}}^{\{k\}[j]} \frac{1}{\sqrt{H^{[j]}}} {\bf x}_{\origin}^{\{k\}[j]} \\

   b_{\changein :i_{j+1}}^{[j]\backstep}
  &\!\!\!\!= 
  -\learnrate \frac{1}{\sqrt{H^{[D-1]} H^{[D-2]} \ldots H^{[j+1]}}} \sum_{k} \gamma_{\origin i_{j+1}}^{\{k\}[j]} \alpha^{[j]} \\
 \end{array}
\label{eq:backproprep}
\end{equation}
where, recursively:
\begin{equation}
 \begin{array}{rl}
  \gamma_{\origin i_{j}}^{\{k\}[j-1]}
  &\!\!\!\!=
  \sum_{i_{j+1}} \gamma_{\origin i_{j+1}}^{\{k\}[j]} W_{\origin i_{j},i^{\prime}_{j+1}}^{[j]} \tau^{[j-1](1)} \left( \tilde{x}_{\origin i_{j}}^{\{k\}[j-1]} \right) \\

  \gamma_{\origin i_D}^{\{k\}[D-1]}
  &\!\!\!\!=
   \nabla_{i_D} E \left( {\bf x}^{\{k\}}, {\bf y}^{\{k\}}, {\bf f} \left( {\bf x}^{\{k\}} \right) \right) \tau^{[D-1](1)} \left( \tilde{x}_{\origin i_D}^{\{k\}[D-1]} \right) \\
 \end{array}
\label{eq:backpropgamma}
\end{equation}

\section{Dual Form of a Neural Network Step} \label{append:dualform}

Our first goal is to rewrite the neural network after a training iteration as:
\[
 \begin{array}{rl}
  {\bf f} \left( {\bf x} \right) 
  &\!\!\!\!= {\bf f}_\origin \left( {\bf x} \right) + {\bf f}_{\changein} \left( {\bf x} \right) \\
 \end{array}
\]
where ${\bf f}_\origin : ( \infset{X} \subset \infset{R}^n ) \to ( \infset{Y} 
\subset \infset{R}^m )$ is the neural network before the iteration and ${\bf 
f}_\changein : ( \infset{X} \subset \infset{R}^n ) \to \infset{R}^m$ is the 
change in network behaviour due to the change ${\bf W}_\changein \in 
\infset{W}_\origin$ in weights and biases for this iteration, so that:
\[
 \begin{array}{rl}
  {\bf f}_{\changein} \left( {\bf x} \right) 
  &\!\!\!\!= \left< 
             {\bg{\Phi}}_{\origin} \left( {\bf x}             \right),
             {\bg{\Psi}}_{\origin} \left( {\bf W}_{\changein} \right)
             \right>_{\mathcal{X}_\origin \times \mathcal{W}_\origin} \\
 \end{array}
\]
where:
\[
 \begin{array}{rl}
  {\bg{\Phi}}_{\origin} &\!\!\!\!: \infset{X}         \to \mathcal{X}_\origin = {\rm span} \left( {{\bg{\Phi}}}_\origin \left( \infset{X}         \right) \right) \subset \infset{R}^{\infty \times m} \\
  {\bg{\Psi}}_{\origin} &\!\!\!\!: \infset{W}_\origin \to \mathcal{W}_\origin = {\rm span} \left( {{\bg{\Psi}}}_\origin \left( \infset{W}_\origin \right) \right) \subset \infset{R}^{\infty \times m} \\
 \end{array}
\]
are feature maps;
\[
 \begin{array}{rl}
  \infset{W}_\origin \subset \mathop{\prod}\limits_{j \in \infset{N}_D} \left( \infset{R}^{H^{[j-1]} \times H^{[j]}} \times \infset{R}^{H^{[j]}} \right) \\
 \end{array}
\]
and $\left<\cdot,\cdot\right>_{\mathcal{X}_\origin \times \mathcal{W}_\origin} 
: \mathcal{X}_\origin \times \mathcal{W}_\origin \to \infset{R}^m$ is the 
bilinear form:
\[
 \begin{array}{rl}
  \left< {\bg{\Xi}}, {\bg{\Omega}} \right>_{\mathcal{X}_\origin \times \mathcal{W}_\origin} &\!\!\!\!= {\rm diag} \left( {\bg{\Xi}}^\tsp {\bg{\Omega}} \right)
 \end{array}
\]
Moreover we aim to show that the feature maps are entirely defined by:
\begin{enumerate}
 \item The structure of the network - that is, the number of layers, their 
       widths and activation functions.
 \item The weights and biases ${\bf W}_\origin \in \infset{W}$ before the 
       iteration.
\end{enumerate}
We will then use this to construct appropriate norms for the weight feature 
space $\mathcal{W}_\origin$ and data feature space $\mathcal{X}_\origin$, 
which will allow us to prove that ${\bf f}_{\changein}$ lies in a reproducing 
kernel Banach space, and analyse the complexity and convergence of the neural 
network.

\subsection{Preliminary: Power Series Notation}

The approach taken here is direct - we construct a power series expansion of 
the network without truncation and rearrange the terms to separate the 
resulting summation into the desired form.  To make the process easier we 
define:
\[

\]
noting that this is linear in the elementwise (Hadamard) product, so:
\[
  \begin{array}{rl}
   {\bg{\varrho}} \left( {\bf a}, {\bf d} \right) \odot {\bg{\varrho}} \left( {\bf a}', {\bf d}' \right) 
   &\!\!\!\!= 
   {\bg{\varrho}} \left( {\bf a} \odot {\bf a}', {\bf d} \odot {\bf d}' \right) \\
  \end{array}
\]
Using this notation it is straightforward to obtain the following useful 
shorthand for the multi-dimensional Taylor expansion (using the smoothness 
assumption) of $\tau^{[j]} : \infset{R} \to \infset{R}$ about $z \in 
\infset{R}$ for ${\bf c}, {\bf c}' \in \infset{R}^p$, $\left| \left< {\bf c}, 
{\bf c}' \right> \right| \leq \rho^{[j]}$:
\begin{equation}
 \begin{array}{rl}
  \tau^{[j]} \left( z + \left< {\bf c}, {\bf c}' \right> \right) 
  &\!\!\!\!= \tau^{[j]} \left( z \right) + 
  \left< {\bg{\varrho}} \left( {\bf g}^{[j]} \left( z \right), {\bf c} \right), 
         {\bg{\varrho}} \left( {\bf 1}_\infty, {\bf c}' \right), \right> \\
 \end{array}
 \label{eq:taylorbase}
\end{equation}
where the derivatives of $\tau^{[j]}$ at $z$ are encoded as:
\[
 %
\]
and $\tau^{[j](r)} (z) = \frac{\partial^r}{\partial z^r} \tau^{[j]} (z)$.

\subsection{Feature Maps for Neural Networks}

We represent the operation before and after the iteration, as well as the 
change due to the iteration, as:
\begin{equation}
 %
 \label{eq:fchange_form}
\end{equation}
Our goal is to write ${\bf f}_\changein : \infset{X} \to \infset{R}^m$ as:
\[
 \begin{array}{rl}
  {\bf f}_{\changein} \left( {\bf x} \right) 
  &\!\!\!\!= \left< 
             {\bg{\Phi}}_{\origin} \left( {\bf x}             \right),
             {\bg{\Psi}}_{\origin} \left( {\bf W}_{\changein} \right)
             \right>_{\mathcal{X}_\origin \times \mathcal{W}_\origin} \\
 \end{array}
\]
where ${\bg{\Phi}}_{\origin} : \infset{X} \to \mathcal{X}_\origin$ and 
${\bg{\Psi}}_{\origin} : \infset{W}_\origin \to \mathcal{W}_\origin$ are 
feature maps and $\left< {\bg{\Xi}},{\bg{\Omega}} \right>_{\mathcal{X}_\origin 
\times \mathcal{W}_\origin} = {\rm diag} ({\bg{\Xi}}^\tsp {\bg{\Omega}})$ is a 
bilinear form.

Starting with the first layer of the network, it is not difficult to see 
that:
\[
 %
\]
where $\shadx^{[0]}_{i_1} \in \infset{R}_+$ are parameters we call scale factors, 
and ${\bf A}_{:l}$ for column $l$ of matrix ${\bf A}$.  
The scale factors will always cancel out in our representation of ${\bf f}$, 
but will be important later when we place norms on feature space.  Using a 
power series expansion as per (\ref{eq:taylorbase}), if $\| \tilde{\bf 
x}_\changein^{[0]} \|_\infty \leq \rho^{[0]}$:
\[
 %
\]

Moving to the second layer, we see that:
\[
 %
\]
where $\shadx_{i_2}^{[1]} \in \infset{R}_+$ are scale factors and 
${\shadw}_{i_1}^{[1]}, {\shadwo}_{i_1,i_2}^{[1]} \in \infset{R}^+$ are shadow 
weights (note that both the shadow weights and scale factors cancel in the 
inner product).  Taking the Taylor expansion and using the properties of 
${\bg{\varrho}}$ we see that, provided $\| \tilde{\bf x}_{\changein}^{[1]} 
\|_\infty \leq \rho^{[1]}$:
\[
 {{
 \begin{array}{rl}
  {\bf x}_\changein^{[2]}
  &\!\!\!\!= \left[ \left< 
             {\bg{\Phi}}_{\origin :i_2}^{[1]} \left( {\bf x} \right),
             {\bg{\Psi}}_{\origin :i_2}^{[1]} \left( {\bf W}_\changein \right)
  \right> \right]_{i_2} \\
 \end{array}%
 }}
\]

Continuing, it is not difficult to see that, for any $j \in \infset{N}_D$, if:
\[
 {{
 \begin{array}{rl}
  {\bf x}_\changein^{[j]}
  &\!\!\!\!= \left[ \left< 
             {\bg{\Phi}}_{\origin :i_j}^{[j-1]} \left( {\bf x} \right),
             {\bg{\Psi}}_{\origin :i_j}^{[j-1]} \left( {\bf W}_\changein \right)
  \right> \right]_{i_{j}} \\
 \end{array}
 }}
\]
then, with scale factors $\shadx_{i_{j+1}}^{[j]} \in \infset{R}_+$ and 
shadow weights $\shadw_{i_{j}}^{[j]}, \shadwo_{i_j,i_{j+1}}^{[j]} \in \infset{R}_+$:
\[
 %
\]

Thus, by induction, so long as $\| \tilde{\bf x}_{\changein}^{[j]} \|_\infty \leq \rho^{[j]}$ $\forall j$:
\begin{equation}
 \begin{array}{rl}
  {\bf f} \left( {\bf x} \right) 
  &\!\!\!\!= {\bf f}_\origin \left( {\bf x} \right) + {\bf f}_{\changein} \left( {\bf x} \right) \\

  {\bf f}_{\changein} \left( {\bf x} \right) 
  &\!\!\!\!= \left< 
             {\bg{\Phi}}_{\origin} \left( {\bf x} \right),
             {\bg{\Psi}}_{\origin} \left( {\bf W}_\changein \right)
             \right>_{\mathcal{X}_\origin \times \mathcal{W}_\origin}
 \end{array}
 \label{eq:yfrombase}
\end{equation}
where:
\[
 {{
 \begin{array}{rl}
  {\bg{\Phi}}_{\origin} \left( {\bf x} \right)
  &\!\!\!\!=
  {\bg{\Phi}}_{\origin}^{[D-1]} \left( {\bf x} \right) \\

  {\bg{\Psi}}_{\origin} \left( {\bf W}_\changein \right)
  &\!\!\!\!=
  {\bg{\Psi}}_{\origin}^{[D-1]} \left( {\bf W}_\changein \right) \\
 \end{array}
 }}
\]
and, working recursively:
\begin{equation}
 {{
 %
 \right] 
             \right) &
             \mbox{if } j = 0 \\
             \end{array} \right. \\
 \end{array} & \;\;\; \forall j \in \infset{N}_D
 \end{array}
 }}
\label{eq:featmapdefs}
\end{equation}
with scale factors $\shadx_{i_{j+1}}^{[j]} \in \infset{R}_+$ and shadow 
weights $\shadw_{i_{j}}^{[j]}, \shadwo_{i_j,i_{j+1}}^{[j]} \in \infset{R}_+$, 
$\forall j \in \infset{N}_D$.

To summarise, we have decomposed the neural network into the desired form 
(\ref{eq:yfrombase}) with the following components:
\begin{itemize}
 \item A term ${\bf f}_\origin \left( {\bf x} \right)$ which is the network 
       prior to the iteration evaluated on ${\bf x}$.
 \item A feature map ${{\bg{\Phi}}}_{\origin} : \infset{X} \to 
       \mathcal{X}_\origin = {\rm span}({{\bg{\Phi}}}_\origin(\infset{X})) 
       \subset \infset{R}^{\infty \times m}$ from data input space to data 
       feature space that is dependent only on the structure of the network 
       (number of layers, their widths and the neuron types) and the weights 
       and biases ${\bf W}_\origin$ prior to the iteration.
 \item A feature map ${\bg{\Psi}}_{\origin} : \infset{W}_\origin \to 
       \mathcal{W}_\origin = {\rm span} ({\bg{\Psi}}_\origin 
       (\infset{W}_\origin)) \subset \infset{R}^{\infty \times m}$ from 
       weight-step input space to weight-step feature space that is dependent 
       only on the structure of the network (number of layers and their 
       widths).
\end{itemize}

\subsection{Density of Feature Maps}

In this section we prove a key property of the feature maps that will be 
required to prove that ${\bf f}_{\changein}$ lies in the reproducing kernel 
Banach space.  As a preliminary we show a consequence of the non-triviality 
assumption, namely:
\begin{lem_denseprekey}
 Let ${\bf f}_\origin \in \mathcal{F}$ be a neural network satisfying the 
 non-triviality assumption.  Then ${\bf 
 f}_\changein ({\bf x})$ varies non-trivially with ${\bf x}$ (that is, it 
 is non-constant).
 \label{lem:denseprekey}
\end{lem_denseprekey}
\begin{proof}
Recall that:
\[
 \begin{array}{rl}
  \tilde{\bf x}_\changein^{[j]}
  &\!\!\!\!= \frac{1}{\sqrt{H^{[j]}}} {\bf W}_\changein^{[j]\tsp} {\bf x}_\origin^{[j]} + 
             \frac{1}{\sqrt{H^{[j]}}} \left( {\bf W}_\origin^{[j]} + {\bf W}_\changein^{[j]} \right)^\tsp {\bf x}_\changein^{[j]} + 
             \alpha^{[j]} {\bf b}_\changein^{[j]} \\

  &\!\!\!\!= \frac{1}{\sqrt{H^{[j]}}} {\bf W}_\changein^{[j]\tsp} \left( {\bf x}_\origin^{[j]} + {\bf x}_\changein^{[j]} \right) + 
             \frac{1}{\sqrt{H^{[j]}}} {\bf W}_\origin^{[j]\tsp} {\bf x}_\changein^{[j]} + 
             \alpha^{[j]} {\bf b}_\changein^{[j]} \\
 \end{array}
\]
we want to know the conditions under which $\tilde{\bf x}_\changein^{[D-1]}$ 
(and hence ${\bf f}_{\changein} ({\bf x})$) is a constant, independent of 
${\bf x}$.  Considering instead $\tilde{\bf x}_\changein^{[j]}$, the only 
component in the above expression that depends on ${\bf x}$ is ${\bf 
x}_\changein^{[j]}$, so $\tilde{\bf x}_\changein^{[j]}$ is constant if 
either ${\bf W}_\changein^{[j]} = -{\bf W}_\origin^{[j]}$ or ${\bf 
x}_\changein^{[j]}$ is constant independent of ${\bf x}$.  So, recursing, we 
find that $\tilde{\bf x}_\changein^{[D-1]}$ is constant (independent of ${\bf 
x}$) iff ${\bf W}_\changein^{[j]} = -{\bf W}_\origin^{[j]}$ for some $j \in 
\infset{N}_D$.  However, recall that our non-triviality assumption explicitly 
rules out this case (which corresponds to trivial neural network post 
iteration), so ${\bf W}_\changein^{[j]} \ne -{\bf W}_\origin^{[j]}$ for all $j 
\in \infset{N}_D$, which suffices to demonstrate the desired result.
\end{proof}
Having addressed this preliminary, we now move to the main result for this 
section:\footnote{See \citep{Lin10} for discussion of density used here.}
\begin{lem_densekey}[Density of Feature Maps]
 The linear span of ${{\bg{\Phi}}}_\origin (\infset{X})$ is dense in 
 $\mathcal{X}_\origin$ with respect to $\left< \cdot, \cdot 
 \right>_{\mathcal{X}_\origin \times \mathcal{W}_\origin}$; and the linear 
 span of ${\bg{\Psi}}_\origin (\infset{W}_\origin)$ is dense in 
 $\mathcal{W}_\origin$ with respect to $\left< \cdot, \cdot 
 \right>_{\mathcal{W}_\origin \times \mathcal{X}_\origin}$, where $\left< 
 {\bg{\Omega}}, {\bg{\Xi}} \right>_{\mathcal{W}_\origin \times 
 \mathcal{X}_\origin} = \left< {\bg{\Xi}}, {\bg{\Omega}} 
 \right>_{\mathcal{X}_\origin \times \mathcal{W}_\origin}^\tsp$.
 \label{lem:densekey}
\end{lem_densekey}
\begin{proof}
By definition \citep{Lin10}, the linear span ${\rm span} 
({{\bg{\Phi}}}_\origin (\infset{X}))$ of ${{\bg{\Phi}}}_\origin 
(\infset{X})$ is dense in $\mathcal{X}_\origin$ with respect to 
$\left< \cdot, \cdot \right>_{\mathcal{X}_\origin \times \mathcal{W}_\origin}$ 
if for any ${\bg{\Omega}}_\changein \in \mathcal{W}_\origin$, the statement:
\[
 \begin{array}{l}
  \left< {\bg{\Phi}}_\origin \left( {\bf x} \right), {\bg{\Omega}}_\changein \right>_{\mathcal{X}_\origin \times \mathcal{W}_\origin} = {\bf 0} \;\;\; \forall {\bf x} \in \infset{X}
 \end{array}
\]
implies that ${\bg{\Omega}}_\changein = {\bf 0}$.  By definition 
$\mathcal{W}_\origin = {\rm span} ({{\bg{\Psi}}}_\origin( 
\infset{W}_\origin))$ so that ${\bg{\Omega}}_\changein = \sum_l \tilde{\alpha}^{\{l\}} {\bf 
W}_\changein^{\{l\}}$ for some $\tilde{\alpha}^{\{0\}}, \tilde{\alpha}^{\{1\}}, \ldots \in 
\infset{R}$, ${\bf W}_{\changein}^{\{0\}}, {\bf W}_{\changein}^{\{1\}}, \ldots \in 
\infset{W}_\origin$, and hence:
\[
 \begin{array}{l}
  \left< {\bg{\Phi}}_\origin \left( {\bf x} \right), {\bg{\Omega}}_\changein \right>_{\mathcal{X}_\origin \times \mathcal{W}_\origin} 
  = 
  \sum_l \tilde{\alpha}^{\{l\}} 
  \left< {\bg{\Phi}}_\origin \left( {\bf x} \right), {\bg{\Psi}}_\origin \left( {\bf W}_{\changein}^{\{l\}} \right) \right>_{\mathcal{X}_\origin \times \mathcal{W}_\origin} 
  = 
  \sum_l \tilde{\alpha}^{\{l\}} 
  {\bf f}_{\changein}^{\{l\}} \left( {\bf x} \right)
 \end{array}
\]
where we denote by ${\bf f}_{\changein}^{\{l\}}$ the change in ${\bf f}$ due to 
weight-step ${\bf W}_{\changein}^{\{l\}}$.  By our preliminary this is ${\bf 0}$ 
for all ${\bf x} \in \infset{X}$ iff $\tilde{\alpha}^{\{0\}} = \tilde{\alpha}^{\{l\}} = \ldots = 
0$, so ${\bg{\Omega}}_\changein = {\bf 0}$, which proves the required density 
property.  Similarly, ${\rm span} ({{\bg{\Psi}}}_\origin 
(\infset{W}_\origin))$ is dense in $\mathcal{W}_\origin$ with respect to 
$\left< \cdot, \cdot \right>_{\mathcal{W}_\origin \times \mathcal{X}_\origin}$ 
if for any ${\bg{\Xi}}_\changein \in \mathcal{X}_\origin$, the observation 
that:
\[
 \begin{array}{l}
  \left< {\bg{\Xi}}_\changein, {\bg{\Psi}}_\origin \left( {\bf W}_\changein \right) \right>_{\mathcal{X}_\origin \times \mathcal{W}_\origin} = {\bf 0} \;\;\; \forall {\bf W}_\changein \in \infset{W}
 \end{array}
\]
implies ${\bg{\Xi}}_\changein = {\bf 0}$.  By definition $\mathcal{X}_\origin 
= {\rm span} ({{\bg{\Phi}}}_\origin (\infset{X}))$ so that ${\bg{\Xi}}_\changein 
= \sum_l \tilde{\beta}^{\{l\}} {\bf x}^{\{l\}}$ for some $\tilde{\beta}^{\{0\}}, 
\tilde{\beta}^{\{1\}}, \ldots \in \infset{R}$, ${\bf x}^{\{0\}}, {\bf x}^{\{1\}}, 
\ldots \in \infset{X}$, and hence:
\[
 \begin{array}{l}
  \left< {\bg{\Xi}}_\changein, {\bg{\Psi}}_\origin \left( {\bf W}_\changein \right) \right>_{\mathcal{X}_\origin \times \mathcal{W}_\origin} 
  = 
  \sum_l \tilde{\beta}^{\{l\}} 
  \left< {\bg{\Phi}}_\origin \left( {\bf x}^{\{l\}} \right), {\bg{\Psi}}_\origin \left( {\bf W}_{\changein} \right) \right>_{\mathcal{X}_\origin \times \mathcal{W}_\origin} 
  = 
  \sum_l \tilde{\beta}^{\{l\}} 
  {\bf f}_{\changein} \left( {\bf x}^{\{l\}} \right)
 \end{array}
\]
But again, by our preliminary, this is ${\bf 0}$ for all ${\bf W}_\changein 
\in \infset{W}_\origin$ iff $\tilde{\beta}^{\{0\}} = \tilde{\beta}^{\{1\}} = 
\ldots = 0$, so ${\bg{\Xi}}_\changein = {\bf 0}$, completing the proof.
\end{proof}

\begin{rem_nontrivalid}
 An objection may be raised that the non-triviality assumption is 
 arbitrary and in any case not guaranteed to hold, bringing the above 
 result into doubt.  However, considering that neural networks are usually 
 initialised with non-trivial weights and biases (or, to be precise, the 
 chance that any one layer has all zero weights and biases drawn from e,g. a 
 uniform or normal distribution is $0$), and that training data is usually 
 non-trivial itself (that is, the targets vary and not just the inputs, so 
 that the trivial case has no special significance in terms of potential 
 optimality giving a fixed output), it seems highly unlikely that the 
 weights and biases would return to a trivial (all weights and biases $0$ in 
 at least one layer) state over any finite number of back-propagation 
 iterations - indeed we conjecture that a result saying that the probability 
 of encountering trivial networks weights and biases during training is 
 precisely $0$ given random weight initialisation and noise-affected data drawn 
 from a distribution should exist, but unfortunately we have been unable to 
 obtain a proof, so this remains speculative for now.
 \label{rem:nontrivalid}
\end{rem_nontrivalid}

\section{Induced Kernels and Induced Norms} \label{sec:inducedkernelsetc}

In the previous section we established that, as a result of a single iteration 
(step-change in weights and biases), the operation of the neural network can 
be written:
\[
 \begin{array}{rl}
  {\bf f} \left( {\bf x} \right) 
  &\!\!\!\!= {\bf f}_\origin \left( {\bf x} \right) + {\bf f}_{\changein} \left( {\bf x} \right) \\
 \end{array}
\]
where ${\bf f}_\origin : \infset{X} \to \infset{Y}$ is the neural network 
before the iteration and ${\bf f}_{\changein}:\infset{X}\to\infset{Y}_\origin$ 
is the change in network behaviour due to the change ${\bf W}_\changein \in 
\infset{W}_\origin$ in weights and biases for this iteration.  Moreover we 
showed that:
\[
 \begin{array}{rl}
  {\bf f}_{\changein} \left( {\bf x} \right) 
  &\!\!\!\!= \left< 
             {\bg{\Phi}}_{\origin} \left( {\bf x}             \right),
             {\bg{\Psi}}_{\origin} \left( {\bf W}_{\changein} \right)
             \right>_{\mathcal{X}_\origin \times \mathcal{W}_\origin} \\
 \end{array}
\]
where:
\[
 \begin{array}{rllll}
  {\bg{\Phi}}_{\origin} &\!\!\!\!: \infset{X}         &\!\!\!\!\to \mathcal{X}_\origin &\!\!\!\!= {\rm span} \left( {{\bg{\Phi}}}_\origin \left( \infset{X}         \right) \right) &\!\!\!\!\subset \infset{R}^{\infty \times m} \\
  {\bg{\Psi}}_{\origin} &\!\!\!\!: \infset{W}_\origin &\!\!\!\!\to \mathcal{W}_\origin &\!\!\!\!= {\rm span} \left( {{\bg{\Psi}}}_\origin \left( \infset{W}_\origin \right) \right) &\!\!\!\!\subset \infset{R}^{\infty \times m} \\
 \end{array}
\]
are feature maps, and $\left< \cdot, \cdot \right>_{\mathcal{X}_\origin \times 
\mathcal{W}_\origin} : \mathcal{X}_\origin \times \mathcal{W}_\origin \to 
\infset{R}^m$ is the bilinear form:
\[
 \begin{array}{rl}
  \left< {\bg{\Xi}}, {\bg{\Omega}} \right>_{\mathcal{X}_\origin \times \mathcal{W}_\origin} &\!\!\!\!= {\rm diag} \left( {\bg{\Xi}}^\tsp {\bg{\Omega}} \right)
 \end{array}
\]
In this section we will place (induce) kernels and norms on 
$\infset{X}_\origin$ and $\infset{W}_\origin$, which will in turn allow us 
to constrain the space of changes in neural network behaviour:
\[
 \begin{array}{l}
  \mathcal{F}_\origin = \left\{ \left. {\bf f}_\changein : \infset{X} \to \infset{Y}_\origin \right| {\bf f}_\changein \mbox{ as above} \right\}
 \end{array}
\]
using H{\"o}lder's inequality:
\[
 \begin{array}{l}
  \left\| {\bf f}_\changein \left( {\bf x} \right) \right\|_2 \leq 
  \left\| {{\bg{\Phi}}}_{\origin} \left( {\bf x} \right) \right\|_{\mathcal{X}_\origin} \left\| {{\bg{\Psi}}}_{\origin} \left( {\bf W}_\changein \right) \right\|_{\mathcal{W}_\origin}
 \end{array}
\]

\subsection{Induced Kernels}

We begin by using the feature maps and the kernel trick to induce kernels on 
$\infset{X}$ and $\infset{W}_\origin$, specifically:
\[
 \begin{array}{rl}
  {{\bf K}}_{\mathcal{X}_\origin} \left( {\bf x}, {\bf x}' \right) 
  &\!\!\!\!= 
  \left< {{\bg{\Phi}}}_{\origin} \left( {\bf x}           \right), {{\bg{\Phi}}}_{\origin} \left( {\bf x}'           \right) \right>_{\mathcal{X}_\origin \times \mathcal{X}_\origin} \\

  {{\bf K}}_{\mathcal{W}_\origin} \left( {\bf W}_\changein, {\bf W}'_\changein \right) 
  &\!\!\!\!= 
  \left< {{\bg{\Psi}}}_{\origin} \left( {\bf W}_\changein \right), {{\bg{\Psi}}}_{\origin} \left( {\bf W}'_\changein \right) \right>_{\mathcal{W}_\origin \times \mathcal{W}_\origin} \\
 \end{array}
\]
where we define the bilinear forms:
\[
 \begin{array}{rl}
  \left< {\bg{\Xi}},    {\bg{\Xi}}'    \right>_{\mathcal{X}_\origin \times \mathcal{X}_\origin} = {\bg{\Xi}}^\tsp    {\bg{\Xi}}'    \\
  \left< {\bg{\Omega}}, {\bg{\Omega}}' \right>_{\mathcal{W}_\origin \times \mathcal{W}_\origin} = {\bg{\Omega}}^\tsp {\bg{\Omega}}' \\
 \end{array}
\]
The matrix-valued kernels ${\bf K}_{\mathcal{X}_\origin}$ and ${\bf 
K}_{\mathcal{W}_\origin}$ are positive-definite (Mercer) by construction.  
Apart from their theoretical use, these could potentially be used (transfered) 
for support vector machines (SVMs), Gaussian Processes (GP) or similar 
kernel-based methods, measuring similarity on $\infset{X}$ and 
$\infset{W}_\origin$, respectively.  

For all layers $j \in \infset{N}_D$ we define:
\[
 \begin{array}{rl}
  {{\bf K}}_{\mathcal{X}_\origin}^{[j]} \left( {\bf x}, {\bf x}' \right) 
  &\!\!\!\!= 
  \left< {{\bg{\Phi}}}_{\origin}^{[j]} \left( {\bf x}           \right), {{\bg{\Phi}}}_{\origin}^{[j]} \left( {\bf x}'           \right) \right>_{\mathcal{X}_\origin \times \mathcal{X}_\origin} \\

  {{\bf K}}_{\mathcal{W}_\origin}^{[j]} \left( {\bf W}_\changein, {\bf W}'_\changein \right) 
  &\!\!\!\!= 
  \left< {{\bg{\Psi}}}_{\origin}^{[j]} \left( {\bf W}_\changein \right), {{\bg{\Psi}}}_{\origin}^{[j]} \left( {\bf W}'_\changein \right) \right>_{\mathcal{W}_\origin \times \mathcal{W}_\origin} \\
 \end{array}
\]
Recalling (\ref{eq:featmapdefs}) we see that, for all $j \in \infset{N}_D 
\backslash \{ 0 \}$:
\[
 \begin{array}{rl}
  K_{\mathcal{X}_\origin i_{j+1},i'_{j+1}}^{[j]} \left( {\bf x}, {\bf x}' \right) 
  &\!\!\!\!= \mathop{\sum}\limits_l
             \varrho_l \Bigg( {\bf g}^{[j]} \left( {\tilde{x}}_{\origin i_{j+1}}^{[j]} \right) \odot {\bf g}^{[j]} \left( {\tilde{x}}_{\origin i'_{j+1}}^{\prime[j]} \right), \ldots \\ & \;\;\;\;\;\;\;\;\;\;\;\;\;\;\;\;\; \ldots  \left.
             {{{\shadx}_{i_{j+1}}^{[j]} {\shadx}_{i'_{j+1}}^{[j]}}} \left[ \begin{array}{c}
                    \left[ \begin{array}{c} \frac{1}{2} \alpha^{[j]2} \\ \frac{1}{{H^{[j]}}} {\bf x}_\origin^{[j]} \odot {\bf x}_\origin^{[j]} \\ \end{array} \right] \\
                    \left[ \begin{array}{c} \frac{{\shadw_{i_j}^{[j]2}} {W_{\origin i_j, i_{j+1}}^{[j]} W_{\origin i_j, i'_{j+1}}^{[j]}}}{{\shadwo}_{i_j,i_{j+1}}^{[j]} {\shadwo}_{i_j,i'_{j+1}}^{[j]} H^{[j]}} {\bg{\Phi}}_{\origin:i_j}^{[j-1]} \left( {\bf x} \right) \odot {\bg{\Phi}}_{\origin:i_j}^{[j-1]} \left( {\bf x}' \right) \\ \end{array} \right]_{i_j} \\
                                                                                   \left[ \begin{array}{c} \frac{{\shadw_{i_j}^{[j]2}}                                                                 }{H^{[j]}} {\bg{\Phi}}_{\origin:i_j}^{[j-1]} \left( {\bf x} \right) \odot {\bg{\Phi}}_{\origin:i_j}^{[j-1]} \left( {\bf x}' \right) \\ \end{array} \right]_{i_j} \\
             \end{array} \right]
             \right) \\

  K_{\mathcal{W}_\origin i_{j+1},i'_{j+1}}^{[j]} \left( {\bf W}_\changein, {\bf W}'_\changein \right) 
  &\!\!\!\!= \mathop{\sum}\limits_l
             \varrho_l \left( {\bf 1}_\infty,
             \frac{1}{{{\shadx}_{i_{j+1}}^{[j]} {\shadx}_{i'_{j+1}}^{[j]}}} \left[ \begin{array}{c}
                    \left[ \begin{array}{c} 2 b_{\changein i_{j+1}}^{[j]} b_{\changein i'_{j+1}}^{\prime [j]} \\ {\bf W}_{\changein :i_{j+1}}^{[j]} \odot {\bf W}_{\changein :i'_{j+1}}^{\prime [j]} \\ \end{array} \right]  \\
                    \left[ \begin{array}{c}  \frac{{\shadwo}_{i_j,i_{j+1}}^{[j]} {\shadwo}_{i_j,i'_{j+1}}^{[j]}}{\shadw_{i_j}^{[j]2}} {\bg{\Psi}}_{\origin :i_j}^{[j-1]} \left( {\bf W}_\changein \right) \odot {\bg{\Psi}}_{\origin :i_j}^{[j-1]} \left( {\bf W}'_\changein \right) \\ \end{array} \right]_{i_j} \\
                                                                         \left[ \begin{array}{c} \frac{W_{\changein i_j, i_{j+1}}^{[j]} W_{\changein i_j, i'_{j+1}}^{\prime [j]}}{\shadw_{i_j}^{[j]2}} {\bg{\Psi}}_{\origin :i_j}^{[j-1]} \left( {\bf W}_\changein \right) \odot {\bg{\Psi}}_{\origin :i_j}^{[j-1]} \left( {\bf W}'_\changein \right) \\ \end{array} \right]_{i_j} \\
             \end{array} \right]
             \right) \\
 \end{array}
\]
and:
\[
 \begin{array}{l}
 \begin{array}{rl}
  K_{\mathcal{X}_\origin i_{1},i'_{1}}^{[0]} \left( {\bf x}, {\bf x}' \right) 
  &\!\!\!\!= \mathop{\sum}\limits_l
             \varrho_l \left( {\bf g}^{[0]} \left( {\tilde{x}}_{\origin i_{1}}^{[0]} \right) \odot {\bf g}^{[0]} \left( {\tilde{x}}_{\origin i'_{1}}^{\prime[0]} \right), 
             {{{\shadx}_{i_{1}}^{[0]} {\shadx}_{i'_{1}}^{[0]}}} \left[ \begin{array}{c}
                    \frac{1}{2} \alpha^{[0]2} \\ 
                    \frac{1}{{H^{[0]}}} \frac{1}{2} {\bf x}_\origin^{[0]} \odot {\bf x}_\origin^{[0]} \\
             \end{array} \right]
             \right) \\

  K_{\mathcal{W}_\origin i_{1},i'_{1}}^{[0]} \left( {\bf W}_\changein, {\bf W}'_\changein \right) 
  &\!\!\!\!= \mathop{\sum}\limits_l
             \varrho_l \left( {\bf 1}_\infty,
             \frac{1}{{{\shadx}_{i_{1}}^{[0]} {\shadx}_{i'_{1}}^{[0]}}} \left[ \begin{array}{c}
                    2 b_{\changein i_{1}}^{[0]} b_{\changein i'_{1}}^{\prime [0]} \\ 
                    2 {\bf W}_{\changein :i_{1}}^{[0]} \odot {\bf W}_{\changein :i'_{1}}^{\prime [0]} \\
             \end{array} \right]
             \right) \\
 \end{array}
 \end{array}
\]
Subsequently $\forall j \in \infset{N}_D \backslash \{0\}$:
\[
{\!\!\!\!\!\!\!\!\!\!\!\!{
 \begin{array}{rl}
  K_{\mathcal{X}_\origin i_{j+1},i'_{j+1}}^{[j]} \left( {\bf x}, {\bf x}' \right) 
  &\!\!\!\!= 
   \mathop{\sum}\limits_{l=1}^\infty \left( \frac{1}{l!} \right)^2 \tau^{[j](l)} \left( {\tilde{x}}_{\origin i_{j+1}}^{[j]} \right) \tau^{[j](l)} \left( {\tilde{x}}_{\origin i'_{j+1}}^{\prime[j]} \right) 
             \Bigg( {{{\shadx}_{i_{j+1}}^{[j]} {\shadx}_{i'_{j+1}}^{[j]}}} \Bigg(
                    \left( \frac{1}{2} \alpha^{[j]2} + \frac{1}{H^{[j]}} \left< {\bf x}_{\origin}^{[j]}, {\bf x}_{\origin}^{\prime[j]} \right> \right)
                    + \ldots \\ 
                    & \;\;\;\; \ldots \mathop{\sum}\limits_{i_j} \frac{1}{{\shadwo}_{i_j,i_{j+1}}^{[j]} {\shadwo}_{i_j,i'_{j+1}}^{[j]}} \frac{{\shadw_{i_j}^{[j]2}} {W_{\origin i_j, i_{j+1}}^{[j]} W_{\origin i_j, i'_{j+1}}^{[j]}}}{H^{[j]}} K_{\mathcal{X}_\origin i_{j},i_{j}}^{[j-1]} \left( {\bf x}, {\bf x}' \right)
                    + 
                                                                                   \mathop{\sum}\limits_{i_j} \frac{{\shadw_{i_j}^{[j]2}}                                                                 }{H^{[j]}} K_{\mathcal{X}_\origin i_{j},i_{j}}^{[j-1]} \left( {\bf x}, {\bf x}' \right)
             \Bigg) \Bigg)^l \\

  K_{\mathcal{W}_\origin i_{j+1},i'_{j+1}}^{[j]} \left( {\bf W}_\changein, {\bf W}'_\changein \right) 
  &\!\!\!\!=
   \mathop{\sum}\limits_{l=1}^\infty 
             \Bigg( \frac{1}{{{\shadx}_{i_{j+1}}^{[j]} {\shadx}_{i'_{j+1}}^{[j]}}} \Bigg(
                    \left( 2 b_{\changein i_{j+1}}^{[j]} b_{\changein i'_{j+1}}^{\prime[j]} + \left< {\bf W}_{\changein : i_{j+1}}^{[j]}, {\bf W}_{\changein : i'_{j+1}}^{\prime [j]} \right> \right)
                    + \ldots \\
                    & \!\!\!\!\!\!\!\!\!\!\!\!\!\!\!\ldots \mathop{\sum}\limits_{i_j} {\shadwo}_{i_j,i_{j+1}}^{[j]} {\shadwo}_{i_j,i'_{j+1}}^{[j]} \frac{1                                                                        }{\shadw_{i_j}^{[j]2}} K_{\mathcal{W}_\origin i_{j},i_{j}}^{[j-1]} \left( {\bf W}_\changein, {\bf W}'_\changein \right)
                    + 
                    \mathop{\sum}\limits_{i_j} \frac{W_{\changein i_j, i_{j+1}}^{[j]} W_{\changein i_j, i'_{j+1}}^{\prime [j]}}{\shadw_{i_j}^{[j]2}} K_{\mathcal{W}_\origin i_{j},i_{j}}^{[j-1]} \left( {\bf W}_\changein, {\bf W}'_\changein \right)
                    \Bigg) \Bigg)^l \\
 \end{array}
}}
\]
and:
\[
 \begin{array}{rl}
  K_{\mathcal{X}_\origin i_{1},i'_{1}}^{[0]} \left( {\bf x}, {\bf x}' \right) 
  &\!\!\!\!= 
   \mathop{\sum}\limits_{l=1}^\infty \left( \frac{1}{l!} \right)^2 \tau^{[0](l)} \left( {\tilde{x}}_{\origin i_{1}}^{[0]} \right) \tau^{[0](l)} \left( {\tilde{x}}_{\origin i'_{1}}^{\prime[0]} \right)
             \left( 
                    {{{\shadx}_{i_{1}}^{[0]} {\shadx}_{i'_{1}}^{[0]}}} \left( \frac{1}{2} \alpha^{[0]2} + \frac{1}{2} \frac{1}{H^{[0]}} \left< {\bf x}_{\origin}^{[0]}, {\bf x}_{\origin}^{\prime[0]} \right> \right)
             \right) ^l \\

  K_{\mathcal{W}_\origin i_{1},i'_{1}}^{[0]} \left( {\bf W}_\changein, {\bf W}'_\changein \right) 
  &\!\!\!\!= 
   \mathop{\sum}\limits_{l=1}^\infty 
             \left( 

                    \frac{1}{{{\shadx}_{i_{1}}^{[0]} {\shadx}_{i'_{1}}^{[0]}}} \left( 2 b_{\changein i_{1}}^{[0]} b_{\changein i'_{1}}^{\prime[0]} + 2 \left< {\bf W}_{\changein : i_{1}}^{[0]}, {\bf W}_{\changein : i'_{1}}^{\prime [0]} \right> \right)
             \right)^l \\
 \end{array}
\]

We define:
\begin{equation}
 \begin{array}{rl}
  \theta \left( \zeta \right) 
  &\!\!\!\!= 
  \mathop{\sum}\limits_{l=1}^\infty \zeta^l = \frac{\zeta}{1-\zeta} \;\;\;\; \forall \zeta \in (-1,1) \\

  \sigma^{[j]}_{z,z'} \left( \zeta \right)
  &\!\!\!\!=
  \mathop{\sum}\limits_{l=1}^\infty \left( \frac{1}{l!} \right)^2 \tau^{[j](l)} \left( z \right) \tau^{[j](l)} \left( z' \right) \zeta^l \\
 \end{array}
\label{eq:euceuc_lambda}
\end{equation}
$\forall j \in \infset{N}_D$, $z \in \infset{R}$, noting that:
\begin{equation}
 \begin{array}{rl}
  \theta^{-1} \left( \zeta \right) 
  &\!\!\!\!= 
  \frac{\zeta}{1+\zeta} \;\;\;\; \forall \zeta \in (1,\infty) \\
 \end{array}
\label{eq:euceuc_lambda_inv}
\end{equation}
Note that $\sigma^{[j]}_{z,z'}$ are derived from the power-series representation of the 
activation functions about $z$ in such a manner that the constant term is 
removed, and, if $z=z'$, so too are all signs, making $\sigma^{[j]}_{z,z}$ 
an increasing function.\footnote{As an aside, we note that the power series 
representations of the activation functions $\tau^{[j]}$ about any $z$ 
represent the same underlying function, minus $\tau^{[j]} (0)$, and so in 
principle we can start with a single such representation and use analytic 
continuation and reconstruct $\tau^{[j]}$ everywhere.  Unfortunately the same 
is not true of $\sigma^{[j]}_{z,z'}$, which is perhaps unfortunate in this 
context.}  Using this definition and observation, provided the argument 
remains in the ROC of all relevant series, we see that:
\begin{equation}
{{
 \begin{array}{rl}
  {\bf K}_{\mathcal{X}_\origin} \left( {\bf x}, {\bf x}' \right) 
  &\!\!\!\!=
  {\bf K}_{\mathcal{X}_\origin}^{[D-1]} \left( {\bf x}, {\bf x}' \right)  \\

  {\bf K}_{\mathcal{W}_\origin} \left( {\bf W}_\changein, {\bf W}'_\changein \right) 
  &\!\!\!\!=
  {\bf K}_{\mathcal{W}_\origin}^{[D-1]} \left( {\bf W}_\changein, {\bf W}'_\changein \right)  \\
 \end{array}
 }}
\label{eq:inducedkdef}
\end{equation}
where, recursively $\forall j \in \infset{N}_D$:
\begin{equation}
{\!\!\!\!{
 \begin{array}{rl}
  K_{\mathcal{X}_\origin i_{j\!+\!1}, i'_{j\!+\!1}}^{[j]} \!\!\!\left( {\bf x}, {\bf x}' \right) 
  &\!\!\!\!\!=\! \left\{ \!\!\begin{array}{ll}
\sigma^{[j]}_{{\tilde{x}}_{\origin i_{j\!+\!1}}^{[j]}, {\tilde{x}}_{\origin i'_{j\!+\!1}}^{\prime[j]}} \!\!\!\left( \!\!{{{\shadx}_{i_{j\!+\!1}}^{[j]} {\shadx}_{i'_{j\!+\!1}}^{[j]}}} \!\!\left( \!\!\begin{array}{l}
                    \left( \frac{1}{2} \alpha^{[j]2} + \frac{1}{H^{[j]}} \left< {\bf x}_{\origin}^{[j]}, {\bf x}_{\origin}^{\prime[j]} \right> \right)
                    + \ldots \\ \hfill \ldots
                    \mathop{\sum}\limits_{i_j} \left( \frac{{W_{\origin i_j, i_{j\!+\!1}}^{[j]} W_{\origin i_j, i'_{j\!+\!1}}^{[j]}}}{{\shadwo}_{i_j,i_{j\!+\!1}}^{[j]} {\shadwo}_{i_j,i'_{j\!+\!1}}^{[j]}} + 1 \right) \frac{\shadw_{i_j}^{[j]2}}{H^{[j]}} K_{\mathcal{X}_\origin i_{j},i_{j}}^{[j-1]} \left( {\bf x}, {\bf x}' \right)
             \end{array} \!\!\!\right) \right) &
                     \mbox{if } j > 0 \\
\sigma^{[0]}_{{\tilde{x}}_{\origin i_{1}}^{[0]}, {\tilde{x}}_{\origin i'_{1}}^{\prime[0]}} \!\!\!\left(
                    {{          {\shadx}_{i_{1}}^{[0]}           {\shadx}_{i'_{1}}^{[0]}}} \left( \frac{1}{2} \alpha^{[0]2} + \frac{1}{2} \frac{1}{H^{[0]}} \left< {\bf x}_{\origin}^{[0]}, {\bf x}_{\origin}^{\prime[0]} \right> \right)
             \right) &
                     \mbox{if } j = 0 \\
   \end{array} \right. \\

  K_{\mathcal{W}_\origin i_{j\!+\!1}, i'_{j\!+\!1}}^{[j]} \!\!\!\left( {\bf W}_\changein, {\bf W}'_\changein \right) 
  &\!\!\!\!\!=\! \left\{ \!\!\begin{array}{ll}
             \theta \!\left( \!\!\frac{1}{{{\shadx}_{i_{j\!+\!1}}^{[j]} {\shadx}_{i'_{j\!+\!1}}^{[j]}}} \!\!\left( \!\!\begin{array}{l}
                    \left( 2 b_{\changein i_{j\!+\!1}}^{[j]} b_{\changein i'_{j\!+\!1}}^{\prime[j]} + \left< {\bf W}_{\changein : i_{j\!+\!1}}^{[j]}, {\bf W}_{\changein : i'_{j\!+\!1}}^{\prime [j]} \right> \right)
                    + \ldots \\ \hfill \ldots
                    \mathop{\sum}\limits_{i_j} \left( {\shadwo}_{i_j,i_{j\!+\!1}}^{[j]} {\shadwo}_{i_j,i'_{j\!+\!1}}^{[j]} + W_{\changein i_j, i_{j\!+\!1}}^{[j]} W_{\changein i_j, i'_{j\!+\!1}}^{\prime [j]} \right) \frac{K_{\mathcal{W}_\origin i_{j},i_{j}}^{[j-1]} \left( {\bf W}_\changein, {\bf W}'_\changein \right)}{\shadw_{i_j}^{[j]2}} 
                  \end{array} \!\!\!\right) \right) &
                     \mbox{if } j > 0 \\
             \theta \!\left(
                    \frac{1}{{{\shadx}_{i_{1}}^{[0]} {\shadx}_{i'_{1}}^{[0]}}} \left( 2 b_{\changein i_{1}}^{[0]} b_{\changein i'_{1}}^{\prime[0]} + 2 \left< {\bf W}_{\changein : i_{1}}^{[0]}, {\bf W}_{\changein : i'_{1}}^{\prime [0]} \right> \right)
                  \right) &
                     \mbox{if } j = 0 \\
   \end{array} \right. \\
 \end{array}
}}
\label{eq:inducedkrec}
\end{equation}

\subsection{Induced Kernel Gradients}

Later in the paper we will require the gradients of induced kernels.  Recalling 
(\ref{eq:inducedkdef}), (\ref{eq:inducedkrec}) and applying the chain rule we 
see that:
\begin{equation}
{{
 \begin{array}{rl}
  \nabla_{\bf x} {\bf K}_{\mathcal{X}_\origin} \left( {\bf x}, {\bf x}' \right) 
  &\!\!\!\!=
  \left[ \begin{array}{c} \left[ \begin{array}{c} 
  \frac{\partial}{\partial x_{i''_1}} \nabla_{\bf x} K_{\mathcal{X}_\origin i_{j+1}, i'_{j+1}}^{[D-1]} \left( {\bf x}, {\bf x}' \right) 
  \end{array} \right]_{i''_{1}} \end{array} \right]_{i_{j+1},i'_{j+1}} \\

  \nabla_{{\bf W}_\changein^{[j']}} {\bf K}_{\mathcal{W}_\origin} \left( {\bf W}_\changein, {\bf W}'_\changein \right) 
  &\!\!\!\!=
  \left[ \begin{array}{c} \left[ \begin{array}{c} 
  \frac{\partial}{\partial W_{\changein i''_{j'},i''_{j'+1}}^{[j']}} K_{\mathcal{W}_\origin i_{j+1}, i'_{j+1}}^{[D-1]} \left( {\bf W}_\changein, {\bf W}'_\changein \right) 
  \end{array} \right]_{i''_{j'},i''_{j'+1}} \end{array} \right]_{i_{j+1},i'_{j+1}} \\

  \nabla_{{\bf b}_\changein^{[j']}} {\bf K}_{\mathcal{W}_\origin} \left( {\bf W}_\changein, {\bf W}'_\changein \right) 
  &\!\!\!\!=
  \left[ \begin{array}{c} \left[ \begin{array}{c} 
  \frac{\partial}{\partial b_{\changein i''_{j'+1}}^{[j']}} K_{\mathcal{W}_\origin i_{j+1}, i'_{j+1}}^{[D-1]} \left( {\bf W}_\changein, {\bf W}'_\changein \right) 
  \end{array} \right]_{i''_{j'+1}} \end{array} \right]_{i_{j+1},i'_{j+1}} \\
 \end{array}
 }}
\label{eq:inducedkdefgrad}
\end{equation}
where, recursively $\forall j \in \infset{N}_D$:
\begin{equation}
 {{
 \begin{array}{rl}
  \frac{\partial}{\partial x_{i''_1}} K_{\mathcal{X}_\origin i_{j\!+\!1}, i'_{j\!+\!1}}^{[j]} \!\!\left( {\bf x}, {\bf x}' \right) 
  &\!\!\!\!\!=\! \left\{ \!\!\!\begin{array}{ll} \!\!\!\!\begin{array}{r}
\sigma^{[j](1)}_{{\tilde{x}}_{\origin i_{j\!+\!1}}^{[j]}, {\tilde{x}}_{\origin i'_{j\!+\!1}}^{\prime[j]}} \!\!\left( \!\!{{{\shadx}_{i_{j\!+\!1}}^{[j]} {\shadx}_{i'_{j\!+\!1}}^{[j]}}} \!\!\left( \!\!\!\begin{array}{l}
                    \left( \frac{1}{2} \alpha^{[j]2} + \frac{1}{H^{[j]}} \left< {\bf x}_{\origin}^{[j]}, {\bf x}_{\origin}^{\prime[j]} \right> \right)
                    + \ldots \\ \hfill \ldots
                    \mathop{\sum}\limits_{i_j} \!\left(\! \frac{{W_{\origin i_j, i_{j\!+\!1}}^{[j]} W_{\origin i_j, i'_{j\!+\!1}}^{[j]}}}{{\shadwo}_{i_j,i_{j\!+\!1}}^{[j]} {\shadwo}_{i_j,i'_{j\!+\!1}}^{[j]}} \!+\! 1 \!\right) 
                    \frac{\shadw_{i_j}^{[j]2}}{H^{[j]}} K_{\mathcal{X}_\origin i_{j},i_{j}}^{[j-1]} \left( {\bf x}, {\bf x}' \right)
             \end{array} \!\!\!\right) \right) \ldots \\ \ldots {\shadx}_{i_{j\!+\!1}}^{[j]} {\shadx}_{i'_{j\!+\!1}}^{[j]} 
             \mathop{\sum}\limits_{i_j} \left( \frac{{W_{\origin i_j, i_{j\!+\!1}}^{[j]} W_{\origin i_j, i'_{j\!+\!1}}^{[j]}}}{{\shadwo}_{i_j,i_{j\!+\!1}}^{[j]} {\shadwo}_{i_j,i'_{j\!+\!1}}^{[j]}} + 1 \right) 
                     \frac{\shadw_{i_j}^{[j]2}}{H^{[j]}} \frac{\partial}{\partial x_{i''_1}}  K_{\mathcal{X}_\origin i_{j},i_{j}}^{[j-1]} \left( {\bf x}, {\bf x}' \right) \!\!\!\!\end{array} &
                     \mbox{if } j > 0 \\ \\

\sigma^{[0](1)}_{{\tilde{x}}_{\origin i_{1}}^{[0]}, {\tilde{x}}_{\origin i'_{1}}^{\prime[0]}} \left(
                    {{{\shadx}_{i_{1}}^{[0]} {\shadx}_{i'_{1}}^{[0]}}} \left( \frac{1}{2} \alpha^{[0]2} + \frac{1}{2} \frac{1}{H^{[0]}} \left< {\bf x}, {\bf x}' \right> \right)
             \right) {{{\shadx}_{i_{1}}^{[0]} {\shadx}_{i'_{1}}^{[0]}}} \frac{1}{H^{[0]}} \delta_{i_1,i''_1} \frac{1}{2} x'_{i'_1} &
                     \mbox{if } j = 0 \\
   \end{array} \right. \\
 \end{array}
 }}
\label{eq:inducedkrecgradx}
\end{equation}
\begin{equation}
 {{
 \begin{array}{l}
  \frac{\partial}{\partial W_{\changein i''_{j'},i''_{j'\!+\!1}}^{[j']}} \!K_{\mathcal{W}_\origin i_{j\!+\!1}, i'_{j\!+\!1}}^{[j]} \!\left( {\bf W}_\changein, {\bf W}'_\changein \right) 
  = \ldots \\ \ldots = \left\{ \!\!\!\begin{array}{ll}
             \begin{array}{r} \!\!\!\theta^{(1)} \!\left( \!\frac{1}{{{\shadx}_{i_{j\!+\!1}}^{[j]} {\shadx}_{i'_{j\!+\!1}}^{[j]}}} \!\left( \!\!\!\begin{array}{l}
                    \left( 2 b_{\changein i_{j\!+\!1}}^{[j]} b_{\changein i'_{j\!+\!1}}^{\prime[j]} + \left< {\bf W}_{\changein : i_{j\!+\!1}}^{[j]}, {\bf W}_{\changein : i'_{j\!+\!1}}^{\prime [j]} \right> \right)
                    + \ldots \\ \hfill \ldots
                    \mathop{\sum}\limits_{i_j} \left( {\shadwo}_{i_j,i_{j\!+\!1}}^{[j]} {\shadwo}_{i_j,i'_{j\!+\!1}}^{[j]} \!\!+\! W_{\changein i_j, i_{j\!+\!1}}^{[j]} W_{\changein i_j, i'_{j\!+\!1}}^{\prime [j]} \right) 
                    \frac{K_{\mathcal{W}_\origin i_{j},i_{j}}^{[j-1]} \!\!\left( {\bf W}_\changein, {\bf W}'_\changein \right)}{\shadw_{i_j}^{[j]2}} 
                  \!\!\!\end{array} \!\right) \!\!\right) \ldots \\
                     \ldots \frac{1}{{{\shadx}_{i_{j\!+\!1}}^{[j]} {\shadx}_{i'_{j\!+\!1}}^{[j]}}} 
                     \mathop{\sum}\limits_{i_j} \left( {\shadwo}_{i_j,i_{j\!+\!1}}^{[j]} {\shadwo}_{i_j,i'_{j\!+\!1}}^{[j]} \!\!+\! W_{\changein i_j, i_{j\!+\!1}}^{[j]} W_{\changein i_j, i'_{j\!+\!1}}^{\prime [j]} \right) 
                    \!\frac{1}{\shadw_{i_j}^{[j]2}} \frac{\partial}{\partial W_{\changein i''_{j'},i''_{j'\!+\!1}}^{[j']}} K_{\mathcal{W}_\origin i_{j},i_{j}}^{[j-1]} \!\!\left( {\bf W}_\changein, {\bf W}'_\changein \right)
                    \end{array}\!\!\! &
                     \!\!\mbox{if } j > j' \\ \\

             \begin{array}{r} \!\!\!\theta^{(1)} \!\left( \!\frac{1}{{{\shadx}_{i_{j\!+\!1}}^{[j]} {\shadx}_{i'_{j\!+\!1}}^{[j]}}} \!\left( \!\!\!\begin{array}{l}
                    \left( 2 b_{\changein i_{j\!+\!1}}^{[j]} b_{\changein i'_{j\!+\!1}}^{\prime[j]} + \left< {\bf W}_{\changein : i_{j\!+\!1}}^{[j]}, {\bf W}_{\changein : i'_{j\!+\!1}}^{\prime [j]} \right> \right)
                    + \ldots \\ \hfill \ldots
                    \mathop{\sum}\limits_{i_j} \left( {\shadwo}_{i_j,i_{j\!+\!1}}^{[j]} {\shadwo}_{i_j,i'_{j\!+\!1}}^{[j]} \!\!+\! W_{\changein i_j, i_{j\!+\!1}}^{[j]} W_{\changein i_j, i'_{j\!+\!1}}^{\prime [j]} \right) 
                    \frac{K_{\mathcal{W}_\origin i_{j},i_{j}}^{[j-1]} \!\!\left( {\bf W}_\changein, {\bf W}'_\changein \right)}{\shadw_{i_j}^{[j]2}} 
                  \!\!\!\end{array} \!\right) \!\!\right) \ldots \\
                     \ldots \frac{1}{{{\shadx}_{i''_{j\!+\!1}}^{[j]} {\shadx}_{i'_{j\!+\!1}}^{[j]}}} 
                     \delta_{i_{j\!+\!1},i''_{j\!+\!1}} \left( 1 + \frac{K_{\mathcal{W}_\origin i''_{j},i''_{j}}^{[j-1]} \left( {\bf W}_\changein, {\bf W}'_\changein \right)}{\shadw_{i''_j}^{[j]2}} \right) W_{\changein i''_j, i'_{j\!+\!1}}^{\prime [j]} 
                     \end{array}\!\!\! &
                     \!\!\mbox{if } j = j' > 0 \\ \\

             \begin{array}{r} \!\!\!\theta^{(1)} \!\left(
                    \frac{1}{{{\shadx}_{i_{1}}^{[0]} {\shadx}_{i'_{1}}^{[0]}}} \left( 2 b_{\changein i_{1}}^{[0]} b_{\changein i'_{1}}^{\prime[0]} + 2 \left< {\bf W}_{\changein : i_{1}}^{[0]}, {\bf W}_{\changein : i'_{1}}^{\prime [0]} \right> \right)
                  \right) \frac{1}{{{\shadx}_{i''_{1}}^{[0]} {\shadx}_{i'_{1}}^{[0]}}} \delta_{i_{1},i''_{1}} 2W_{\changein i''_0, i'_{1}}^{\prime [0]} \end{array}\!\!\! &
                     \!\!\mbox{if } j = j' = 0 \\
   \end{array} \right. \\
 \end{array}
 }}
\label{eq:inducedkrecgradw}
\end{equation}
\begin{equation}
 {{
 \begin{array}{l}
  \frac{\partial}{\partial b_{\changein i''_{j'\!+\!1}}^{[j']}} \!K_{\mathcal{W}_\origin i_{j\!+\!1}, i'_{j\!+\!1}}^{[j]} \!\left( {\bf W}_\changein, {\bf W}'_\changein \right) 
  = \ldots \\ \ldots = \left\{ \!\!\!\begin{array}{ll}
             \begin{array}{r} \!\!\!\theta^{(1)} \!\left( \!\frac{1}{{{\shadx}_{i_{j\!+\!1}}^{[j]} {\shadx}_{i'_{j\!+\!1}}^{[j]}}} \!\left( \!\!\!\begin{array}{l}
                    \left( 2 b_{\changein i_{j\!+\!1}}^{[j]} b_{\changein i'_{j\!+\!1}}^{\prime[j]} + \left< {\bf W}_{\changein : i_{j\!+\!1}}^{[j]}, {\bf W}_{\changein : i'_{j\!+\!1}}^{\prime [j]} \right> \right)
                    + \ldots \\ \hfill \ldots
                    \mathop{\sum}\limits_{i_j} \left( {\shadwo}_{i_j,i_{j\!+\!1}}^{[j]} {\shadwo}_{i_j,i'_{j\!+\!1}}^{[j]} \!\!+\! W_{\changein i_j, i_{j\!+\!1}}^{[j]} W_{\changein i_j, i'_{j\!+\!1}}^{\prime [j]} \right) 
                    \frac{K_{\mathcal{W}_\origin i_{j},i_{j}}^{[j-1]} \!\!\left( {\bf W}_\changein, {\bf W}'_\changein \right)}{\shadw_{i_j}^{[j]2}} 
                  \!\!\!\end{array} \!\right) \!\!\right) \ldots \\
                     \ldots \frac{1}{{{\shadx}_{i_{j\!+\!1}}^{[j]} {\shadx}_{i'_{j\!+\!1}}^{[j]}}} 
                     \mathop{\sum}\limits_{i_j} \left( {\shadwo}_{i_j,i_{j\!+\!1}}^{[j]} {\shadwo}_{i_j,i'_{j\!+\!1}}^{[j]} \!\!+\! W_{\changein i_j, i_{j\!+\!1}}^{[j]} W_{\changein i_j, i'_{j\!+\!1}}^{\prime [j]} \right) 
                     \frac{1}{\shadw_{i_j}^{[j]2}}  
                    \!\frac{\partial}{\partial b_{\changein i''_{j'\!+\!1}}^{[j']}} 
                     K_{\mathcal{W}_\origin i_{j},i_{j}}^{[j-1]} \!\!\left( {\bf W}_\changein, {\bf W}'_\changein \right)
                     \end{array}\!\!\! &
                     \!\!\mbox{if } j > j' \\ \\

             \begin{array}{r} \!\!\!\theta^{(1)} \!\left( \!\frac{1}{{{\shadx}_{i_{j\!+\!1}}^{[j]} {\shadx}_{i'_{j\!+\!1}}^{[j]}}} \!\left( \!\!\!\begin{array}{l}
                    \left( 2b_{\changein i_{j\!+\!1}}^{[j]} b_{\changein i'_{j\!+\!1}}^{\prime[j]} + \left< {\bf W}_{\changein : i_{j\!+\!1}}^{[j]}, {\bf W}_{\changein : i'_{j\!+\!1}}^{\prime [j]} \right> \right)
                    + \ldots \\ \hfill \ldots
                    \mathop{\sum}\limits_{i_j} \left( {\shadwo}_{i_j,i_{j\!+\!1}}^{[j]} {\shadwo}_{i_j,i'_{j\!+\!1}}^{[j]} \!\!+\! W_{\changein i_j, i_{j\!+\!1}}^{[j]} W_{\changein i_j, i'_{j\!+\!1}}^{\prime [j]} \right) 
                    \frac{K_{\mathcal{W}_\origin i_{j},i_{j}}^{[j-1]} \!\!\left( {\bf W}_\changein, {\bf W}'_\changein \right)}{\shadw_{i_j}^{[j]2}} 
                  \!\!\!\end{array} \!\right) \!\!\right) \ldots \\
                     \ldots \frac{1}{{{\shadx}_{i''_{j\!+\!1}}^{[j]} {\shadx}_{i'_{j\!+\!1}}^{[j]}}} 
                     \delta_{i_{j\!+\!1},i''_{j\!+\!1}} 2 b_{\changein i'_{j\!+\!1}}^{\prime [j]} \end{array}\!\!\! &
                     \!\!\mbox{if } j = j' > 0 \\ \\

             \begin{array}{r} \!\!\!\theta^{(1)} \!\left(
                    \frac{1}{{{\shadx}_{i_{1}}^{[0]} {\shadx}_{i'_{1}}^{[0]}}} \left( 2 b_{\changein i_{1}}^{[0]} b_{\changein i'_{1}}^{\prime[0]} + 2 \left< {\bf W}_{\changein : i_{1}}^{[0]}, {\bf W}_{\changein : i'_{1}}^{\prime [0]} \right> \right)
                  \right) \frac{1}{{{\shadx}_{i''_{1}}^{[0]} {\shadx}_{i'_{1}}^{[0]}}} \delta_{i_{1},i''_{1}} 2 b_{\changein i'_{1}}^{\prime [0]} \end{array}\!\!\! &
                     \!\!\mbox{if } j = j' = 0 \\
   \end{array} \right. \\
 \end{array}
 }}
\label{eq:inducedkrecgradb}
\end{equation}

\subsection{Induced Norms} \label{sec:induced_norm}

We now consider the norms induced by the feature maps.  Precisely, we wish to 
impose norms on $\mathcal{X}_\origin \subset \infset{R}^{\infty \times m}$ and 
$\mathcal{W}_\origin \subset \infset{R}^{\infty \times m}$ and calculate $\| 
{{\bg{\Phi}}}_{\origin} ({\bf x}) \|_{\mathcal{X}_\origin}$, $\| 
{{\bg{\Psi}}}_{\origin} ({\bf W}_\changein) \|_{\mathcal{W}_\origin}$ 
from this, which will subsequently provide the {\em induced} norms we will 
require when constructing the reproducing kernel Banach space for ${\bf 
f}_\changein$.  In principle we could apply an arbitrary norm to these spaces.  
However, recall that:
\[
 \begin{array}{l}
  {\bf f}_{\changein} \left( {\bf x} \right) = 
  \left< {{\bg{\Phi}}}_{\origin} \left( {\bf x} \right), {{\bg{\Psi}}}_{\origin} \left( {\bf W}_\changein \right) \right>_{\mathcal{X}_\origin \times \mathcal{W}_\origin} 
 \end{array}
\]
and hence:
\[
 \begin{array}{l}
  \left\| {\bf f}_{\changein} \left( {\bf x} \right) \right\|_2 = 
  \left< {{\bg{\Phi}}}_{\origin} \left( {\bf x} \right), {{\bg{\Psi}}}_{\origin} \left( {\bf W}_\changein \right) \right>_F 
 \end{array}
\]
so a natural choice is a pair of dual norms - that is, a pair of norms $\| 
\cdot \|_{\mathcal{X}_\origin}$ and $\| \cdot \|_{\mathcal{W}_\origin}$ such 
that:
\[
 \begin{array}{rl}
  \left\| {\bg{\Xi}} \right\|_{\mathcal{X}_\origin} = \mathop{\sup}\limits_{{\bg{\Omega}} \in \mathcal{W}_\origin} \left\{ \left. 
  \left< {{\bg{\Xi}}}, {{\bg{\Omega}}} \right>_F
  \right| 
  \left\| {\bg{\Omega}} \right\|_{\mathcal{W}_\origin} \leq 1 \right\}
 \end{array}
\]
and hence by H{\"o}lder's inequality:
\[
 \begin{array}{l}
  \left\| {\bf f}_\changein \left( {\bf x} \right) \right\|_2 \leq 
  \left\| {{\bg{\Phi}}}_{\origin} \left( {\bf x} \right) \right\|_{\mathcal{X}_\origin} \left\| {{\bg{\Psi}}}_{\origin} \left( {\bf W}_\changein \right) \right\|_{\mathcal{W}_\origin}
 \end{array}
\]

In the present context we find it convenient to use the Frobenius (elementwise 
Euclidean) norms:
\[
 \begin{array}{rl}
  \left\| {\bg{\Xi}}    \right\|_{\mathcal{X}_\origin}^2 &\!\!\!\!= \left\| {\bg{\Xi}}    \right\|_F^2 = 
\sum_{i_D} \left\| {\bg{\Xi}}_{:i_D}    \right\|_2^2 \\

  \left\| {\bg{\Omega}} \right\|_{\mathcal{W}_\origin}^2 &\!\!\!\!= \left\| {\bg{\Omega}} \right\|_F^2 = 
\sum_{i_D} \left\| {\bg{\Omega}}_{:i_D} \right\|_2^2 \\
 \end{array}
\]
and so:
\[
 \begin{array}{rl}
  \left\| {{\bg{\Phi}}}_{\origin} \left( {\bf x} \right) \right\|_{\mathcal{X}_\origin}^2
  &\!\!\!\!=
  {\rm Tr} \left( {\bf K}_{\mathcal{X}_\origin} \left( {\bf x}, {\bf x} \right) \right) \\

  \left\| {{\bg{\Psi}}}_{\origin} \left( {\bf W}_\changein \right) \right\|_{\mathcal{W}_\origin}^2
  &\!\!\!\!=
  {\rm Tr} \left( {\bf K}_{\mathcal{W}_\origin} \left( {\bf W}_\changein, {\bf W}_\changein \right) \right) \\
 \end{array}
\]

Before proceeding we define:
\[
 \begin{array}{rl}
 \begin{array}{rl}
  s^{[j]2}
  &\!\!\!\!=
  \left\{ \begin{array}{ll}
   \frac{1}{2} \alpha^{[0]2} + \frac{1}{2} \frac{H^{ [-1]}}{H^{[0]}} M^{ [-1]2} & \mbox{if } j = 0 \\
   \frac{1}{2} \alpha^{[j]2} +             \frac{H^{[j-1]}}{H^{[j]}} M^{[j-1]2} & \mbox{otherwise} \\
  \end{array} \right. \\

  t_{\changein i_{j+1}}^{[j]2} 
  &\!\!\!\!= 
  \left\{ \begin{array}{ll}
   \left[ 2b_{\changein i_{  1}}^{[0]2} + 2 \left\| {\bf W}_{\changein :i_{  1}}^{[0]} \right\|_2^2 \right]_{i_{  1}} & \mbox{if } j = 0 \\
   \left[ 2b_{\changein i_{j+1}}^{[j]2} +   \left\| {\bf W}_{\changein :i_{j+1}}^{[j]} \right\|_2^2 \right]_{i_{j+1}} & \mbox{otherwise} \\
  \end{array} \right. \\
 \end{array} & \forall j \in \infset{N}_D, i_{j+1}
 \end{array}
\]
where, roughly speaking, ${t}^{[j-1]2}$ represents the magnitude of the 
weight-step for layer $j$ and $s^{[j]2}$ represents the upper bound on the 
image of a vector ${\bf x} \in \infset{X}$ at the input to layer $j$.  Using 
this notation:
\begin{equation}
{{
 \begin{array}{rl}
  \left\| {{\bg{\Phi}}}_{\origin} \left( {\bf x} \right) \right\|_{\mathcal{X}_\origin}^2
  &\!\!\!\!=
  \sum_{i_D} \left\| {{\bg{\Phi}}}_{\origin:i_{D}}^{[D-1]} \left( {\bf x} \right) \right\|_2^2 \\

  \left\| {{\bg{\Psi}}}_{\origin} \left( {\bf W}_\changein \right) \right\|_{\mathcal{W}_\origin}^2
  &\!\!\!\!=
  \sum_{i_D} \left\| {{\bg{\Psi}}}_{\origin:i_{D}}^{[D-1]} \left( {\bf W}_\changein \right) \right\|_2^2 \\
 \end{array}
 }}
 \label{eq:euceuc_normfindef}
\end{equation}
where, recursively $\forall j \in \infset{N}_D$:
\begin{equation}
 {{
 \begin{array}{rl}
  \left\| {{\bg{\Phi}}}_{\origin:i_{j+1}}^{[j]} \left( {\bf x} \right) \right\|_2^2
  &\!\!\!\!= \left\{ \begin{array}{ll}
\sigma^{[j]}_{\!\!{\tilde{x}}_{\origin i_{j+1}}^{[j]}, {\tilde{x}}_{\origin i_{j+1}}^{[j]}} \left( \begin{array}{l} {\shadx}_{i_{j+1}}^{[j]2} \Bigg(
                    \left( \frac{1}{2} \alpha^{[j]2} + \frac{1}{H^{[j]}} \left\| {\bf x}_{\origin}^{[j]} \right\|_2^2 \right)
                    + \ldots \\ \;\;\; \ldots
                    \mathop{\sum}\limits_{i_j} \left( 1 + \frac{W_{\origin i_j, i_{j+1}}^{[j]2}}{{\shadwo}_{i_j,i_{j+1}}^{[j]2}} \right) \frac{\shadw_{i_j}^{[j]2}}{H^{[j]}} \left\| {{\bg{\Phi}}}_{\origin:i_{j}}^{[j-1]} \left( {\bf x} \right) \right\|_2^2
             \Bigg) \end{array} \right) &
                     \mbox{if } j > 0 \\
\sigma^{[0]}_{{\tilde{x}}_{\origin i_{1}}^{[0]}, {\tilde{x}}_{\origin i_{1}}^{[0]}} \left(
                    {{\shadx}_{i_{1}}^{[0]2}} \left( \frac{1}{2} \alpha^{[0]2} + \frac{1}{2} \frac{1}{H^{[0]}} \left\| {\bf x}_{\origin}^{[0]} \right\|_2^2 \right)
             \right) &
                     \mbox{if } j = 0 \\
   \end{array} \right. \\

  \left\| {{\bg{\Psi}}}_{\origin:i_{j+1}}^{[j]} \left( {\bf W}_\changein \right) \right\|_2^2
  &\!\!\!\!= \left\{ \begin{array}{ll}
             \theta \left( \frac{1}{{\shadx}_{i_{j+1}}^{[j]2}} \left(
                    t_{\changein i_{j+1}}^{[j]2} 
                    +
                    \mathop{\sum}\limits_{i_j} \left( {\shadwo}_{i_j,i_{j+1}}^{[j]2} + W_{\changein i_j, i_{j+1}}^{[j]2} \right) \frac{\left\| {{\bg{\Psi}}}_{\origin:i_{j}}^{[j-1]} \left( {\bf W}_\changein \right) \right\|_2^2}{\shadw_{i_j}^{[j]2}} 
                  \right) \right) &
                     \mbox{if } j > 0 \\
             \theta \left(
                    \frac{1}{\shadx_{i_1}^{[0]2}} t_{\changein i_1}^{[0]2} 
                  \right) &
                     \mbox{if } j = 0 \\
   \end{array} \right. \\
 \end{array}
 }}
\label{eq:euceuc_normsdefof}
\end{equation}

\subsection{Induced Norm Gradients}

Later in the paper we will require the gradients of induced norms.  Recalling 
(\ref{eq:euceuc_normfindef}), (\ref{eq:euceuc_normsdefof}) and applying the chain rule we 
see that:
\begin{equation}
{{
 \begin{array}{rl}
  \nabla_{\bf x} \left\| {\bg{\Phi}}_{\origin} \left( {\bf x} \right) \right\|_{\mathcal{X}_\origin}^2
  &\!\!\!\!=
  \sum_{i_D} 
  \left[ \begin{array}{c} \frac{\partial}{\partial x_{i''_1}} \left\| {\bg{\Phi}}_{\origin :i_D}^{[D-1]} \left( {\bf x} \right) \right\|_2^2 \end{array} \right]_{i''_1} \\

  \nabla_{{\bf W}_\changein^{[j']}} \left\| {\bg{\Psi}}_{\origin} \left( {\bf W}_\changein \right) \right\|_{\mathcal{W}_\origin}^2
  &\!\!\!\!=
  \sum_{i_D} 
  \left[ \begin{array}{c} \frac{\partial}{\partial W_{\changein i''_{j'},i''_{j'+1}}^{[j']}} \left\| {\bg{\Psi}}_{\origin :i_D}^{[D-1]} \left( {\bf W}_\changein \right) \right\|_2^2 \end{array} \right]_{i''_{j'},i''_{j'+1}} \\

  \nabla_{{\bf b}_\changein^{[j']}} \left\| {\bg{\Psi}}_{\origin} \left( {\bf W}_\changein \right) \right\|_{\mathcal{W}_\origin}^2
  &\!\!\!\!=
  \sum_{i_D} 
  \left[ \begin{array}{c} \frac{\partial}{\partial b_{\changein i''_{j'}}^{[j']}} \left\| {\bg{\Psi}}_{\origin :i_D}^{[D-1]} \left( {\bf W}_\changein \right) \right\|_2^2 \end{array} \right]_{i''_{j'+1}} \\
 \end{array}
 }}
\label{eq:inducedndefgrad}
\end{equation}
where, recursively $\forall j \in \infset{N}_D$:
\begin{equation}
 {{
 \begin{array}{l}
  \frac{\partial}{\partial x_{i'_1}} \left\| {\bg{\Phi}}_{\origin :i_{j+1}}^{[j]} \left( {\bf x} \right) \right\|_2^2
  = \ldots \\ \ldots = \left\{ \begin{array}{ll} \!\!\!\!\begin{array}{r}
\sigma^{[j](1)}_{{\tilde{x}}_{\origin i_{j+1}}^{[j]}, {\tilde{x}}_{\origin i_{j+1}}^{[j]}} \left( {{{\shadx}_{i_{j+1}}^{[j]2}}} \left( 
                    \left( \frac{1}{2} \alpha^{[j]2} + \frac{1}{H^{[j]}} \left\| {\bf x}_{\origin}^{[j]} \right\|_2^2 \right)
                    +
                    \mathop{\sum}\limits_{i_j} \left( \frac{{W_{\origin i_j, i_{j+1}}^{[j]2}}}{{\shadwo}_{i_j,i_{j+1}}^{[j]2}} + 1 \right) 
                    \frac{\shadw_{i_j}^{[j]2}}{H^{[j]}} \left\| {\bg{\Phi}}_{\origin :i_{j}}^{[j-1]} \left( {\bf x} \right) \right\|_2^2
             \right) \right) \ldots \\ \ldots {\shadx}_{i_{j+1}}^{[j]2} \mathop{\sum}\limits_{i_j} \left( \frac{{W_{\origin i_j, i_{j+1}}^{[j]2}}}{{\shadwo}_{i_j,i_{j+1}}^{[j]2}} + 1 \right) 
                   \frac{\shadw_{i_j}^{[j]2}}{H^{[j]}} \frac{\partial}{\partial x_{i'_1}} \left\| {\bg{\Phi}}_{\origin :i_{j}}^{[j-1]} \left( {\bf x} \right) \right\|_2^2
             \!\!\!\!\end{array} &
                     \mbox{if } j > 0 \\ \\

\sigma^{[0](1)}_{{\tilde{x}}_{\origin i_{1}}^{[0]}, {\tilde{x}}_{\origin i_{1}}^{[0]}} \left(
                    {{{\shadx}_{i_{1}}^{[0]2}}} \left( \frac{1}{2} \alpha^{[0]2} + \frac{1}{2} \frac{1}{H^{[0]}} \left\| {\bf x} \right\|_2^2 \right)
             \right) {{{\shadx}_{i_{1}}^{[0]2}}} \frac{1}{H^{[0]}} \delta_{i_1,i'_1} x_{i'_1} &
                     \mbox{if } j = 0 \\
   \end{array} \right. \\
 \end{array}
 }}
\label{eq:inducednrecgradx}
\end{equation}
\begin{equation}
 {{
 \begin{array}{l}
  \frac{\partial}{\partial W_{\changein i'_{j'},i'_{j'+1}}^{[j']}} \left\| {\bg{\Psi}}_{\origin :i_{j+1}}^{[j]} \left( {\bf W}_\changein \right) \right\|_2^2
  = \ldots \\ \ldots = \left\{ \begin{array}{ll}
             \begin{array}{r} \theta^{(1)} \left( \frac{1}{{{\shadx}_{i_{j+1}}^{[j]2}}} \left( 
                    t_{\changein i_{j+1}}^{[j]} 
                    +
                    \mathop{\sum}\limits_{i_j} \left( {\shadwo}_{i_j,i_{j+1}}^{[j]2} + W_{\changein i_j, i_{j+1}}^{[j]2} \right) 
                    \frac{\left\| {\bg{\Psi}}_{\origin :i_{j}}^{[j-1]} \left( {\bf W}_\changein \right) \right\|_2^2}{\shadw_{i_j}^{[j]2}} 
                    \right) \right) \ldots \\
                     \ldots \frac{1}{{{\shadx}_{i_{j+1}}^{[j]2}}} 
                     \mathop{\sum}\limits_{i_j} \left( {\shadwo}_{i_j,i_{j+1}}^{[j]2} + W_{\changein i_j, i_{j+1}}^{[j]2} \right) 
                     \frac{1}{\shadw_{i_j}^{[j]2}} \frac{\partial}{\partial W_{\changein i'_{j'},i'_{j'+1}}^{[j']}} \left\| {\bg{\Psi}}_{\origin :i_{j}}^{[j-1]} \left( {\bf W}_\changein \right) \right\|_2^2
                     \end{array} &
                     \mbox{if } j > j' \\ \\

             \begin{array}{r} \theta^{(1)} \left( \frac{1}{{{\shadx}_{i_{j+1}}^{[j]2}}} \left( 
                    t_{\changein i_{j+1}}^{[j]} 
                    + 
                    \mathop{\sum}\limits_{i_j} \left( {\shadwo}_{i_j,i_{j+1}}^{[j]2} + W_{\changein i_j, i_{j+1}}^{[j]2} \right) 
                    \frac{\left\| {\bg{\Psi}}_{\origin :i_{j}}^{[j-1]} \left( {\bf W}_\changein \right) \right\|_2^2}{\shadw_{i_j}^{[j]2}} 
                    \right) \right) \ldots \\
                     \ldots \frac{1}{{{\shadx}_{i'_{j+1}}^{[j]2}}} 
                     \delta_{i_{j+1},i'_{j+1}} 2 \left( 1 + \frac{\left\| {\bg{\Psi}}_{\origin :i'_{j}}^{[j-1]} \left( {\bf W}_\changein \right) \right\|_2^2}{\shadw_{i'_j}^{[j]2}} \right) W_{\changein i'_j, i_{j+1}}^{[j]} 
                    \end{array} &
                     \mbox{if } j = j' > 0 \\ \\

             \begin{array}{r} \theta^{(1)} \left(
                    \frac{1}{{{\shadx}_{i_{1}}^{[0]2}}} t_{\changein i_1}^{[0]} 
                  \right) \frac{1}{{{\shadx}_{i'_{1}}^{[0]2}}} \delta_{i_{1},i'_{1}} 4W_{\changein i'_0, i_{1}}^{[0]} \end{array} &
                     \mbox{if } j = j' = 0 \\
   \end{array} \right. \\
 \end{array}
 }}
\label{eq:inducednrecgradw}
\end{equation}
\begin{equation}
 {\!\!\!\!{
 \begin{array}{rl}
  \frac{\partial}{\partial b_{\changein i'_{j'+1}}^{[j']}} \left\| {\bg{\Psi}}_{\origin :i_{j+1}}^{[j]} \left( {\bf W}_\changein \right) \right\|_2^2
  &\!\!\!\!\!=\! \left\{ \!\!\!\begin{array}{ll}
             \begin{array}{r} \!\!\!\theta^{(1)} \!\left( \!\frac{1}{{{\shadx}_{i_{j+1}}^{[j]2}}} \!\left( 
                    t_{\changein i_{j+1}}^{[j]} 
                    +
                    \mathop{\sum}\limits_{i_j} \left( {\shadwo}_{i_j,i_{j+1}}^{[j]2} \!\!+\! W_{\changein i_j, i_{j+1}}^{[j]2} \right) 
                    \frac{\left\| {\bg{\Psi}}_{\origin :i_{j}}^{[j-1]} \left( {\bf W}_\changein \right) \right\|_2^2}{\shadw_{i_j}^{[j]2}} 
                    \!\right) \!\!\right) \ldots \\
                     \ldots \frac{1}{{{\shadx}_{i_{j+1}}^{[j]2}}} 
                     \mathop{\sum}\limits_{i_j} \!\left( {\shadwo}_{i_j,i_{j+1}}^{[j]2} \!\!+\! W_{\changein i_j, i_{j+1}}^{[j]2} \right) 
                     \frac{1}{\shadw_{i_j}^{[j]2}} \frac{\partial}{\partial b_{\changein i'_{j'+1}}^{[j']}} \left\| {\bg{\Psi}}_{\origin :i_{j}}^{[j-1]} \left( {\bf W}_\changein \right) \right\|_2^2
                     \end{array}\!\!\! &
                     \mbox{if } j > j' \\ \\

             \begin{array}{r} \!\!\!\theta^{(1)} \!\left( \!\frac{1}{{{\shadx}_{i_{j+1}}^{[j]2}}} \!\left( 
                    t_{\changein i_{j+1}}^{[j]} 
                    + 
                    \mathop{\sum}\limits_{i_j} \left( {\shadwo}_{i_j,i_{j+1}}^{[j]2} \!\!+\! W_{\changein i_j, i_{j+1}}^{[j]2} \right) 
                    \frac{\left\| {\bg{\Psi}}_{\origin :i_{j}}^{[j-1]} \left( {\bf W}_\changein \right) \right\|_2^2}{\shadw_{i_j}^{[j]2}} 
                    \!\right) \!\!\right) \ldots \\
                     \ldots \frac{1}{{{\shadx}_{i'_{j+1}}^{[j]2}}} 
                     \delta_{i_{j+1},i'_{j+1}} 4b_{\changein i_{j+1}}^{[j]} 
                    \end{array}\!\!\! &
                     \!\!\mbox{if } j = j' > 0 \\ \\

             \begin{array}{r} \!\!\!\theta^{(1)} \!\left(
                    \frac{1}{{{\shadx}_{i_{1}}^{[0]2}}} t_{\changein i_{1}}^{[0]} 
                  \right) \frac{1}{{{\shadx}_{i'_{1}}^{[0]2}}} \delta_{i_{1},i'_{1}} 4b_{\changein i_{1}}^{[0]} \end{array}\!\!\! &
                     \!\!\mbox{if } j = j' = 0 \\
   \end{array} \right. \\
 \end{array}
 }}
\label{eq:inducednrecgradb}
\end{equation}

\subsection{Properties of $\theta$ and $\sigma$} \label{sec:thetasigmaprop}

Here we present an important preliminary property of the $\theta$ and and 
$\sigma_{z,z'}$ functions.  Recall that:
\[
 \begin{array}{rl}
  \theta \left( \zeta \right) 
  &\!\!\!\!= 
  \mathop{\sum}\limits_{l=1}^\infty \zeta^l = \frac{\zeta}{1-\zeta} \;\;\;\; \forall \zeta \in (-1,1) \\

  \sigma^{[j]}_{z,z'} \left( \zeta \right)
  &\!\!\!\!=
  \mathop{\sum}\limits_{l=1}^\infty \left( \frac{1}{l!} \right)^2 \tau^{[j](l)} \left( z \right) \tau^{[j](l)} \left( z' \right) \zeta^l \\
 \end{array}
\]
where:
\[
 \begin{array}{rl}
  \theta^{-1}    \left( \zeta \right) &\!\!\!\!= \frac{\zeta}{1+\zeta} \\
  \theta^{(1)}   \left( \zeta \right) &\!\!\!\!= \frac{1}{(1-\zeta)^2} \\
  \theta^{(1)-1} \left( \zeta \right) &\!\!\!\!= \frac{\sqrt{\zeta}-1}{\sqrt{\zeta}} \\
 \end{array}
\]
We note that, for $z \in \infset{R}$, $j \in \infset{N}_D$:
\begin{itemize}
 \item $\sigma_{z,z}^{[j]} : (-\sqrt{\rho^{[j]}},\sqrt{\rho^{[j]}}) \to \infset{R}$   is, in general, unbounded, smooth and non-Lipchitz.
 \item $\sigma_{z,z}^{[j]} : [0,\sqrt{\rho^{[j]}})                  \to \infset{R}_+$ is, in general, unbounded, smooth, strictly increasing and non-Lipschitz.
 \item $\theta         : (-1,1)     \to \infset{R}$ is unbounded, smooth and non-Lipchitz.
 \item $\theta         : [0,1)      \to [0,\infty)$ is unbounded, smooth, strictly increasing and non-Lipschitz.
 \item $\theta^{-1}    : [0,\infty) \to [0,1)$      is   bounded, smooth, strictly increasing and Lipschitz.
 \item $\theta^{(1)}   : [0,1)      \to [1,\infty)$ is unbounded, smooth, strictly increasing and non-Lipschitz. 
 \item $\theta^{(1)-1} : [1,\infty) \to [0,1)$      is   bounded, smooth, strictly increasing and Lipschitz. 
\end{itemize}
The unbounded, non-Lipschitz nature of $\sigma_{z,z}^{[j]}$ and $\theta$ 
function is problematic when analysing the convergence properties of our 
feature maps.  However if we restrict the domains of $\sigma_{z,z}^{[j]}$ and 
$\theta$ by defining $\psidual^{[j]}, \phidual^{[j]} \in (0,1)$ $\forall j \in 
\infset{N}_D$ and:
\[
 \begin{array}{l}
  \overline{\sigma}^{[j]} \left( \zeta \right) = \mathop{\max}\limits_{z \in \infset{R}_+ \cup \{0\}} \left\{ \sigma_{z,z}^{[j]} \left( \zeta \right) \right\}
 \end{array}
\]
then:
\begin{itemize}
 \item $\overline{\sigma}^{[j]} : [0,(1-\phidual^{[j]}) \sqrt{{\rho}^{[j]}}] 
       \to [0,\frac{\phi^{[j]}}{H^{[j]}}] \subset \infset{R}_+$ is bounded, 
       smooth, strictly increasing and $L_\phi^{[j]}$-Lipschitz, where:
       \[
        \begin{array}{rl}
         \frac{\phi^{[j]}}{H^{[j]}} &\!\!\!\!= \overline{\sigma}^{[j]} \left( (1-\phidual^{[j]}) \sqrt{{\rho}^{[j]}} \right) \\
         L_\phi^{[j]}               &\!\!\!\!= \nabla_{+1} \overline{\sigma}^{[j]} \left( \left( 1-\phidual^{[j]} \right)) \sqrt{{\rho}^{[j]}} \right) \\
        \end{array}
       \]
       and we note that $\frac{\phi^{[j]}}{H^{[j]}} \to \infty$ as $\phidual^{[j]} \to 0$.
 \item $\overline{\sigma}^{[j]-1} : [0,\frac{\phi^{[j]}}{H^{[j]}}] \to [0,(1-\phidual^{[j]}) \sqrt{{\rho}^{[j]}}]$ 
       is bounded, smooth and strictly increasing.
 \item $\theta : [0,1-\psidual^{[j]}] \to [0,\frac{\psi^{[j]}}{H^{[j]}}] \subset \infset{R}_+$ 
       is bounded, smooth, strictly increasing and $L_\psi^{[j]}$-Lipschitz, where:
       \[
        \begin{array}{rl}
         \frac{\psi^{[j]}}{H^{[j]}} &\!\!\!\!= \theta \left( 1 - \psidual^{[j]} \right) = \frac{1-\psidual^{[j]}}{\psidual^{[j]}} \\
         L_\psi^{[j]}               &\!\!\!\!= \frac{\partial}{\partial \zeta} \theta \left( 1 - \psidual^{[j]} \right) = \frac{1}{\psidual^{[j]2}} \\
        \end{array}
       \]
       and we note that $\frac{\psi^{[j]}}{H^{[j]}} \to \infty$ as $\psidual^{[j]} \to 0$.
 \item $\theta^{-1} : [0,\frac{\psi^{[j]}}{H^{[j]}}] \to [0,1-\psidual^{[j]}]$ 
       is bounded, smooth and strictly increasing.
\end{itemize}
This restriction in domain will be used in the following section when we 
analyse the convergence and finiteness of our induced norms.  We also find it 
useful to define the related functions:
\[
 \begin{array}{rl}
  \kappa       \left( \zeta \right) &\!\!\!\!= \zeta \theta^{(1)} \left( \zeta \right) = \frac{\zeta}{\left( 1-\zeta \right)^2} \\
  \kappa^{-1}  \left( \zeta \right) &\!\!\!\!= \frac{2\zeta+1-\sqrt{4\zeta+1}}{2\zeta} \\
  \lambda^{-1} \left( \zeta \right) &\!\!\!\!= \frac{1}{\zeta} \kappa^{-1} \left( \zeta \right) = \frac{2\zeta+1-\sqrt{4\zeta+1}}{2\zeta^2} \\
  \lambda      \left( \zeta \right) &\!\!\!\!= \frac{1-\sqrt{y}}{y} \\
 \end{array}
\]
where we note that:
\begin{itemize}
 \item $\kappa       : [0,1)      \to [0,\infty)$ is strictly increasing. 
 \item $\kappa^{-1}  : [0,\infty) \to [0,1)$      is strictly increasing. 
 \item $\lambda      : (0,1]      \to [0,\infty)$ is strictly decreasing.
 \item $\lambda^{-1} : [0,\infty) \to (0,1]$      is strictly decreasing.
\end{itemize}

\subsection{Convergence Conditions} \label{append:convergecond}

We finish this section by considering the issue of the convergence of the 
induced norms, noting that these results also apply to the induced kernels as:
\[
 \begin{array}{l}
  \det \left( {\bf K}_{\mathcal{X}_\origin} \left( {\bf x}, {\bf x}' \right) \right)
  \leq
  \left( \frac{{\rm Tr} \left( {\bf K}_{\mathcal{X}_\origin} \left( {\bf x}, {\bf x}' \right) \right)}{m} \right)^m
  \leq
  \left( \frac{\left< {\bg{\Phi}}_\origin \left( {\bf x} \right), {\bg{\Phi}}_\origin \left( {\bf x}' \right) \right>_F}{m} \right)^m
  \leq
  \left( \frac{\left\| {\bg{\Phi}}_\origin \left( {\bf x} \right) \right\|_{\mathcal{X}_\origin} \left\| {\bg{\Phi}}_\origin \left( {\bf x}' \right) \right\|_{\mathcal{X}_\origin}}{m} \right)^m \\

  \det \left( {\bf K}_{\mathcal{W}_\origin} \left( {\bf W}_\changein, {\bf W}'_\changein \right) \right)
  \leq
  \left( \frac{{\rm Tr} \left( {\bf K}_{\mathcal{W}_\origin} \left( {\bf W}_\changein, {\bf W}'_\changein \right) \right)}{m} \right)^m
  \leq
  \left( \frac{\left< {\bg{\Psi}}_\origin \left( {\bf W}_\changein \right), {\bg{\Psi}}_\origin \left( {\bf W}'_\changein \right) \right>_F}{m} \right)^m
  \leq
  \left( \frac{\left\| {\bg{\Psi}}_\origin \left( {\bf W}_\changein \right) \right\|_{\mathcal{W}_\origin} \left\| {\bg{\Psi}}_\origin \left( {\bf W}'_\changein \right) \right\|_{\mathcal{W}_\origin}}{m} \right)^m
 \end{array}
\]
and using the positive definiteness of the induced kernels.  
Whereas for the feature representation of ${\bf f}_\changein$ 
it sufficed to ensure that $\| \tilde{\bf x}_{\changein}^{[j]} \|_\infty \leq 
\rho^{[j]}$ for all $i_{j+1}, j$, we cannot use this directly here as we the 
feature map ${\bg{\Phi}}$ does not have access to ${\bf W}_\changein$ (which 
is indeterminant in this context in any case), and likewise the feature map 
${\bg{\Psi}}$ does not have access to ${\bf x}$.

Recall our definition:
\[
 \begin{array}{rl}
 \begin{array}{rl}
  s^{[j]2}
  &\!\!\!\!=
  \left\{ \begin{array}{ll}
   \frac{1}{2} \alpha^{[0]2} + \frac{1}{2} \frac{H^{ [-1]}}{H^{[0]}} M^{ [-1]2} & \mbox{if } j = 0 \\
   \frac{1}{2} \alpha^{[j]2} +             \frac{H^{[j-1]}}{H^{[j]}} M^{[j-1]2} & \mbox{otherwise} \\
  \end{array} \right. \\

  {t}_{\changein i_{j+1}}^{[j]2} 
  &\!\!\!\!= 
  \left\{ \begin{array}{ll}
   \left[ 2b_{\changein i_{  1}}^{[0]2} + 2 \left\| {\bf W}_{\changein :i_{  1}}^{[0]} \right\|_2^2 \right]_{i_{  1}} & \mbox{if } j = 0 \\
   \left[ 2b_{\changein i_{j+1}}^{[j]2} +   \left\| {\bf W}_{\changein :i_{j+1}}^{[j]} \right\|_2^2 \right]_{i_{j+1}} & \mbox{otherwise} \\
  \end{array} \right. \\
 \end{array} & \forall j \in \infset{N}_D, i_{j+1}
 \end{array}
\]
Using this notation, the following theorems present conditions for the 
convergence (finiteness) of the induced norms for all ${\bf x} \in \infset{X}$ 
and given weight-step ${\bf W}_\changein$:
\begin{th_convergephi}
 Let $\phidual^{[j]} \in (0,1)$ $\forall j \in \infset{N}_D$ and for a given 
 neural network and initial weights ${\bf W}_\origin$ define:
 \[
  \begin{array}{l}
   \frac{\phi^{[j]}}{H^{[j]}}
   =
   \overline{\sigma}^{[j]} \left( \left( 1-\phidual^{[j]} \right) \sqrt{{\rho}^{[j]}} \right) 
   \;\;\;\; \forall j \in \infset{N}_D
  \end{array}
 \]
 If the scale factors satisfy:
 \[
  \begin{array}{rll}
                     {\shadx}_{i_{1}}^{[0]2}
                     &\!\!\!\!\leq
                     \frac{1}{\left( \frac{s^{[0]2}}{\left( 1-\phidual^{[0]} \right) \sqrt{{{\rho}}^{[0]}}} \right)} 
                     & \forall i_{1} \\

                     {\shadx}_{i_{j+1}}^{[j]2}
                     &\!\!\!\!\leq
                     \frac{1}{\frac{s^{[j]2}}{\left( 1-\phidual^{[j]} \right) \sqrt{{{\rho}}^{[j]}}}
                     +
                     \frac{1}{H^{[j]}} \sum_{i_j} \shadw_{i_j}^{[j]2} \left( \frac{W_{\origin i_j i_{j+1}}^{[j]2}}{\shadwo_{i_j,i_{j+1}}^{[j]2}} + 1 \right) 
                     \frac{\frac{\phi^{[j-1]}}{H^{[j-1]}}}{\left( 1-\phidual^{[j]} \right) \sqrt{{{\rho}}^{[j]}}}} 
                     & \forall j \in \infset{N}_D \backslash \left\{ 0 \right\}, i_{j+1}
  \end{array}
 \]
 then $\left\| {\bg{\Phi}}_\origin^{[j]} ({\bf x}) \right\|_F^2 \leq 
 \phi^{[j]}$ $\forall {\bf x} \in \infset{X}$, $j \in \infset{N}_D$.
 \label{th:convergephi}
\end{th_convergephi}
\begin{proof}
Recalling (\ref{eq:euceuc_normfindef}), (\ref{eq:euceuc_normsdefof}), we have 
that:
\[
{{
 \begin{array}{l}
  \left\| {{\bg{\Phi}}}_{\origin:i_{j+1}}^{[j]} \left( {\bf x} \right) \right\|_2^2
  = \ldots \\ \ldots = \left\{ \begin{array}{ll}
\sigma^{[j]}_{\!\!{\tilde{x}}_{\origin i_{j+1}}^{[j]}, {\tilde{x}}_{\origin i_{j+1}}^{[j]}} \left( {\shadx}_{i_{j+1}}^{[j]2} \left( 
                    \left( \alpha^{[j]2} + \frac{2}{H^{[j]}} \left\| {\bf x}_{\origin}^{[j]} \right\|_2^2 \right)
                    +
                    \sum_{i_j} \frac{\shadw_{i_j}^{[j]2}}{H^{[j]}} \left( \frac{W_{\origin i_j i_{j+1}}^{[j]2}}{\shadwo_{i_j,i_{j+1}}^{[j]2}} + 1 \right) \left\| {{\bg{\Phi}}}_{\origin:i_{j}}^{[j-1]} \left( {\bf x} \right) \right\|_2^2
             \right) \right) &
                     \mbox{if } j > 0 \\
\sigma^{[0]}_{{\tilde{x}}_{\origin i_{1}}^{[0]}, {\tilde{x}}_{\origin i_{1}}^{[0]}} \left(
                    {\shadx}_{i_{1}}^{[0]2} \left( \alpha^{[0]2} + \frac{1}{H^{[0]}} \left\| {\bf x}_{\origin}^{[0]} \right\|_2^2 \right)
             \right) &
                     \mbox{if } j = 0 \\
   \end{array} \right. \\
 \end{array}
}}
\]
where as discussed in section \ref{sec:thetasigmaprop} on the restricted 
range $\sigma^{[j]}_{z,z} : [0,(1-\phidual^{[j]}) \sqrt{{\rho}^{[j]}}] \to 
[0,\phi^{[j]}]$ is bounded, increasing and Lipschitz for all $z \in 
\infset{R}$.  By our assumptions we have that:
\[
 \begin{array}{l}
  \alpha^{[0]2} + \frac{1}{H^{[0]}} \left\| {\bf x}_{\origin}^{[0]} \right\|_2^2 \leq s^{[0]2} \\
  \alpha^{[j]2} + \frac{2}{H^{[j]}} \left\| {\bf x}_{\origin}^{[j]} \right\|_2^2 \leq s^{[j]2} \;\;\;\; \forall j \in \infset{N}_D \backslash \left\{ 0 \right\} \\
 \end{array}
\]
and hence:
\[
{{
 \begin{array}{rl}
  \left\| {{\bg{\Phi}}}_{\origin:i_{j+1}}^{[j]} \left( {\bf x} \right) \right\|_2^2
  &\!\!\!\!\leq \left\{ \begin{array}{ll}
\sigma^{[j]}_{{\tilde{x}}_{\origin i_{j+1}}^{[j]}, {\tilde{x}}_{\origin i_{j+1}}^{[j]}} \left( {\shadx}_{i_{j+1}}^{[j]2} \left( 
                    s^{[j]2}
                    +
                    \sum_{i_j} \frac{\shadw_{i_j}^{[j]2}}{H^{[j]}} \left( \frac{W_{\origin i_j i_{j+1}}^{[j]2}}{\shadwo_{i_j,i_{j+1}}^{[j]2}} + 1 \right) \left\| {{\bg{\Phi}}}_{\origin:i_{j}}^{[j-1]} \left( {\bf x} \right) \right\|_2^2
             \right) \right) &
                     \mbox{if } j > 0 \\
\sigma^{[0]}_{{\tilde{x}}_{\origin i_{1}}^{[0]}, {\tilde{x}}_{\origin i_{1}}^{[0]}} \left( {\shadx}_{i_{1}}^{[0]2} s^{[0]2} \right) &
                     \mbox{if } j = 0 \\
   \end{array} \right. \\
 \end{array}
}}
\]
We will construct sufficient conditions for convergence by bounding this bound. 
Starting with the input layer, the convergence of $\| {{\bg{\Phi}}}_{\origin 
:i_1}^{[0]} ({\bf x}) \|_2^2$ is assured if the argument of 
$\sigma^{[0]}_{{\tilde{x}}_{\origin i_{1}}^{[0]}, {\tilde{x}}_{\origin 
i_{1}}^{[0]}}$ lies in the restricted ROC, i.e.:
\begin{equation}
 \begin{array}{l}
                    {\shadx}_{i_{1}}^{[0]2} s^{[0]2}
                    \leq
                    \left( 1-\phidual^{[0]} \right) \sqrt{{{\rho}}^{[0]}} \;\;\;\; \forall i_1
 \end{array}
\label{eq:layer0con}
\end{equation}
and moreover if this condition is met then:
\begin{equation}
 \begin{array}{l}
                    \left\| {{\bg{\Phi}}}_{\origin :i_{1}}^{[0]} \left( {\bf x} \right) \right\|_2^2
                    \leq
                    \frac{\phi^{[0]}}{H^{[0]}} \;\;\;\; \forall i_1 
 \end{array}
\label{eq:convbndlay0}
\end{equation}

For layer $j \in \infset{N}_D \backslash \{0\}$, convergence of $\| 
{{\bg{\Phi}}}_{\origin :i_{j+1}}^{[j]} ({\bf x}) \|_2^2$ is assured if the 
argument of $\sigma^{[j]}_{{\tilde{x}}_{\origin i_{j+1}}^{[j]}, 
{\tilde{x}}_{\origin i_{j+1}}^{[j]}}$ lies in the restricted ROC:
\begin{equation}
 \begin{array}{l}
                    {\shadx}_{i_{j+1}}^{[j]2}
                    \left(
                    s^{[j]2}
                    +
                    \sum_{i_j} \frac{\shadw_{i_j}^{[j]2}}{H^{[j]}} \left( \frac{W_{\origin i_j i_{j+1}}^{[j]2}}{\shadwo_{i_j,i_{j+1}}^{[j]2}} + 1 \right) 
                    \left\| {{\bg{\Phi}}}_{\origin:i_{j}}^{[j-1]} \left( {\bf x} \right) \right\|_2^2
                    \right)
                    \leq
                    \left( 1-\phidual^{[j]} \right) \sqrt{{{\rho}}^{[j]}} \;\;\;\; \forall i_{j+1}
 \end{array}
\label{eq:layerjcon}
\end{equation}
and moreover if this condition is met then:
\[
 \begin{array}{l}
                    \left\| {{\bg{\Phi}}}_{\origin :i_{j+1}}^{[j]} \left( {\bf x} \right) \right\|_2^2
                    \leq
                    \frac{\phi^{[j]}}{H^{[j]}} \;\;\;\; \forall i_{j+1}
 \end{array}
\]
Assuming layers $j' < j$ satisfy (\ref{eq:layer0con}), (\ref{eq:layerjcon}) 
then convergence of $\| {{\bg{\Phi}}}_{\origin :i_{j+1}}^{[j]} ({\bf x}) 
\|_2^2$ is assured if:
\[
 \begin{array}{l}
                    {\shadx}_{i_{j+1}}^{[j]2}
                    \left(
                    s^{[j]2}
                    +
                    \sum_{i_j} \frac{\shadw_{i_j}^{[j]2}}{H^{[j]}} \left( \frac{W_{\origin i_j i_{j+1}}^{[j]2}}{\shadwo_{i_j,i_{j+1}}^{[j]2}} + 1 \right) 
                    \frac{\phi^{[j-1]}}{H^{[j-1]}}
                    \right)
                    \leq
                    \left( 1-\phidual^{[j]} \right) \sqrt{{{\rho}}^{[j]}} \;\;\;\; \forall i_{j+1}
 \end{array}
\]
or, sufficiently:
\[
 \begin{array}{l}
                    {\shadx}_{i_{j+1}}^{[j]2}
                    \leq
                    \frac{\left( 1-\phidual^{[j]} \right) \sqrt{{{\rho}}^{[j]}}}{s^{[j]2}
                    +
                    \frac{1}{H^{[j]}} \sum_{i_j} \shadw_{i_j}^{[j]2} \left( \frac{W_{\origin i_j i_{j+1}}^{[j]2}}{\shadwo_{i_j,i_{j+1}}^{[j]2}} + 1 \right) 
                    \frac{\phi^{[j-1]}}{H^{[j-1]}}} \;\;\;\; \forall j \in \infset{N}_D \backslash \left\{ 0 \right\}, i_{j+1}
 \end{array}
\]
which completes the proof.
\end{proof}

\begin{th_convergepsi}
 Let $\psidual^{[j]} \in (0,1)$ $\forall j \in \infset{N}_D$ and for a given 
 neural network and initial weights ${\bf W}_\origin$ define:
 \[
  \begin{array}{rl}
  \begin{array}{rl}
   \frac{\psi^{[j]}}{H^{[j]}}
   =
   \theta \left( 1-\psidual^{[j]} \right)
   =
   \frac{1-\psidual^{[j]}}{\psidual^{[j]}}
  \end{array}
  \end{array}
 \]
 For a weight-step ${\bf W}_\changein$, if:
 \[
  \begin{array}{rl}
   t_{\changein i_{1}}^{[0]2}
   &\!\!\!\!\leq
   \left( 1 - \psidual^{[0]} \right) {\shadx}_{i_{1}}^{[0]2}
   \;\;\;\; \forall i_{1}  \\

   t_{\changein i_{j+1}}^{[j]2} + \sum_{i_j} \frac{\left\| {{\bg{\Psi}}}_{\origin:i_{j}}^{[j-1]} \left( {\bf W}_\changein \right) \right\|_2^2}{\shadw_{i_j}^{[j]2}} \left( {\shadwo}_{i_j,i_{j+1}}^{[j]2} + W_{\changein i_j, i_{j+1}}^{[j]2} \right)
   &\!\!\!\!\leq
   \left( 1 - \psidual^{[j]} \right) {\shadx}_{i_{j+1}}^{[j]2}
   \;\;\;\; \forall j \in \infset{N}_D \backslash \left\{ 0 \right\}, i_{j+1} \\
  \end{array}
 \]
 then $\left\| {{\bg{\Psi}}}_\origin^{[j]} ({\bf W}_\changein) 
 \right\|_F^2 \leq \psi^{[j]}$ for all $j \in \infset{N}_D$.  
 \label{th:convergepsi}
\end{th_convergepsi}
\begin{proof}
Recalling (\ref{eq:euceuc_normfindef}), (\ref{eq:euceuc_normsdefof}), we have 
that:
\[
 {\!\!\!\!\!\!\!\!\!\!\!\!\!\!\!\!\!\!\!\!\!\!\!\!\!\!\!\!{
 \begin{array}{rl}
  \left\| {{\bg{\Psi}}}_{\origin:i_{j+1}}^{[j]} \left( {\bf W}_\changein \right) \right\|_2^2
  &\!\!\!\!= \left\{ \begin{array}{ll}
             \theta \left( \frac{1}{{\shadx}_{i_{j+1}}^{[j]2}} \left(
                    t_{\changein i_{j+1}}^{[j]2}
                    +
                    \sum_{i_j} \frac{1}{\shadw_{i_j}^{[j]2}} \left( {\shadwo}_{i_j,i_{j+1}}^{[j]2} + W_{\changein i_j, i_{j+1}}^{[j]2} \right) \left\| {{\bg{\Psi}}}_{\origin:i_{j}}^{[j-1]} \left( {\bf W}_\changein \right) \right\|_2^2
                  \right) \right) &
                     \mbox{if } j > 0 \\
             \theta \left(
                    \frac{1}{\shadx_{i_1}^{[0]2}} t_{\changein i_{1}}^{[0]2}
                  \right) &
                     \mbox{if } j = 0 \\
   \end{array} \right. \\
 \end{array}
 }}
\]
where as discussed in section \ref{sec:thetasigmaprop} on the restricted 
range $\theta : [0,(1-\psidual^{[j]})] \to [0,\psi^{[j]}]$ is bounded, 
increasing and $\frac{1}{\psidual^{[j]2}}$-Lipschitz.  Starting at layer 
$0$ we see that if:
\[
 \begin{array}{rl}
  \frac{t_{\changein i_{1}}^{[0]2}}{\shadx_{i_{1}}^{[0]2}} 
  &\!\!\!\!\leq 
  1 - \psidual^{[0]}
 \end{array}
\]
Then:
\[
 \begin{array}{rll}
  \left\| {{\bg{\Psi}}}_{\origin:i_{1}}^{[0]} \left( {\bf W}_\changein \right) \right\|_2^2
  &\!\!\!\!\leq
  \frac{\psi^{[0]}}{H^{[0]}}
  &\forall i_{1} \\
 \end{array}
\]
Moreover for layer $j$, if:
\[
 \begin{array}{rll}
             \frac{1}{{\shadx}_{i_{j+1}}^{[j]2}} \left(
                    t_{\changein i_{j+1}}^{[j]2}
                    +
                    \sum_{i_j} \frac{\left\| {{\bg{\Psi}}}_{\origin:i_{j}}^{[j-1]} \left( {\bf W}_\changein \right) \right\|_2^2}{\shadw_{i_j}^{[j]2}} \left( {\shadwo}_{i_j,i_{j+1}}^{[j]2} + W_{\changein i_j, i_{j+1}}^{[j]2} \right)
                  \right)
  &\!\!\!\!\leq
  1 - \psidual^{[j]}
  &\forall i_{j+1} \\
 \end{array}
\]
Then:
\[
 \begin{array}{rll}
  \left\| {{\bg{\Psi}}}_{\origin:i_{2}}^{[1]} \left( {\bf W}_\changein \right) \right\|_2^2
  &\!\!\!\!\leq
  \frac{\psi^{[j]}}{H^{[j]}}
  &\forall i_{2} \\
 \end{array}
\]
which gives our sufficient conditions in the general case:
\[
 \begin{array}{rll}
  {\shadx}_{i_{1}}^{[0]2}
  &\!\!\!\!\geq
             \frac{t_{\changein i_{1}}^{[0]2}}{1 - \psidual^{[0]}}
  &\forall i_{1} \\

  {\shadx}_{i_{j+1}}^{[j]2}
  &\!\!\!\!\geq
             \frac{1}{1 - \psidual^{[j]}} \left(
                    t_{\changein i_{j+1}}^{[j]2}
                    +
                    \sum_{i_j} \frac{\left\| {{\bg{\Psi}}}_{\origin:i_{j}}^{[j-1]} \left( {\bf W}_\changein \right) \right\|_2^2}{\shadw_{i_j}^{[j]2}} \left( {\shadwo}_{i_j,i_{j+1}}^{[j]2} + W_{\changein i_j, i_{j+1}}^{[j]2} \right)
                  \right)
  &\forall j \in \infset{N}_D \backslash \left\{ 0 \right\}, i_{j+1} \\
 \end{array}
\]
which completes the proof.
\end{proof}

Note that this theorem may be equivalently stated using a bound on the 
scale factors:
 \[
  \begin{array}{rl}
   {\shadx}_{i_{1}}^{[0]2}
   &\!\!\!\!\geq
   \frac{1}{1 - \psidual^{[0]}} t_{\changein i_{1}}^{[0]2}
   \;\;\;\; \forall i_{1}  \\

   {\shadx}_{i_{j+1}}^{[j]2}
   &\!\!\!\!\geq
   \frac{1}{1 - \psidual^{[j]}} \left( t_{\changein i_{j+1}}^{[j]2} + \sum_{i_j} \frac{\left\| {{\bg{\Psi}}}_{\origin:i_{j}}^{[j-1]} \left( {\bf W}_\changein \right) \right\|_2^2}{\shadw_{i_j}^{[j]2}} \left( {\shadwo}_{i_j,i_{j+1}}^{[j]2} + W_{\changein i_j, i_{j+1}}^{[j]2} \right) \right)
   \;\;\;\; \forall j \in \infset{N}_D \backslash \left\{ 0 \right\}, i_{j+1} \\
  \end{array}
 \]
rather than as a bound on the weight-step.  Thus theorems \ref{th:convergephi} 
and \ref{th:convergepsi} give some insight into the role of the scale factors 
in our scheme.  Loosely speaking, the scale factors must be chosen as a 
tradeoff between the convergence of the feature maps ${\bg{\Phi}}_\origin$ and 
${\bg{\Psi}}_\origin$, and these theorems how this trade-off may be tuned, and 
to what extent.  In particular:
\begin{itemize}
 \item
 The upper bounds on the scale factors given in theorem \ref{th:convergephi} 
 get larger (less strict) as the ``size'' of $\infset{X}$ and and the range 
 of outputs of all layers of the network, as measured by $s^{[j]}$, gets 
 smaller, diverging to $\infty$ (unbounded) in the limit $s^{[j]} \to 
 0^+$.  Of course the factors $s^{[j]}$ are determined by the structure of the 
 network, the range of the dataset $\infset{X}$ and our choice of $\alpha$, so 
 in a practice this upper bound on the scale factors is fixed.
 \item
 The lower bounds on the scale factors given in theorem \ref{th:convergepsi} 
 get smaller (less strict) as the size of the weight-step, plus a factor 
 dependent on ${{\shadwo}}_{i_j,i_{j+1}}^{[j]}$, gets smaller, decreasing to $0$ 
 (unbounded) as these go to zero.  Thus we see that the scale factors are 
 effectively bounded below by the convergence conditions on the feature map 
 ${\bg{\Psi}}_{\origin}$, with the bound being determined by the size of the 
 weight-step.  Unlike the upper bound given in theorem \ref{th:convergephi}, 
 we can make the lower bound arbitrarily small by requiring that the weight-step 
 and offset be sufficiently small.
\end{itemize}
The interaction of these two observations - the upper and lower bounds on 
the scale factors - determines what weight-steps we can model over the set 
of all inputs $\infset{X}$.  Furthermore, the presence of the scale factors 
in the lower bounds in theorem \ref{th:convergepsi}, gives an indication of how 
they may be chosen.  To be precise, for the bounds to make sense we need that:
\begin{itemize}
 \item The shadow weights $\shadwo_{i_j,i_{j+1}}^{[j]}$ must be of the same (or lower) 
       order (scale) as the weight-step to ensure that, for a sufficiently small 
       weight-step, the lower bound on the scale factors becomes lower than the 
       upper bound on the scale factors.
 \item The shadow weights $\shadw_{i_j}^{[j]}$ must be of the same (or greater) 
       order (scale) as the norm $\| {{\bg{\Psi}}}_{\origin:i_{j}}^{[j-1]} ({\bf 
       W}_\changein) \|_2^2$ to cancel out its influence on the lower bound on 
       the scale factors.
\end{itemize}
We make this intuition concrete by combining these theorems to obtain sufficient 
conditions on the size of the weight-step, the shadow weights and the scale 
factors to ensure that both ${\bg{\Phi}}_\origin$ and ${\bg{\Psi}}_\origin$ are 
convergent for all ${\bf x} \in \infset{X}$:
\begin{th_convergephipsi}
 Let $\psidualscale, \phidual^{[j]}, \psidual \in (0,1)$ $\forall j \in 
 \infset{N}_D$ and for a given neural network and initial weights ${\bf W}_\origin$ 
 define:
 \[

 \]
 recursively, and subsequently:
 \[
  %
 \]
 For a weight-step ${\bf W}_\changein$, if the shadow weights satisfy:
 \[
  \begin{array}{rll}
                    \shadw_{i_j}^{[j]2} 
                    &\!\!\!\!\in \left[ \left\| {{\bg{\Psi}}}_{\origin:i_{j}}^{[j-1]} \left( {\bf W}_\changein \right) \right\|_2^2, 
                    \frac{\psi^{[j-1]}}{H^{[j-1]}} \right] 
                    & \forall j \in \infset{N}_D, i_j \\

                     \left\| {\bg{\shadwo}}_{:i_{j+1}}^{[j]} \right\|_\infty^2
                     &\!\!\!\!\leq \psidualscale
                     \frac{1-\psidual^{[j]}}
                     {H^{[j-1]} \left( \frac{s^{[j]2}}{\left( 1-\phidual^{[j]} \right) \sqrt{{{\rho}}^{[j]}}}
                     +
                     \frac{\left( 1 - \psidualscale \right) \left( 1-\psidual^{[j]} \right)}{\left\| {\bf W}_{\origin}^{[j]} \right\|_{2,\infty}^2}
                     \right)}
                    & \forall j \in \infset{N}_D \backslash \left\{ 0 \right\}, i_{j+1} \\
  \end{array}
 \]
 and the weight-step satisfies:
 \[
  \begin{array}{rll}
                     t_{\changein i_{1}}^{[0]2}
                     &\!\!\!\!\leq
                     \frac{1-\psidual^{[0]}}{\left( \frac{s^{[0]2}}{\left( 1-\phidual^{[0]} \right) \sqrt{{{\rho}}^{[0]}}} \right)} 
                     & \forall i_{1} \\

                     t_{\changein i_{j\!+\!1}}^{[j]2}
                     +
                     \sum_{i_j} \frac{\left\| {{\bg{\Psi}}}_{\origin:i_{j}}^{[j-1]} \left( {\bf W}_\changein \right) \right\|_2^2}{\shadw_{i_j}^{[j]2}} 
                     \left( {\shadwo}_{i_j,i_{j\!+\!1}}^{[j]2} + W_{\changein i_j, i_{j\!+\!1}}^{[j]2} \right)
                     &\!\!\!\!\leq
                     \frac{1-\psidual^{[j]}}{\frac{s^{[j]2}}{\left( 1-\phidual^{[j]} \right) \sqrt{{{\rho}}^{[j]}}}
                     +
                     \left( \frac{\left\| {\bf W}_{\origin :i_{j\!+\!1}}^{[j]} \right\|_2^2}{H^{[j-1]} \left\| {\bg{\shadwo}}_{:i_{j\!+\!1}}^{[j]} \right\|_{-\infty}^2} + 1 \right) 
                     \frac{\left( 1 - \psidualscale \right) \left( 1-\psidual^{[j]} \right)}{\left\| {\bf W}_{\origin}^{[j]} \right\|_{2,\infty}^2}
                     } 
                     & \forall j \in \infset{N}_D \backslash \left\{ 0 \right\}, i_{j\!+\!1} \\
  \end{array}
 \]
 and the scale factors satisfy:
 \[
 {\!\!\!{
  \begin{array}{rll}
                     {\shadx}_{i_{1}}^{[0]2}
                     &\!\!\!\!\!\in\!
                     \left[
                     \frac{1}{1-\psidual^{[0]}} t_{\changein i_{1}}^{[0]2}, 
                     \frac{1}{\left( \frac{s^{[0]2}}{\left( 1-\phidual^{[0]} \right) \sqrt{{{\rho}}^{[0]}}} \right)} 
                     \right]
                     & \forall i_{1} \\

                     {\shadx}_{i_{j\!+\!1}}^{[j]2}
                     &\!\!\!\!\!\in\!
                     \left[
                     \frac{1}{1\!-\!\psidual^{[j]}} \!\left( \!t_{\changein i_{j\!+\!1}}^{[j]2} \!\!+\! \mathop{\sum}\limits_{i_j} \frac{\left\| {{\bg{\Psi}}}_{\origin:i_{j}}^{[j\!-\!1]} \!\left(\! {\bf W}_\changein \right) \right\|_2^2}{\shadw_{i_j}^{[j]2}} \!\left( \!{\shadwo}_{i_j,i_{j\!+\!1}}^{[j]2} \!+\! W_{\changein i_j, i_{j\!+\!1}}^{[j]2} \!\right) \!\!\right)\!\!, 
                     \frac{1}{\frac{s^{[j]2}}{\left( \!1\!-\!\phidual^{[j]} \!\right) \sqrt{\!{{\rho}}^{[j]}}}
                     \!+\!
                     \left( \!\frac{\left\| {\bf W}_{\origin :i_{j\!+\!1}}^{[j]} \right\|_2^2}{H^{[j\!-\!1]} \left\| {\bg{\shadwo}}_{:i_{j\!+\!1}}^{[j]} \right\|_{-\infty}^2} \!+\! 1 \!\right) 
                     \!\frac{\left( \!1 \!-\! \psidualscale \right) \left( \!1\!-\!\psidual^{[j]} \!\right)}{\left\| {\bf W}_{\origin}^{[j]} \right\|_{2,\infty}^2}
                     } 
                     \right]\!\!\!\!\!
                     & \forall j \in \infset{N}_D \backslash \left\{ 0 \right\}, i_{j\!+\!1}
  \end{array}
  }}
 \]
 then $\left\| {\bg{\Phi}}_\origin^{[j]} ({\bf x}) \right\|_F^2 \leq 
 \phi^{[j]}$ $\forall {\bf x} \in \infset{X}$, $j \in \infset{N}_D$ and 
 $\left\| {{\bg{\Psi}}}_\origin^{[j]} ({\bf W}_\changein) \right\|_F^2 \leq 
 \psi^{[j]}$ for all $j \in \infset{N}_D$.
 \label{th:convergephipsi}
\end{th_convergephipsi}
\begin{proof}
Combining theorems \ref{th:convergephi} and \ref{th:convergepsi}, we 
see that for convergence in both ${\bg{\Phi}}_\origin$, ${\bg{\Psi}}_\origin$ 
we require that:
 \[
  \begin{array}{rll}
                     t_{\changein i_{1}}^{[0]2}
                     &\!\!\!\!\leq
                     \frac{\left( 1-\psidual^{[0]} \right) \left( 1-\phidual^{[0]} \right) \sqrt{{{\rho}}^{[0]}}}{s^{[0]2}} 
                     & \forall i_{1} \\

                     t_{\changein i_{j+1}}^{[j]2}
                     +
                     \sum_{i_j} \frac{\left\| {{\bg{\Psi}}}_{\origin:i_{j}}^{[j-1]} \left( {\bf W}_\changein \right) \right\|_2^2}{\shadw_{i_j}^{[j]2}} 
                     \left( {\shadwo}_{i_j,i_{j+1}}^{[j]2} + W_{\changein i_j, i_{j+1}}^{[j]2} \right)
                     &\!\!\!\!\leq
                     \frac{\left( 1-\psidual^{[j]} \right) \left( 1-\phidual^{[j]} \right) \sqrt{{{\rho}}^{[j]}}}{s^{[j]2}
                     +
                     \frac{1}{H^{[j]}} \sum_{i_j} \shadw_{i_j}^{[j]2} \left( \frac{W_{\origin i_j i_{j+1}}^{[j]2}}{\shadwo_{i_j,i_{j+1}}^{[j]2}} + 1 \right) 
                     \frac{\phi^{[j-1]}}{H^{[j-1]}}} 
                     & \forall j \in \infset{N}_D \backslash \left\{ 0 \right\}, i_{j+1}
  \end{array}
 \]
Using our condition $\shadw_{i_j}^{[j]2} \in \left[ \| {{\bg{\Psi}}}_{\origin:i_{j}}^{[j-1]} ({\bf W}_\changein) \|_2^2, 
\frac{\psi^{[j-1]}}{H^{[j-1]}} \right]$ then it suffices that:
\[
 \begin{array}{rll}
                    t_{\changein i_{1}}^{[0]2}
                    &\!\!\!\!\leq
                    \frac{\left( 1-\psidual^{[0]} \right) \left( 1-\phidual^{[0]} \right) \sqrt{{{\rho}}^{[0]}}}{s^{[0]2}} 
                    & \forall i_{1} \\

                    t_{\changein i_{j+1}}^{[j]2}
                    +
                    H^{[j-1]} \left\| {\bg{\shadwo}}_{:i_{j+1}}^{[j]} \right\|_\infty^2 + \left\| {\bf W}_{\changein :i_{j+1}}^{[j]} \right\|_2^2
                    &\!\!\!\!\leq
                    \frac{\left( 1-\psidual^{[j]} \right) \left( 1-\phidual^{[j]} \right) \sqrt{{{\rho}}^{[j]}}}{s^{[j]2}
                    +
                    \frac{1}{H^{[j]}} \left( \sum_{i_j} \frac{W_{\origin i_j i_{j+1}}^{[j]2}}{\shadwo_{i_j,i_{j+1}}^{[j]2}} + H^{[j-1]} \right) 
                    \frac{\psi^{[j-1]}}{H^{[j-1]}} \frac{\phi^{[j-1]}}{H^{[j-1]}}} 
                    & \forall j \in \infset{N}_D \backslash \left\{ 0 \right\}, i_{j+1}
 \end{array}
\]
We require that this may be satisfied for a sufficiently small weight step, 
so $\forall j \in \infset{N}_D \backslash \left\{ 0 \right\}, i_{j+1}$:
\[
 \begin{array}{rl}
  &
                    H^{[j-1]} \left\| {\bg{\shadwo}}_{:i_{j+1}}^{[j]} \right\|_\infty^2 
                    \leq
                    \frac{\left( 1-\psidual^{[j]} \right) \left( 1-\phidual^{[j]} \right) \sqrt{{{\rho}}^{[j]}}}{s^{[j]2}
                    +
                    \frac{1}{H^{[j]}} \left( \sum_{i_j} \frac{W_{\origin i_j i_{j+1}}^{[j]2}}{\shadwo_{i_j,i_{j+1}}^{[j]2}} + H^{[j-1]} \right) 
                    \frac{\psi^{[j-1]}}{H^{[j-1]}} \frac{\phi^{[j-1]}}{H^{[j-1]}}} \\

  \Rightarrow &
                    H^{[j-1]}  
                    \leq
                    \frac{\left( 1-\psidual^{[j]} \right) \left( 1-\phidual^{[j]} \right) \sqrt{{{\rho}}^{[j]}}}{
                    s^{[j]2} \left\| {\bg{\shadwo}}_{:i_{j+1}}^{[j]} \right\|_\infty^2
                    +
                    \frac{1}{H^{[j]}} \left( \frac{1}{\chi_{i_{j+1}}^{[j]}} \left\| {\bf W}_{\origin :i_{j+1}}^{[j]} \right\|_2^2 + H^{[j-1]} \left\| {\bg{\shadwo}}_{:i_{j+1}}^{[j]} \right\|_\infty^2 \right) 
                    \frac{\psi^{[j-1]}}{H^{[j-1]}} \frac{\phi^{[j-1]}}{H^{[j-1]}}} \\

  \Rightarrow &
                    H^{[j\!-\!1]} s^{[j]2} \left\| {\bg{\shadwo}}_{:i_{j\!+\!1}}^{[j]} \right\|_\infty^2
                    \!\!+\!
                    \frac{H^{[j\!-\!1]}}{H^{[j]}} \!\left( \!\frac{1}{\chi_{i_{j\!+\!1}}^{[j]}} \left\| {\bf W}_{\origin :i_{j\!+\!1}}^{[j]} \right\|_2^2 + H^{[j\!-\!1]} \left\| {\bg{\shadwo}}_{:i_{j\!+\!1}}^{[j]} \right\|_\infty^2 \right) 
                    \frac{\psi^{[j\!-\!1]}}{H^{[j\!-\!1]}} \frac{\phi^{[j\!-\!1]}}{H^{[j\!-\!1]}}
                    \leq
                    \left( 1-\psidual^{[j]} \right) \left( 1-\phidual^{[j]} \right) \sqrt{{{\rho}}^{[j]}} \\
\end{array}
\]
where:
\[
 \begin{array}{rl}
  \chi_{i_{j+1}}^{[j]}
  &\!\!\!\!=
  \frac{\left\| {\bg{\shadwo}}_{:i_{j+1}}^{[j]} \right\|_{-\infty}^2}{\left\| {\bg{\shadwo}}_{:i_{j+1}}^{[j]} \right\|_{\infty}^2}
 \end{array}
\]
To be well-defined, we require that the right-hand-side is positive, for which it suffices that:
\[
 \begin{array}{rl}
  \frac{\psi^{[j-1]}}{H^{[j-1]}} 
  =
  \left( 1 - \psidualscale \right) \left( 1-\psidual^{[j]} \right) \left( 1-\phidual^{[j]} \right) \sqrt{{{\rho}}^{[j]}} 
  \frac{H^{[j]}}{H^{[j-1]}}
  \frac{\chi_{i_{j+1}}^{[j]}}{\left\| {\bf W}_{\origin}^{[j]} \right\|_{2,\infty}^2}
  \frac{H^{[j-1]}}{\phi^{[j-1]}}
 \end{array}
\]
which we may rewrite in terms of $\psidual^{[j-1]}$:
\[
 \begin{array}{rl}
  \psidual^{[j-1]}
  &\!\!\!\!=
  \theta^{-1} \left( \left( 1 - \psidualscale \right) 
  \left( 1-\psidual^{[j]} \right) \left( 1-\phidual^{[j]} \right) \sqrt{{{\rho}}^{[j]}} 
  \frac{H^{[j]}}{H^{[j-1]}}
  \frac{\chi_{i_{j+1}}^{[j]}}{\left\| {\bf W}_{\origin}^{[j]} \right\|_{2,\infty}^2}
  \frac{H^{[j-1]}}{\phi^{[j-1]}}
  \right) \\
  &\!\!\!\!=
  \frac{\left( 1 - \psidualscale \right) \left( 1-\psidual^{[j]} \right) \left( 1-\phidual^{[j]} \right) \sqrt{{{\rho}}^{[j]}}}
  {\frac{H^{[j-1]}}{H^{[j]}}
  \frac{\left\| {\bf W}_{\origin}^{[j]} \right\|_{2,\infty}^2}{\chi_{i_{j+1}}^{[j]}}
  \frac{\phi^{[j-1]}}{H^{[j-1]}}+\left( 1 - \psidualscale \right) \left( 1-\psidual^{[j]} \right) \left( 1-\phidual^{[j]} \right) \sqrt{{{\rho}}^{[j]}}}
 \end{array}
\]
in which case our sufficient conditions become:
\[
 \begin{array}{rl}
                    \left\| {\bg{\shadwo}}_{:i_{j+1}}^{[j]} \right\|_\infty^2
                    \leq \psidualscale
                    \frac{\left( 1-\psidual^{[j]} \right)}
                    {H^{[j-1]} \left( \frac{s^{[j]2}}{\left( 1-\phidual^{[j]} \right) \sqrt{{{\rho}}^{[j]}}}
                    +
                    \left( 1 - \psidualscale \right) \left( 1-\psidual^{[j]} \right) 
                    \frac{\chi_{i_{j+1}}^{[j]}}{\left\| {\bf W}_{\origin}^{[j]} \right\|_{2,\infty}^2}
                    \right)} \\
 \end{array}
\]
and:
 \[
  \begin{array}{rll}
                     t_{\changein i_{1}}^{[0]2}
                     &\!\!\!\!\leq
                     \frac{\left( 1-\psidual^{[0]} \right)}{\left( \frac{s^{[0]2}}{\left( 1-\phidual^{[0]} \right) \sqrt{{{\rho}}^{[0]}}} \right)} 
                     & \forall i_{1} \\

                     t_{\changein i_{j\!+\!1}}^{[j]2}
                     +
                     \mathop{\sum}\limits_{i_j} \frac{\left\| {{\bg{\Psi}}}_{\origin:i_{j}}^{[j\!-\!1]} \left( {\bf W}_\changein \right) \right\|_2^2}{\shadw_{i_j}^{[j]2}} 
                     \!\left( {\shadwo}_{i_j,i_{j\!+\!1}}^{[j]2} \!+\! W_{\changein i_j, i_{j\!+\!1}}^{[j]2} \right)
                     &\!\!\!\!\leq
                     \frac{\left( 1-\psidual^{[j]} \right)}{\frac{s^{[j]2}}{\left( 1-\phidual^{[j]} \right) \sqrt{{{\rho}}^{[j]}}}
                     +
                     \left( \frac{\left\| {\bf W}_{\origin :i_{j\!+\!1}}^{[j]} \right\|_2^2}{H^{[j\!-\!1]} \left\| {\bg{\shadwo}}_{:i_{j\!+\!1}}^{[j]} \right\|_{-\infty}^2} + 1 \right) 
                     \frac{\left( 1 - \psidualscale \right) \left( 1-\psidual^{[j]} \right)  \chi_{i_{j\!+\!1}}^{[j]}}{\left\| {\bf W}_{\origin}^{[j]} \right\|_{2,\infty}^2}
                     } 
                     & \forall j \in \infset{N}_D \backslash \left\{ 0 \right\}, i_{j\!+\!1} \\
  \end{array}
 \]
In terms of the scale factors, recalling theorems \ref{th:convergephi} and 
\ref{th:convergepsi}, we require that (sufficiently):
 \[
 {\!\!\!\!{
  %
 }}
 \]
Noting that $\chi_{i_{j+1}}^{[j]} \leq 1$, we can further tighten our 
constraints to get sufficient conditions:
\[
 \begin{array}{rl}
                    \left\| {\bg{\shadwo}}_{:i_{j+1}}^{[j]} \right\|_\infty^2
                    \leq \psidualscale
                    \frac{1-\psidual^{[j]}}
                    {H^{[j-1]} \left( \frac{s^{[j]2}}{\left( 1-\phidual^{[j]} \right) \sqrt{{{\rho}}^{[j]}}}
                    +
                    \frac{\left( 1 - \psidualscale \right) \left( 1-\psidual^{[j]} \right)}{\left\| {\bf W}_{\origin}^{[j]} \right\|_{2,\infty}^2}
                    \right)} \\
 \end{array}
\]
 \[
 {\!\!\!{
  \begin{array}{rll}
                     t_{\changein i_{1}}^{[0]2}
                     &\!\!\!\!\leq
                     \frac{1-\psidual^{[0]}}{\left( \frac{s^{[0]2}}{\left( 1-\phidual^{[0]} \right) \sqrt{{{\rho}}^{[0]}}} \right)} 
                     & \forall i_{1} \\

                     t_{\changein i_{j+1}}^{[j]2}
                     +
                     \mathop{\sum}\limits_{i_j} \frac{\left\| {{\bg{\Psi}}}_{\origin:i_{j}}^{[j-1]} \left( {\bf W}_\changein \right) \right\|_2^2}{\shadw_{i_j}^{[j]2}} 
                     \left( {\shadwo}_{i_j,i_{j+1}}^{[j]2} + W_{\changein i_j, i_{j+1}}^{[j]2} \right)
                     &\!\!\!\!\leq
                     \frac{1-\psidual^{[j]}}{\frac{s^{[j]2}}{\left( 1-\phidual^{[j]} \right) \sqrt{{{\rho}}^{[j]}}}
                     +
                     \left( \frac{\left\| {\bf W}_{\origin :i_{j+1}}^{[j]} \right\|_2^2}{H^{[j-1]} \left\| {\bg{\shadwo}}_{:i_{j+1}}^{[j]} \right\|_{-\infty}^2} + 1 \right) 
                     \frac{\left( 1 - \psidualscale \right) \left( 1-\psidual^{[j]} \right)}{\left\| {\bf W}_{\origin}^{[j]} \right\|_{2,\infty}^2}
                     } 
                     & \forall j \in \infset{N}_D \backslash \left\{ 0 \right\}, i_{j+1} \\
  \end{array}
 }}
 \]
 \[
 {\!\!\!{
  \begin{array}{rll}
                     {\shadx}_{i_{1}}^{[0]2}
                     &\!\!\!\!\!\in\!
                     \left[
                     \frac{1}{1-\psidual^{[0]}} t_{\changein i_{1}}^{[0]2}, 
                     \frac{1}{\left( \frac{s^{[0]2}}{\left( 1-\phidual^{[0]} \right) \sqrt{{{\rho}}^{[0]}}} \right)} 
                     \right]
                     & \forall i_{1} \\

                     {\shadx}_{i_{j\!+\!1}}^{[j]2}
                     &\!\!\!\!\!\in\!
                     \left[
                     \frac{1}{1\!-\!\psidual^{[j]}} \!\left( \!t_{\changein i_{j\!+\!1}}^{[j]2} \!\!+\! \mathop{\sum}\limits_{i_j} \!\frac{\left\| {{\bg{\Psi}}}_{\origin:i_{j}}^{[j\!-\!1]} \!\left( \!{\bf W}_\changein \right) \right\|_2^2}{\shadw_{i_j}^{[j]2}} \left( {\shadwo}_{i_j,i_{j\!+\!1}}^{[j]2} \!\!+\! W_{\changein i_j, i_{j\!+\!1}}^{[j]2} \!\right) \!\!\right), 
                     \frac{1}{\frac{s^{[j]2}}{\left( \!1\!-\!\phidual^{[j]}\! \right) \sqrt{\!{{\rho}}^{[j]}}}
                     \!+\!
                     \left( \!\frac{\left\| {\bf W}_{\origin :i_{j\!+\!1}}^{[j]} \right\|_2^2}{H^{[j\!-\!1]} \left\| {\bg{\shadwo}}_{:i_{j\!+\!1}}^{[j]} \right\|_{-\infty}^2} + \!1\! \right) 
                     \!\frac{\left( \!1\!-\!\psidualscale\! \right) \left( \!1\!-\!\psidual^{[j]}\! \right)}{\left\| {\bf W}_{\origin}^{[j]} \right\|_{2,\infty}^2}
                     } 
                     \right]
                     & \forall j \!\in\! \infset{N}_D \!\backslash\! \left\{ 0 \right\}\!, i_{j\!+\!1}
  \end{array}
 }}
 \]
which completes the proof.
\end{proof}

We finish this section by considering a special case that will be important 
shortly.
\begin{th_convergephipsi_spec}
 Let $\phidual^{[j]}, \psidual^{[j]}, \wtune \in (0,1)$ $\forall j \in 
 \infset{N}_D$ and for a given neural network with initial weights ${\bf 
 W}_\origin$ and weight-step ${\bf W}_\changein$, using the shadow weights:
 \[
  %
 \]
 If the weight-step satisfies:
 \[
  \begin{array}{rll}
                     t_{\changein i_{1}}^{[0]2}
                     &\!\!\!\!\leq
                     \frac{\left( 1-\psidual^{[0]} \right)}{\left( \frac{s^{[0]2}}{\left( 1-\phidual^{[0]} \right) \sqrt{{{\rho}}^{[0]}}} \right)} 
                     & \forall i_{1} \\

                     t_{\changein i_{j+1}}^{[j]2} + \frac{1}{1-\wtune} \left\| {\bf W}_{\changein :i_{j+1}}^{[j]} \right\|_2^2
                     &\!\!\!\!\leq
                     \frac{\left( 1-\psidual^{[j]} \right)}{\frac{s^{[j]2}}{\left( 1-\phidual^{[j]} \right) \sqrt{{{\rho}}^{[j]}}}
                     +
                     \frac{1}{H^{[j]}} \left( \frac{\left\| {\bf W}_{\origin :i_{j+1}}^{[j]} \right\|_2^2}{
\frac{1}{1-\wtune} \left\| {\bf W}^{[j]}_{\changein :i_{j+1}} \right\|_2^2 - \left\| {\bf W}^{[j]}_{\changein :i_{j+1}} \right\|_\infty^2
} + H^{[j-1]} \right) 
                     \frac{\frac{\psi^{[j-1]}}{H^{[j-1]}} \frac{\phi^{[j-1]}}{H^{[j-1]}}}{\left( 1-\phidual^{[j]} \right) \sqrt{{{\rho}}^{[j]}}}} 
                     & \forall j \in \infset{N}_D \backslash \left\{ 0 \right\}, i_{j+1} \\
  \end{array}
 \]
 and the scale factors satisfy:
 \[
 {\!\!\!{
  %
  \end{array}
 }}
 \]
 then $\left\| {\bg{\Phi}}_\origin^{[j]} ({\bf x}) \right\|_F^2 \leq 
 \phi^{[j]}$ $\forall {\bf x} \in \infset{X}$, $j \in \infset{N}_D$ and 
 $\left\| {{\bg{\Psi}}}_\origin^{[j]} ({\bf W}_\changein) \right\|_F^2 \leq 
 \psi^{[j]}$ for all $j \in \infset{N}_D$.
 \label{th:convergephipsi_spec}
\end{th_convergephipsi_spec}
\begin{proof}
Our proof is a simple analogue of the proof of theorem ref{th:convergephipsi} 
with adjustments for our pre-defined scale factors.  Combining theorems 
\ref{th:convergephi} and \ref{th:convergepsi}, we see that for convergence in 
both ${\bg{\Phi}}_\origin$, ${\bg{\Psi}}_\origin$ we require that, using our 
definitions of the scale factors:
\[
 \begin{array}{rll}
                    t_{\changein i_{1}}^{[0]2}
                    &\!\!\!\!\leq
                    \frac{\left( 1-\psidual^{[0]} \right) \left( 1-\phidual^{[0]} \right) \sqrt{{{\rho}}^{[0]}}}{s^{[0]2}} 
                    & \forall i_{1} \\

                    t_{\changein i_{j+1}}^{[j]2}
                    +
                    \frac{1}{1-\wtune} \left\| {\bf W}^{[j]}_{\changein :i_{j+1}} \right\|_2^2
                    &\!\!\!\!\leq
                    \frac{\left( 1-\psidual^{[j]} \right) \left( 1-\phidual^{[j]} \right) \sqrt{{{\rho}}^{[j]}}}{s^{[j]2}
                    +
                    \frac{1}{H^{[j]}} \sum_{i_j} \left\| {{\bg{\Psi}}}_{\origin:i_{j}}^{[j-1]} \left( {\bf W}_\changein \right) \right\|_2^2 \left( \frac{W_{\origin i_j i_{j+1}}^{[j]2}}{
\frac{1}{1-\wtune} \left\| {\bf W}^{[j]}_{\changein :i_{j+1}} \right\|_2^2 - W^{[j]2}_{\changein i_j,i_{j+1}}
} + 1 \right) 
                    \frac{\phi^{[j-1]}}{H^{[j-1]}}} 
                    & \forall j \in \infset{N}_D \backslash \left\{ 0 \right\}, i_{j+1} \\
 \end{array}
\]
and so it suffices that:
\[
 \begin{array}{rll}
                    t_{\changein i_{1}}^{[0]2}
                    &\!\!\!\!\leq
                    \frac{\left( 1-\psidual^{[0]} \right)}{\left( \frac{s^{[0]2}}{\left( 1-\phidual^{[0]} \right) \sqrt{{{\rho}}^{[0]}}} \right)} 
                    & \forall i_{1} \\

                    t_{\changein i_{j+1}}^{[j]2}
                    +
                    \frac{1}{1-\wtune} \left\| {\bf W}^{[j]}_{\changein :i_{j+1}} \right\|_2^2
                    &\!\!\!\!\leq
                    \frac{\left( 1-\psidual^{[j]} \right)}{\frac{s^{[j]2}}{\left( 1-\phidual^{[j]} \right) \sqrt{{{\rho}}^{[j]}}}
                    +
                    \frac{1}{H^{[j]}} \left( \frac{\left\| {\bf W}_{\origin :i_{j+1}}^{[j]} \right\|_2^2}{
\frac{1}{1-\wtune} \left\| {\bf W}^{[j]}_{\changein :i_{j+1}} \right\|_F^2 - \left\| {\bf W}^{[j]}_{\changein :i_{j+1}} \right\|_\infty^2
} + H^{[j-1]} \right) 
                    \frac{\frac{\psi^{[j-1]}}{H^{[j-1]}} \frac{\phi^{[j-1]}}{H^{[j-1]}}}{\left( 1-\phidual^{[j]} \right) \sqrt{{{\rho}}^{[j]}}}} 
                    & \forall j \in \infset{N}_D \backslash \left\{ 0 \right\}, i_{j+1} \\
 \end{array}
\]
In terms of the scale factors, recalling theorems \ref{th:convergephi} and 
\ref{th:convergepsi}, we require that (sufficiently):
 \[
  \begin{array}{rl}
                     {\shadx}_{i_{1}}^{[0]2}
                     &\!\!\!\!\in
                     \left[
                     \frac{1}{1-\psidual^{[0]}} t_{\changein i_{1}}^{[0]2}, 
                     \frac{1}{\left( \frac{s^{[0]2}}{\left( 1-\phidual^{[0]} \right) \sqrt{{{\rho}}^{[0]}}} \right)} 
                     \right] \\

                     {\shadx}_{i_{j+1}}^{[j]2}
                     &\!\!\!\!\in
                     \left[
                     \frac{1}{1-\psidual^{[j]}} \left( t_{\changein i_{j+1}}^{[j]2} + \frac{1}{1-\wtune} \left\| {\bf W}^{[j]}_{\changein i_{j+1}} \right\|_2^2 \right), 
                     \frac{1}{\frac{s^{[j]2}}{\left( 1-\phidual^{[j]} \right) \sqrt{{{\rho}}^{[j]}}}
                    +
                    \frac{1}{H^{[j]}} \left( \frac{\left\| {\bf W}_{\origin :i_{j+1}}^{[j]} \right\|_2^2}{
\frac{1}{1-\wtune} \left\| {\bf W}^{[j]}_{\changein :i_{j+1}} \right\|_F^2 - \left\| {\bf W}^{[j]}_{\changein :i_{j+1}} \right\|_\infty^2
} + H^{[j-1]} \right) 
                    \frac{\frac{\psi^{[j-1]}}{H^{[j-1]}} \frac{\phi^{[j-1]}}{H^{[j-1]}}}{\left( 1-\phidual^{[j]} \right) \sqrt{{{\rho}}^{[j]}}}}
                     \right] \\
  \end{array}
 \]
$\forall i_1$, $\forall j \in \infset{N}_D \backslash \left\{ 0 \right\}, i_{j+1}$, 
which completes the proof.
\end{proof}

\section{Canonical Scaling} \label{sec:canon_scale}

In the next section we will be considering regularised risk minimisation 
problems of the form:
\begin{equation}
 \begin{array}{l}
  {\bf W}_\changein^\regstep = \mathop{\rm argmin}\limits_{{\bf W}_\changein \in \infset{W}_\origin} 
  R_\lambda \left( {\bf W}_\changein \right) = 
  \lambda h \left( \left\| {\bg{\Psi}}_\origin \left( {\bf W}_\changein \right) \right\|_{\mathcal{W}_\origin} \right)
  + 
  \sum_k E \left( {\bf x}^{\{k\}}, {\bf y}^{\{k\}}, {\bf f}_\origin \left( {\bf x}^{\{k\}} \right) + \left< {\bg{\Phi}}_\origin \left( {\bf x}^{\{k\}} \right), {\bg{\Psi}}_\origin \left( {\bf W}_\changein \right) \right>_{\mathcal{X}_\origin \times \mathcal{W}_\origin} \right)
 \end{array}
 \label{eq:Wchangein}
\end{equation}
The question we will be addressing is:
\begin{quote}
 {\it For a given neural network with initial weights and biases ${\bf 
 W}_\origin$, let ${\bf W}_\changein^\backstep$ be the back-propagation 
 weight-step (gradient descent with learning rate $\learnrate$) defined by 
 (\ref{eq:backproprep}), and let ${\bf W}_\changein^\regstep$ be a weight-step 
 solving the regularised risk minimisation problem (\ref{eq:Wchangein}).  
 Given the gradient-descent derived weight-step ${\bf W}_\changein^\backstep$, 
 is there a selection shadow weights and scaling factors, possibly dependent 
 on ${\bf W}_\changein^\backstep$, and regularisation parameter $\lambda$ that 
 would ensure that ${\bf W}_\changein^\regstep = {\bf W}_\changein^\backstep$?}
\end{quote}
If the answer is yes (which we demonstrate) then we can gain understanding of 
back-propagation by analysing (\ref{eq:Wchangein}).  Now, the solution to 
(\ref{eq:Wchangein}) must satisfy first-order optimality conditions (we assume 
differentiability for simplicity here):
\[
{{
 \begin{array}{l}
  \left. \frac{\partial}{\partial {\bf W}_\changein} 
  \left\| {\bg{\Psi}}_\origin \left( {\bf W}_\changein \right) \right\|_{\mathcal{W}_\origin} 
  \right|_{{\bf W}_\changein = {\bf W}_\changein^\regstep}
  = \ldots \\ \;\; \ldots =
  - \frac{1}{\lambda}
  \frac{1}{h^{(1)} \left( \left\| {\bg{\Psi}}_\origin \left( {\bf W}_\changein \right) \right\|_{\mathcal{W}_\origin} \right)}
  \left. \frac{\partial}{\partial {\bf W}_\changein} 
  \sum_k E \left( {\bf x}^{\{k\}}, {\bf y}^{\{k\}}, {\bf f}_\origin \left( {\bf x}^{\{k\}} \right) + \left< {\bg{\Phi}}_\origin \left( {\bf x}^{\{k\}} \right), {\bg{\Psi}}_\origin \left( {\bf W}_\changein \right) \right>_{\mathcal{X}_\origin \times \mathcal{W}_\origin} \right)
  \right|_{{\bf W}_\changein = {\bf W}_\changein^\regstep}
 \end{array}
}}
\]
Now, noting that the derivative of the second term in (\ref{eq:Wchangein}) 
corresponds to the gradient in back-propagation, if the gradient of the first 
(regularisation) term satisfies:
\[
 \begin{array}{rl}
  \left. \frac{\partial}{\partial {\bf W}_\changein} \left\| {\bg{\Psi}}_\origin \left( {\bf W}_\changein \right) \right\|_{\mathcal{W}_\origin} \right|_{{\bf W}_\changein = {\bf W}_\changein^\backstep} = \bangrad {\bf W}_\changein^\backstep
 \end{array}
\]
for some $\bangrad \in \infset{R}_+$, and:
\[
 \begin{array}{rl}
  \lambda = \frac{1}{\learnrate \bangrad h^{(1)} \left( \left\| {\bg{\Psi}}_\origin \left( {\bf W}_\changein^\backstep \right) \right\|_{\mathcal{W}_\origin} \right)}
 \end{array}
\]
then:
\[
 \begin{array}{l}
  {\bf W}_\changein^\regstep 
  =
  {\bf W}_\changein^\backstep 
  =
  -\learnrate \left. \frac{\partial}{\partial {\bf W}_\changein} 
  \sum_k E \left( {\bf x}^{\{k\}}, {\bf y}^{\{k\}}, {\bf f}_\origin \left( {\bf x}^{\{k\}} \right) + \left< {\bg{\Phi}}_\origin \left( {\bf x}^{\{k\}} \right), {\bg{\Psi}}_\origin \left( {\bf W}_\changein \right) \right>_{\mathcal{X}_\origin \times \mathcal{W}_\origin} \right)
  \right|_{{\bf W}_\changein = {\bf W}_\changein^\backstep}
 \end{array}
\]
Thus the question whether there exists scaling factors, shadow weights and 
$\lambda$ such that the regularised risk minimisation weight-step corresponds to 
the gradient-descent weight-step for a specified learning rate $\learnrate$ can 
be answered in the affirmative by proving the existence of canonical scalings, 
which we define as follows:
\begin{def_canonical_scaling}[Canonical Scaling]
 For a given neural network, initial weights ${\bf W}_\origin$ and weight step 
 ${\bf W}_{\changein}^\backstep$ generated by back-propagation, we define a 
 {\em canonical scaling} to be a set of shadow weights and scaling factors for 
 which:
 \[
  \begin{array}{rl}
   \left. \frac{\partial}{\partial {\bf W}_\changein} \left\| {\bg{\Psi}}_\origin \left( {\bf W}_\changein \right) \right\|_{\mathcal{W}_\origin} \right|_{{\bf W}_\changein = {\bf W}_\changein^\backstep} = \bangrad {\bf W}_\changein^\backstep
  \end{array}
 \]
 for some $\bangrad \in \infset{R}_+$, and $\left\| {\bg{\Psi}}_\origin ( 
 {\bf W}_\changein^\backstep) \right\|_{\mathcal{W}_\origin} < \infty$, 
 $\left\| {\bg{\Phi}}_\origin ({\bf x}) \right\|_{\mathcal{X}_\origin} < 
 \infty$ $\forall {\bf x} \in \infset{X}$.  We call the kernels and norms 
 induced using a canonical scaling {\em canonical induced kernels} and {\em 
 canonical induced norms}, respectively.
\end{def_canonical_scaling}

With regard to existence we have the following theorem that sets out 
sufficient conditions for the weight-step for a canonical scaling to exist.  
Defining $\wtune \in (0,1)$ and:
\[
 \begin{array}{rl}
 \begin{array}{rl}
  {t}_{\changein i_{j+1}}^{[j]\backstep 2} 
  &\!\!\!\!= 
  \left\{ \begin{array}{ll}
   2b_{\changein i_{  1}}^{[0]\backstep 2} + 2 \left\| {\bf W}_{\changein :i_{  1}}^{[0]\backstep} \right\|_2^2 
   & \mbox{if } j = 0 \\
   2b_{\changein i_{j+1}}^{[j]\backstep 2} + \frac{2-\wtune}{1-\wtune} \left\| {\bf W}_{\changein :i_{j+1}}^{[j]\backstep} \right\|_2^2
   & \mbox{otherwise} \\
  \end{array} \right. \\
 \end{array} & \forall j \in \infset{N}_D, i_{j+1}
 \end{array}
\]
our central result is as follows:
\begin{th_canexistgen}
 Let $\phidual, \psidual, \wtune, \chi \in (0,1)$ and for a given neural network 
 with initial weights ${\bf W}_\origin$, and let ${\bf W}_\changein^\backstep$ 
 be the weight-step for this derived from back-propagation.  Let:
 \[
  \begin{array}{c}
  \begin{array}{rl}
   0 \leq \psidual^{[j]}
   &\!\!\!\!\leq
   \left\{ \begin{array}{ll}
   1-\frac{s^{[D-1]2}}{\left( 1-\phidual \right) \sqrt{{{\rho}}^{[D-1]}}} \left\| {\bf t}_{\changein}^{[D-1]\backstep} \right\|_\infty^2
   & \mbox{if } j = D-1 \\

   \frac{1}{1-\wtune} \frac{\left\| {\bf W}_{\changein}^{[j+1]\backstep} \right\|_F^2}{\left\| {{\bf t}_{\changein}^{[j]\backstep}} \right\|_\infty^2}
   & \mbox{otherwise} \\
   \end{array} \right. \\

   \frac{\psi^{[j]}}{H^{[j]}}
   &\!\!\!\!=
   \theta \left( 1-\psidual^{[j]} \right)
   =
   \frac{1-\psidual^{[j]}}{\psidual^{[j]}} \\
  \end{array} \;\;\;\; \forall j \in \infset{N}_D \\
  \end{array}
 \]
 and:
 \[
 {\!\!\!\!{ %{\!\!\!\!\!\!\!\!\!\!\!\!\!\!\!\!\!\!\!\!\!\!\!\!{
  \begin{array}{c}
  \begin{array}{rl}
   1 \geq \phidual^{[j]}
   &\!\!\!\!\geq
   \left\{ \begin{array}{ll}
      \phidual
   & \mbox{if } j = D-1 \\ \\

      1 - \frac{1}{\sqrt{\rho^{[D-2]}}} \overline{\sigma}^{[D-2]-1} \left(
      \frac{{\frac{1}{1-\chi}\frac{1-\psidual^{[D-1]}}{\left\| {\bf t}_{\changein}^{[D-1]\backstep} \right\|_\infty^2}}
            -
            \frac{s^{[D-1]2}}{\left( 1-\phidual^{[D-1]} \right) \sqrt{{{\rho}}^{[D-1]}}}}
           {\frac{1}{H^{[D-1]}} \mathop{\max}\limits_{i_D} \left\{ \frac{\left\| {\bf W}_{\origin :i_{D}}^{[D-1]} \right\|_2^2}{
            \frac{1}{1-\wtune} \left\| {\bf W}^{[D-1]\backstep}_{\changein :i_{D}} \right\|_2^2 - \left\| {\bf W}^{[D-1]\backstep}_{\changein :i_{D}} \right\|_\infty^2
            } + H^{[D-2]} \right\}
            \frac{\frac{\psi^{[D-2]}}{H^{[D-2]}}}{\left( 1-\phidual^{[D-1]} \right) \sqrt{{{\rho}}^{[D-1]}}}}
       \right)
   & \mbox{if } j = D-2 \\ \\

     1 - \frac{1}{\sqrt{\rho^{[j]}}} \overline{\sigma}^{[j] -1} \left(
     \frac{
     \frac{1}{{\left\| {\bf t}_{\changein}^{[j+1]\backstep} \right\|_{\infty}^2}} {\left( 1
      -
      \frac{1}{1-\wtune} \frac{\left\| {\bf W}_{\changein}^{[j+2]\backstep} \right\|_F^2}{\left\| {\bf t}_{\changein}^{[j+1]\backstep} \right\|_{\infty}^2} \right)}
      -
      \frac{s^{[j+1]2}}{\left( 1-\phidual^{[j+1]} \right) \sqrt{{{\rho}}^{[j+1]}}}
      }{
                       {\frac{1}{H^{[j+1]}} \mathop{\max}\limits_{i_{j+2}} \left\{ \frac{\left\| {\bf W}_{\origin :i_{j+2}}^{[j+1]} \right\|_2^2}{
        \frac{1}{1-\wtune} \left\| {\bf W}^{[j+1]\backstep}_{\changein :i_{j+2}} \right\|_2^2 - \left\| {\bf W}^{[j+1]\backstep}_{\changein :i_{j+2}} \right\|_\infty^2
        } + H^{[j]} \right\}
                       \frac{\frac{\psi^{[j]}}{H^{[j]}} }{\left( 1-\phidual^{[j+1]} \right) \sqrt{{{\rho}}^{[j+1]}}}}
      }
      \right)
   & \mbox{otherwise} \\
   \end{array} \right. \\

   \frac{\phi^{[j]}}{H^{[j]}}
   &\!\!\!\!=
   \overline{\sigma}^{[j]} \left( \left( 1-\phidual^{[j]} \right) \sqrt{{\rho}^{[j]}} \right) \\
  \end{array}  \\
  \end{array}
 }}
 \]
 $\forall j \in \infset{N}_D$.  For some $\alpha^{[j]} \in \infset{R}_+$ there exists $\degrowth^{[j]} \in (0,1)$ 
 $\forall j \in \infset{N}_D \backslash \{ 0 \}$ such that:\footnote{Note that 
 $b_{\changein i_{j+1}}^{[j]\backstep}$ is proportional to $\alpha^{[j]}$, so we can 
 always increase ${t}_{\changein i_{j+1}}^{[j]\backstep 2}$ to ensure 
 the condition holds by increasing $\alpha^{[j]}$ sufficiently.}
 \[
  \begin{array}{rl}
   \left\| {\bf W}_{\changein}^{[j+1]\backstep} \right\|_F^2
   =
   \left( 1 - \wtune \right) \left( 1 - \degrowth^{[j+1]} \right) \left\| {\bf t}_{\changein}^{[j]\backstep} \right\|_{\infty}^2 \\
  \end{array}
 \]
 If the weight-step satisfies:
 \[
  \begin{array}{rl}
   \left\| {\bf t}_{\changein}^{[j]\backstep} \right\|_{\infty}^2
   &\!\!\!\!<
   \left\{ \begin{array}{ll}
   \frac{\left( 1 - \chi \right) \left( 1-\psidual^{[D-1]} \right)}{\left( \frac{s^{[D-1]2}}{\left( 1-\phidual^{[D-1]} \right) \sqrt{{{\rho}}^{[D-1]}}} \right)}
   & \mbox{if } j = D-1 \\
   \frac{\degrowth^{[j+1]} \left( 1-\psidual^{[j]} \right)}{\left( \frac{s^{[j]2}}{\left( 1-\phidual^{[j]} \right) \sqrt{{{\rho}}^{[j]}}} \right)}
   & \mbox{otherwise} \\
   \end{array} \right.
   \;\;\;\; \forall j \in \infset{N}_D
  \end{array}
 \]
 then there exists a canonical scaling:
 \[
  \begin{array}{rl}
   \left. \frac{\partial}{\partial {\bf W}_\changein} \left\| {\bg{\Psi}}_\origin \left( {\bf W}_\changein \right) \right\|_{\mathcal{W}_\origin} \right|_{{\bf W}_\changein = {\bf W}_\changein^\backstep} = \bangrad {\bf W}_\changein^\backstep
  \end{array}
 \]
 where:
 \[
  \begin{array}{l}
   \bangrad
   =
   \frac{4}{\left\| {\bf t}_{\changein}^{[D-1]\backstep} \right\|_\infty^2}
   \kappa \left( \frac{1 - \psidual^{[D-1]}}{1-\chi} \right) \\
  \end{array}
 \]
 and $\left\| {\bg{\Phi}}_\origin^{[j]} ({\bf x}) \right\|_F^2 \leq 
 \phi^{[j]}$ $\forall {\bf x} \in \infset{X}$, $j \in \infset{N}_D$ and 
 $\left\| {{\bg{\Psi}}}_\origin^{[j]} ({\bf W}_\changein) \right\|_F^2 \leq 
 \psi^{[j]}$ for all $j \in \infset{N}_D$.
 \label{th:canexistgen}
\end{th_canexistgen}
\begin{proof}
We aim to derive a canonical scalings for general, a-priori weight-steps 
${\bf W}_\changein^\backstep$.  Recall from (\ref{eq:inducedndefgrad}), 
(\ref{eq:inducednrecgradw}) and (\ref{eq:inducednrecgradb}) that:
\[
{{
 \begin{array}{rl}
  \left. \frac{\partial}{\partial {\bf W}_\changein^{[j']}} \left\| {\bg{\Psi}}_{\origin} \left( {\bf W}_\changein \right) \right\|_{\mathcal{W}_\origin}^2 \right|_{{\bf W}_\changein = {\bf W}_\changein^\backstep}
  &\!\!\!\!=
  \sum_{i_D} 
  \left[ \begin{array}{c} \left. \frac{\partial}{\partial W_{\changein i'_{j'},i'_{j'+1}}^{[j']}} \left\| {\bg{\Psi}}_{\origin :i_D}^{[D-1]} \left( {\bf W}_\changein \right) \right\|_2^2 \right|_{{\bf W}_\changein = {\bf W}_\changein^\backstep} \end{array} \right]_{i'_{j'},i'_{j'+1}} \\

  \left. \frac{\partial}{\partial {\bf b}_\changein^{[j']}} \left\| {\bg{\Psi}}_{\origin} \left( {\bf W}_\changein \right) \right\|_{\mathcal{W}_\origin}^2 \right|_{{\bf W}_\changein = {\bf W}_\changein^\backstep}
  &\!\!\!\!=
  \sum_{i_D} 
  \left[ \begin{array}{c} \left. \frac{\partial}{\partial b_{\changein i'_{j'+1}}^{[j']}} \left\| {\bg{\Psi}}_{\origin :i_D}^{[D-1]} \left( {\bf W}_\changein \right) \right\|_2^2 \right|_{{\bf W}_\changein = {\bf W}_\changein^\backstep} \end{array} \right]_{i'_{j'+1}} \\
 \end{array}
 }}
\]
where, recursively $\forall j \in \infset{N}_D$:
\[
 {{
 \begin{array}{l}
  \left. \frac{\partial}{\partial W_{\changein i'_{j'},i'_{j'+1}}^{[j']}} \left\| {\bg{\Psi}}_{\origin :i_{j+1}}^{[j]} \left( {\bf W}_\changein \right) \right\|_2^2 \right|_{{\bf W}_\changein = {\bf W}_\changein^\backstep}
  = \ldots \\ \;\;\;\;\ldots = \left\{ \begin{array}{ll}
                    \frac{1}{{{\shadx}_{i_{j+1}}^{[j]2}}} 
                    \theta^{(1)} \left( \frac{{t}_{\changein i_{j+1}}^{[j]\backstep 2}}{{{\shadx}_{i_{j+1}}^{[j]2}}} \right)
                    \mathop{\sum}\limits_{i_j} 
                    \frac{{\shadwo}_{i_j,i_{j+1}}^{[j]2} + W_{\changein i_j, i_{j+1}}^{[j]\backstep 2}}{\shadw_{i_j}^{[j]2}} 
                    \left( \left. \frac{\partial}{\partial W_{\changein i'_{j'},i'_{j'+1}}^{[j']}} \left\| {\bg{\Psi}}_{\origin :i_{j}}^{[j-1]} \left( {\bf W}_\changein \right) \right\|_2^2 \right|_{{\bf W}_\changein = {\bf W}_\changein^\backstep} \right)
                    & \mbox{if } j > j' \\ \\

                    \frac{1}{{{\shadx}_{i'_{j+1}}^{[j]2}}} 
                    \theta^{(1)} \left( \frac{{t}_{\changein i_{j+1}}^{[j]\backstep 2}}{{{\shadx}_{i_{j+1}}^{[j]2}}} \right)
                    \delta_{i_{j+1},i'_{j+1}} 
                    2 \left( 1 + \frac{\left\| {\bg{\Psi}}_{\origin :i'_{j}}^{[j-1]} \left( {\bf W}_\changein^\backstep \right) \right\|_2^2}{\shadw_{i'_j}^{[j]2}} \right) W_{\changein i'_j, i_{j+1}}^{[j]\backstep} 
                    & \mbox{if } j = j' > 0 \\ \\

                    \frac{1}{{{\shadx}_{i'_{1}}^{[0]2}}} 
                    \theta^{(1)} \left( \frac{{t}_{\changein i_{1}}^{[0]\backstep 2}}{{{\shadx}_{i_{1}}^{[0]2}}} \right)
                    \delta_{i_{1},i'_{1}} 
                    4W_{\changein i'_0, i_{1}}^{[0]\backstep}
                    & \mbox{if } j = j' = 0 \\
   \end{array} \right. \\
 \end{array}
 }}
\]
\[
{{
 \begin{array}{rl}
  \left. \frac{\partial}{\partial b_{\changein i'_{j'+1}}^{[j']}} \left\| {\bg{\Psi}}_{\origin :i_{j+1}}^{[j]} \left( {\bf W}_\changein \right) \right\|_2^2 \right|_{{\bf W}_\changein = {\bf W}_\changein^\backstep}
  = \ldots \\ \;\;\;\;\ldots = \left\{ \begin{array}{ll}
                    \frac{1}{{{\shadx}_{i_{j+1}}^{[j]2}}} 
                    \theta^{(1)} \left( \frac{{t}_{\changein i_{j+1}}^{[j]\backstep 2}}{{{\shadx}_{i_{j+1}}^{[j]2}}} \right)
                    \mathop{\sum}\limits_{i_j} 
                    \frac{{\shadwo}_{i_j,i_{j+1}}^{[j]2} + W_{\changein i_j, i_{j+1}}^{[j]\backstep 2}}{\shadw_{i_j}^{[j]2}}  
                    \left( \left. \frac{\partial}{\partial b_{\changein i'_{j'+1}}^{[j']}} \left\| {\bg{\Psi}}_{\origin :i_{j}}^{[j-1]} \left( {\bf W}_\changein \right) \right\|_2^2 \right|_{{\bf W}_\changein = {\bf W}_\changein^\backstep} \right)
                    & \mbox{if } j > j' \\ \\

                    \frac{1}{{{\shadx}_{i'_{j+1}}^{[j]2}}} 
                    \theta^{(1)} \left( \frac{{t}_{\changein i_{j+1}}^{[j]\backstep 2}}{{{\shadx}_{i_{j+1}}^{[j]2}}} \right)
                    \delta_{i_{j+1},i'_{j+1}} 
                    4b_{\changein i_{j+1}}^{[j]\backstep} 
                    & \mbox{if } j = j' \\

   \end{array} \right. \\
 \end{array}
 }}
\]
and we have defined:
\[
 \begin{array}{rl}
 \begin{array}{rl}
  {t}_{\changein i_{j+1}}^{[j]\backstep 2} 
  &\!\!\!\!= 
  \left\{ \begin{array}{ll}
   \left[ 2b_{\changein i_{  1}}^{[0]\backstep 2} + 2 \left\| {\bf W}_{\changein :i_{  1}}^{[0]\backstep} \right\|_2^2 \right]_{i_{  1}} 
   & \mbox{if } j = 0 \\
   \left[ 2b_{\changein i_{j+1}}^{[j]\backstep 2} +   \left\| {\bf W}_{\changein :i_{j+1}}^{[j]\backstep} \right\|_2^2
          + \mathop{\sum}\limits_{i_j} \left( {\shadwo}_{i_j,i_{j+1}}^{[j]2} + W_{\changein i_j, i_{j+1}}^{[j]\backstep 2} \right) 
            \frac{\left\| {\bg{\Psi}}_{\origin :i_{j}}^{[j-1]} \left( {\bf W}_\changein^\backstep \right) \right\|_2^2}{\shadw_{i_j}^{[j]2}} 
   \right]_{i_{j+1}} 
   & \mbox{otherwise} \\
  \end{array} \right. \\
 \end{array} & \forall j \in \infset{N}_D, i_{j+1}
 \end{array}
\]
so that:
\[
 \begin{array}{rl}
  \left\| {\bg{\Psi}}_{\origin i_{j+1}}^{[j]} \left( {\bf W}_\changein^\backstep \right) \right\|_2^2 
  = 
  \theta \left( \frac{1}{\shadx_{i_{j+1}}^{[j]}} {t}_{\changein i_{j+1}}^{[j]\backstep 2} \right)
  =
  \frac{{t}_{\changein i_{j+1}}^{[j]\backstep 2}}{\shadx_{i_{j+1}}^{[j]2}-{t}_{\changein i_{j+1}}^{[j]\backstep 2}}
 \end{array}
\]

Note that the recursive fomula for the norm gradient with respect to a specific 
weight $W_{\changein i'_{j'}, i'_{j'+1}}^{[j']}$ or bias $b_{\changein 
i'_{j'+1}}^{[j']}$ precisely mirrors the structure of the network, where the 
calculation begins at the ouput layer and recurses backwards along all possible 
paths to the neuron $i'_{j'+1}$ in layer $j'$.  Indeed, if we define the set of 
all paths from any neuron in the output layer to neural $i'_{j'+1}$ in layer $j'$ 
as:
\[
 \begin{array}{rl}
  \mathcal{P}_{i'_{j'+1}}^{[j']} = \left\{ {\tt i} = \left( i_D, i_{D-1}, \ldots, i_{j'+1} \right) : i_{j+1} \in \infset{N}_{H^{[j]}} \forall j \in \infset{N}_D \backslash \infset{N}_{j'+1}, i_{j'+1} = i'_{j'+1} \right\}
 \end{array}
\]
then we can re-write the norm gradient as a pathwise sum:
\[
{{
 \begin{array}{l}
  \left. \frac{\partial}{\partial W_{\changein i'_{j'}, i'_{j'+1}}^{[j']}} \left\| {\bg{\Psi}}_{\origin} \left( {\bf W}_\changein \right) \right\|_{\mathcal{W}_\origin}^2 \right|_{{\bf W}_\changein = {\bf W}_\changein^\backstep}
  = \ldots \\ \;\;\;\;\;\;\ldots =
  \left( 
  \mathop{\sum}\limits_{{\tt i} \in \mathcal{P}_{i'_{j'+1}}^{[j']}}
  g_{i_{j'+1}}^{[j']\backstep} 
  \mathop{\prod}\limits_{j'' = j'+1}^{D-1}
  g_{i_{j''+1}}^{[j'']\backstep} h_{i_{j''},i_{j''+1}}^{[j''-1,j'']\backstep}
  \right)
  \left\{ \begin{array}{ll}
                    2 \left( 1 + \frac{\left\| {\bg{\Psi}}_{\origin :i'_{j}}^{[j-1]} \left( {\bf W}_\changein^\backstep \right) \right\|_2^2}{\shadw_{i'_j}^{[j]2}} \right) W_{\changein i'_j, i_{j+1}}^{[j]\backstep} 
                    & \mbox{if } j' > 0 \\ \\

                    4W_{\changein i'_0, i_{1}}^{[0]\backstep}
                    & \mbox{otherwise} \\
  \end{array} \right. \\

  \left. \frac{\partial}{\partial b_{\changein i'_{j'+1}}^{[j']}} \left\| {\bg{\Psi}}_{\origin} \left( {\bf W}_\changein \right) \right\|_{\mathcal{W}_\origin}^2 \right|_{{\bf W}_\changein = {\bf W}_\changein^\backstep}
  = \ldots \\ \;\;\;\;\;\;\ldots =
  \left( 
  \mathop{\sum}\limits_{{\tt i} \in \mathcal{P}_{i'_{j'+1}}^{[j']}}
  g_{i_{j'+1}}^{[j']\backstep} 
  \mathop{\prod}\limits_{j'' = j'+1}^{D-1}
  g_{i_{j''+1}}^{[j'']\backstep} h_{i_{j''},i_{j''+1}}^{[j''-1,j'']\backstep}
  \right)
  4b_{\changein i_{j+1}}^{[j]\backstep} \\
 \end{array}
 }}
\]
where each term in the sum of paths is the product of neuron costs:
\[

\]
and link costs:
\[
 \begin{array}{rl}
  h^{[j-1,j]\backstep}_{i_j,i_{j+1}}
  &\!\!\!\!=
  \shadwo_{i_j,i_{j+1}}^{[j]2} + W_{\changein i_j, i_{j+1}}^{[j]\backstep 2}
 \end{array}
\]
for all neurons and links between neurons on that path from the output 
layer back to neuron $i'_{j'+1}$ in layer $j'$..

Ignoring for the moment the question of convergence, we see that to obtain 
a canonical scaling we must use the scaling factors and shadow weights to 
``even out'' the path-dependence of the neuron and link costs and flatten 
the dependence on the feature-map norm at the terminating neuron $i_{j+1}$ 
in layer $j$.  To this end it is straightforward to see that, if we select:
\begin{equation}
 %
\label{eq:canscalweight}
\end{equation}
where $\wtune \in (0,1)$ (remember that we are assuming the 
back-propagation weight-step ${\bf W}_\changein^\backstep$ is calculated 
a-priori and then attempting to make the regularisation-based weight-step 
${\bf W}_\changein^\regstep$ correspond to this) then:
\[
{{
 %
\]
Using (\ref{eq:euceuc_normsdefof}), we see that
\[
 \begin{array}{rl}
  \left\| {\bg{\Psi}}_{\origin i_{j+1}}^{[j]} \left( {\bf W}_\changein^\backstep \right) \right\|_2^2 
  = 
  \theta \left( \frac{1}{\shadx_{i_{j+1}}^{[j]2}} {t}_{\changein i_{j+1}}^{[j]\backstep 2} \right)
  =
  \frac{{t}_{\changein i_{j+1}}^{[j]\backstep 2}}{\shadx_{i_{j+1}}^{[j]2}-{t}_{\changein i_{j+1}}^{[j]\backstep 2}}
 \end{array}%
\]

Take as given than the scale factors (\ref{eq:canscalweight}) are used from 
here on.  Having used our scale factors to even out the link costs in the norm 
gradient, our next task is to use the scale factors to even out the neuron 
costs and obtain a canonical scaling.  Again neglecting (for the moment) the 
question of convergence, consider the gradients of the weights and biases in 
the output layer, which take the particularly simple form:
\[
{{
 \begin{array}{rl}
  \left. \frac{\partial}{\partial W_{\changein i'_{D-1}, i'_{D}}^{[D-1]}} \left\| {\bg{\Psi}}_{\origin} \left( {\bf W}_\changein \right) \right\|_{\mathcal{W}_\origin}^2 \right|_{{\bf W}_\changein = {\bf W}_\changein^\backstep}
  &\!\!\!\!=
  g_{i'_{D}}^{[D-1]\backstep}
  4W_{\changein i'_{D-1}, i_{D}}^{[D-1]\backstep} \\

  \left. \frac{\partial}{\partial b_{\changein i'_{D}}^{[D-1]}} \left\| {\bg{\Psi}}_{\origin} \left( {\bf W}_\changein \right) \right\|_{\mathcal{W}_\origin}^2 \right|_{{\bf W}_\changein = {\bf W}_\changein^\backstep}
  &\!\!\!\!=
  g_{i'_{D}}^{[D-1]\backstep}
  4b_{\changein i_{D}}^{[D-1]\backstep} \\
 \end{array}
 }}
\]
For canonical scaling, we must have that:
\[
 \begin{array}{l}
  \frac{4}{{{\shadx}_{i_{D}}^{[D-1]2}}} 
  \theta^{(1)} \left( \frac{{t}_{\changein i_{D}}^{[D-1]\backstep 2}}{{{\shadx}_{i_{D}}^{[D-1]2}}} \right)
  =

  \frac{4}{{t}_{\changein i_{D}}^{[D-1]\backstep 2}}
  \kappa \left( \frac{{t}_{\changein i_{D}}^{[D-1]\backstep 2}}{{{\shadx}_{i_{D}}^{[D-1]2}}} \right)
  =
  \bangrad
 \end{array}
\]
where we recall that:
\[
 \begin{array}{rl}
  \kappa       \left( \zeta \right) &\!\!\!\!= \zeta \theta^{(1)} \left( \zeta \right) = \frac{\zeta}{\left( 1-\zeta \right)^2} \\
  \kappa^{-1}  \left( \zeta \right) &\!\!\!\!= \frac{2\zeta+1-\sqrt{4\zeta+1}}{2\zeta} \\
 \end{array}
\]
and hence:
\[
 \begin{array}{l}
  \shadx_{i_{D}}^{[D-1]2}
  =
  \frac{{t}_{\changein i_{D}}^{[D-1]\backstep 2}}
       {\kappa^{-1} \left( 
        \frac{{t}_{\changein i_{D}}^{[D-1]\backstep 2} \bangrad}{4}
        \right)}
  =
  \frac{2\frac{\bangrad {t}_{\changein i_{D}}^{[D-1]\backstep 2}}{4}}
       {2\frac{\bangrad {t}_{\changein i_{D}}^{[D-1]\backstep 2}}{4}+1-\sqrt{4\frac{\bangrad {t}_{\changein i_{D}}^{[D-1]\backstep 2}}{4}+1}}
  {t}_{\changein i_{D}}^{[D-1]\backstep 2}
 \end{array}
\]
Letting $\shadx_{i_{D}}^{[D-1]2}$ take this value, consder the gradients of 
the weights and biases for layer $D-2$, which simplify to:
\[
{{
 \begin{array}{rl}
  \left. \frac{\partial}{\partial W_{\changein i'_{D-2}, i'_{D-1}}^{[D-2]}} \left\| {\bg{\Psi}}_{\origin} \left( {\bf W}_\changein \right) \right\|_{\mathcal{W}_\origin}^2 \right|_{{\bf W}_\changein = {\bf W}_\changein^\backstep}
  &\!\!\!\!=
  \bangrad
  \left\| {\bf W}^{[D-1]} \right\|_F^2
  g_{i_{D-1}}^{[D-2]\backstep}
  W_{\changein i'_{D-2}, i_{D-1}}^{[D-2]\backstep} \\

  \left. \frac{\partial}{\partial b_{\changein i'_{D-1}}^{[D-2]}} \left\| {\bg{\Psi}}_{\origin} \left( {\bf W}_\changein \right) \right\|_{\mathcal{W}_\origin}^2 \right|_{{\bf W}_\changein = {\bf W}_\changein^\backstep}
  &\!\!\!\!=
  \bangrad
  \left\| {\bf W}^{[D-1]} \right\|_F^2
  g_{i_{D-1}}^{[D-2]\backstep}
  b_{\changein i_{D-1}}^{[D-2]\backstep} \\
 \end{array}
 }}
\]
and we see that, for a canonical scaling, we must have that:
\[
 \begin{array}{rl}
  &
  \frac{1}{1-\wtune} \left\| {\bf W}_{\changein}^{[D-1]\backstep} \right\|_F^2
  \frac{\shadx_{i_{D-1}}^{[D-2]2}-t_{\changein i_{D-1}}^{[D-2]2}}{t_{\changein i_{D-1}}^{[D-2]2}}
  \frac{1}{{{\shadx}_{i_{D-1}}^{[D-2]2}}}
  \frac{1}{\left( 1 - \frac{{t}_{\changein i_{D-1}}^{[D-2]\backstep 2}}{{{\shadx}_{i_{D-1}}^{[D-2]2}}} \right)^2}
  =
  1 \\

  \Rightarrow &
  \frac{1}{1-\wtune} \left\| {\bf W}_{\changein}^{[D-1]\backstep} \right\|_F^2
  \frac{\shadx_{i_{D-1}}^{[D-2]2}-t_{\changein i_{D-1}}^{[D-2]2}}{t_{\changein i_{D-1}}^{[D-2]2}}
  \frac{\shadx_{i_{D-1}}^{[D-2]2}}{\left( \shadx_{i_{D-1}}^{[D-2]2} - {t}_{\changein i_{D-1}}^{[D-2]\backstep 2} \right)^2}
  =
  1 \\

  \Rightarrow &
  \frac{1}{1-\wtune} \left\| {\bf W}_{\changein}^{[D-1]\backstep} \right\|_F^2
  \shadx_{i_{D-1}}^{[D-2]2}
  =
  {t_{\changein i_{D-1}}^{[D-2]2}} \shadx_{i_{D-1}}^{[D-2]2} 
  - 
  {t}_{\changein i_{D-1}}^{[D-2]\backstep 4} \\

  \Rightarrow &
  \left(
  {t_{\changein i_{D-1}}^{[D-2]2}}
  -
  \frac{1}{1-\wtune} \left\| {\bf W}_{\changein}^{[D-1]\backstep} \right\|_F^2
  \right)
  \shadx_{i_{D-1}}^{[D-2]2}
  =
  {t}_{\changein i_{D-1}}^{[D-2]\backstep 4} \\

  \Rightarrow &
  \shadx_{i_{D-1}}^{[D-2]2}
  =
  \frac{{t}_{\changein i_{D-1}}^{[D-2]\backstep 2}}
       {{t_{\changein i_{D-1}}^{[D-2]2}} - \frac{1}{1-\wtune} \left\| {\bf W}_{\changein}^{[D-1]\backstep} \right\|_F^2}
  {t}_{\changein i_{D-1}}^{[D-2]\backstep 2} \\
 \end{array}
\]
For layer $D-3$, using the scale factors derived thus far:
\[
{{
 \begin{array}{rl}
  \left. \frac{\partial}{\partial W_{\changein i'_{D-3}, i'_{D-2}}^{[D-3]}} \left\| {\bg{\Psi}}_{\origin} \left( {\bf W}_\changein \right) \right\|_{\mathcal{W}_\origin}^2 \right|_{{\bf W}_\changein = {\bf W}_\changein^\backstep}
  &\!\!\!\!=
  \bangrad
  \left\| {\bf W}^{[D-2]} \right\|_F^2
  g_{i_{D-2}}^{[D-3]\backstep}
  W_{\changein i'_{D-3}, i_{D-2}}^{[D-3]\backstep} \\

  \left. \frac{\partial}{\partial b_{\changein i'_{D-2}}^{[D-3]}} \left\| {\bg{\Psi}}_{\origin} \left( {\bf W}_\changein \right) \right\|_{\mathcal{W}_\origin}^2 \right|_{{\bf W}_\changein = {\bf W}_\changein^\backstep}
  &\!\!\!\!=
  \bangrad
  \left\| {\bf W}^{[D-2]} \right\|_F^2
  g_{i_{D-2}}^{[D-3]\backstep}
  b_{\changein i_{D-2}}^{[D-3]\backstep} \\
 \end{array}
 }}
\]
and so on.  Working back through all layers, therefore, we find that canonical 
scaling is attained if we select shadow weights:
\begin{equation}
 \begin{array}{rll}
  \shadw_{i_j}^{[j]2} &\!\!\!\!= \left\| {{\bg{\Psi}}}_{\origin:i_{j}}^{[j-1]} \left( {\bf W}_\changein^\backstep \right) \right\|_2^2 
  & \forall j \in \infset{N}_D, i_j \\

  \shadwo_{i_j,i_{j+1}}^{[j]2} &\!\!\!\!= \frac{1}{1-\wtune} \left\| {\bf W}^{[j]\backstep}_{\changein :i_{j+1}} \right\|_F^2 - W^{[j]\backstep 2}_{\changein i_j,i_{j+1}}
  & \forall j \in \infset{N}_D \backslash \left\{ 0 \right\}, i_j, i_{j+1} \\
 \end{array} 
\label{eq:canscalweightb}
\end{equation}
and scale factors:
\begin{equation}
{\!\!\!\!\!\!\!\!\!\!\!\!\!\!\!\!\!\!{
 \begin{array}{rll}
  \shadx_{i_{D}}^{[D-1]2}
  &\!\!\!\!=
  \frac{{t}_{\changein i_{D}}^{[D-1]\backstep 2}}
       {\kappa^{-1} \left( 
        \frac{{t}_{\changein i_{D}}^{[D-1]\backstep 2} \bangrad}{4}
        \right)}
  =
  \frac{2\frac{\bangrad {t}_{\changein i_{D}}^{[D-1]\backstep 2}}{4}}
       {2\frac{\bangrad {t}_{\changein i_{D}}^{[D-1]\backstep 2}}{4}+1-\sqrt{4\frac{\bangrad {t}_{\changein i_{D}}^{[D-1]\backstep 2}}{4}+1}}
  {t}_{\changein i_{D}}^{[D-1]\backstep 2}
  & \forall i_D \\

  \shadx_{i_{j+1}}^{[j]2}
  &\!\!\!\!=
  \frac{{t}_{\changein i_{j+1}}^{[j]\backstep 2}}{{t}_{\changein i_{j+1}}^{[j]\backstep 2} 
  - 
  \frac{1}{1-\wtune} \left\| {\bf W}_{\changein}^{[j+1]\backstep} \right\|_F^2}
  {{t}_{\changein i_{j+1}}^{[j]\backstep 2}} 
  & \forall j \in \infset{N}_{D-1}, i_{j+1} \\
 \end{array}
}}
\label{eq:canshadweightb}
\end{equation}
We also obtain our first additional constraint on the weight-step, 
namely, using the requisit positivity of $\shadx_{i_{j+1}}^{[j]2}$:
\begin{equation}
 \begin{array}{rl}
  \frac{1}{1-\wtune} \left\| {\bf W}_{\changein}^{[j+1]\backstep} \right\|_F^2
  \leq
  \left( 1 - \degrowth^{[j+1]} \right) \left\| {\bf t}_{\changein}^{[j]\backstep} \right\|_{\infty}^2 \\
 \end{array}
\label{eq:canaddcon}
\end{equation}

Finally we must consider the question of convergence given scale factors and 
shadow weights (\ref{eq:canscalweightb}), (\ref{eq:canshadweightb}).  Recall 
from theorem \ref{th:convergephipsi_spec} that, given scale factors 
(\ref{eq:canscalweightb}), using the definitions therein, if:
 \begin{equation}
  \begin{array}{rll}
                     t_{\changein i_{1}}^{[0]2}
                     &\!\!\!\!\leq
                     \frac{\left( 1-\psidual^{[0]} \right)}{\left( \frac{s^{[0]2}}{\left( 1-\phidual^{[0]} \right) \sqrt{{{\rho}}^{[0]}}} \right)} 
                     & \forall i_{1} \\

                     t_{\changein i_{j+1}}^{[j]2}
                     +
                     \frac{1}{1-\wtune} \left\| {\bf W}^{[j]}_{\changein :i_{j+1}} \right\|_2^2
                     &\!\!\!\!\leq
                     \frac{\left( 1-\psidual^{[j]} \right)}{\frac{s^{[j]2}}{\left( 1-\phidual^{[j]} \right) \sqrt{{{\rho}}^{[j]}}}
                     +
                     \frac{1}{H^{[j]}} \left( \frac{\left\| {\bf W}_{\origin :i_{j+1}}^{[j]} \right\|_2^2}{
\frac{1}{1-\wtune} \left\| {\bf W}^{[j]}_{\changein :i_{j+1}} \right\|_2^2 - \left\| {\bf W}^{[j]}_{\changein :i_{j+1}} \right\|_\infty^2
} + H^{[j-1]} \right) 
                     \frac{\frac{\psi^{[j-1]}}{H^{[j-1]}} \frac{\phi^{[j-1]}}{H^{[j-1]}}}{\left( 1-\phidual^{[j]} \right) \sqrt{{{\rho}}^{[j]}}}} 
                     & \forall j \in \infset{N}_D \backslash \left\{ 0 \right\}, i_{j+1} \\
  \end{array}
 \label{eq:eqnX}
\end{equation}
which we take as given, and:
\begin{equation}
 {\!\!\!\!\!{
  \begin{array}{rl}
                     {\shadx}_{i_{1}}^{[0]2}
                     &\!\!\!\!\in
                     \left[
                     \frac{1}{1-\psidual^{[0]}} t_{\changein i_{1}}^{[0]2}, 
                     \frac{1}{\left( \frac{s^{[0]2}}{\left( 1-\phidual^{[0]} \right) \sqrt{{{\rho}}^{[0]}}} \right)} 
                     \right] \\

                     {\shadx}_{i_{j+1}}^{[j]2}
                     &\!\!\!\!\in
                     \left[
                     \frac{1}{1-\psidual^{[j]}} \left( t_{\changein i_{j+1}}^{[j]2} + \frac{1}{1-\wtune} \left\| {\bf W}^{[j]}_{\changein :i_{j+1}} \right\|_2^2 \right), 
                     \frac{1}{\frac{s^{[j]2}}{\left( 1-\phidual^{[j]} \right) \sqrt{{{\rho}}^{[j]}}}
                    +
                    \frac{1}{H^{[j]}} \left( \frac{\left\| {\bf W}_{\origin :i_{j+1}}^{[j]} \right\|_2^2}{
\frac{1}{1-\wtune} \left\| {\bf W}^{[j]}_{\changein :i_{j+1}} \right\|_2^2 - \left\| {\bf W}^{[j]}_{\changein :i_{j+1}} \right\|_\infty^2
} + H^{[j-1]} \right) 
                    \frac{\frac{\psi^{[j-1]}}{H^{[j-1]}} \frac{\phi^{[j-1]}}{H^{[j-1]}}}{\left( 1-\phidual^{[j]} \right) \sqrt{{{\rho}}^{[j]}}}}
                     \right] \\
  \end{array}
 \label{eq:eqnY}
}\!\!}
\end{equation}
$\forall i_{1}$, $\forall j \in \infset{N}_D \backslash \left\{ 0 \right\}, i_{j+1}$, 
then the norms will convergence.  We must show that the scale factors for our 
canonical scaling satisfy the bounds required for convergence.  Consider first 
$\shadx_{i_D}^{[D-1]2}$.  In this case we require that, using (\ref{eq:canshadweightb}):
\[
 \begin{array}{rl}
  &
  \frac{{t}_{\changein i_{D}}^{[D-1]\backstep 2}}
       {\kappa^{-1} \left( 
        \frac{{t}_{\changein i_{D}}^{[D-1]\backstep 2} \bangrad}{4}
        \right)}
  \geq
                    \frac{1}{1-\psidual^{[D-1]}} {t}_{\changein i_{D}}^{[D-1]\backstep 2}
  \;\;\;\;\forall i_{D} \\

  \Rightarrow &
  \frac{\left\| {\bf t}_{\changein i_{D}}^{[D-1]\backstep} \right\|_\infty^2}{4} \bangrad
  \leq
  \kappa \left( 1-\psidual^{[D-1]} \right)
  =
  \frac{1-\psidual^{[D-1]}}{\psidual^{[D-1]2}} 
  = 
  \frac{1}{\psidual^{[D-1]}} 
  \frac{\psi^{[D-1]}}{H^{[D-1]}} \\
 \end{array}
\]
and $\forall i_D$:
\[
 \begin{array}{rl}
  &
  \frac{{t}_{\changein i_{D}}^{[D-1]\backstep 2}}
       {\kappa^{-1} \left( 
        \frac{{t}_{\changein i_{D}}^{[D-1]\backstep 2} \bangrad}{4}
        \right)}
  \leq
                     \frac{1}{\frac{s^{[D-1]2}}{\left( 1-\phidual^{[D-1]} \right) \sqrt{{{\rho}}^{[D-1]}}}
                    +
                    \frac{1}{H^{[D-1]}} \left( \frac{\left\| {\bf W}_{\origin :i_{D}}^{[D-1]} \right\|_2^2}{
\frac{1}{1-\wtune} \left\| {\bf W}^{[D-1]\backstep}_{\changein :i_{D}} \right\|_2^2 - \left\| {\bf W}^{[D-1]\backstep}_{\changein :i_{D}} \right\|_\infty^2
} + H^{[D-2]} \right) 
                    \frac{\frac{\psi^{[D-2]}}{H^{[D-2]}} \frac{\phi^{[D-2]}}{H^{[D-2]}}}{\left( 1-\phidual^{[D-1]} \right) \sqrt{{{\rho}}^{[D-1]}}}} \\

  \Rightarrow &
                    \frac{1}{H^{[D\!-\!1]}} \!\left( \!\frac{\left\| {\bf W}_{\origin :i_{D}}^{[D\!-\!1]} \right\|_2^2}{
\frac{1}{1-\wtune} \left\| {\bf W}^{[D\!-\!1]\backstep}_{\changein :i_{D}} \right\|_2^2 - \left\| {\bf W}^{[D\!-\!1]\backstep}_{\changein :i_{D}} \right\|_\infty^2
} \!+\! H^{[D-2]} \!\right)\! 
                    \frac{\frac{\psi^{[D-2]}}{H^{[D-2]}} \frac{\phi^{[D-2]}}{H^{[D-2]}}}{\left( 1-\phidual^{[D\!-\!1]} \right) \sqrt{{{\rho}}^{[D\!-\!1]}}}
  \leq
       {\frac{1}{{t}_{\changein i_{D}}^{[D\!-\!1]\backstep 2}} \kappa^{\!-\!1} \left( 
        \frac{{t}_{\changein i_{D}}^{[D\!-\!1]\backstep 2} \bangrad}{4}
        \right)}
                    -
                    \frac{s^{[D\!-\!1]2}}{\left( 1-\phidual^{[D\!-\!1]} \right) \sqrt{{{\rho}}^{[D\!-\!1]}}} \\
\end{array}
\]
we maximise the right-side of this bound while satisfying the first bound by selecting:
\[
 \begin{array}{l}
  \bangrad
  =
  \frac{4}{\left\| {\bf t}_{\changein}^{[D-1]\backstep} \right\|_\infty^2}
  \kappa \left( \frac{1 - \psidual^{[D-1]}}{1-\chi} \right) \\
 \end{array}
\]
so the conditions become (the first condition is the positivity of the right-side of the 
above, the second enforces the inequality):
\[
 \begin{array}{rl}
  \psidual^{[D-1]}
  &\!\!\!\!<
  1-\left(1-\chi\right)\frac{s^{[D-1]2}}{\left( 1-\phidual^{[D-1]} \right) \sqrt{{{\rho}}^{[D-1]}}} \left\| {t}_{\changein}^{[D-1]\backstep} \right\|_\infty^2 \\

  \frac{\phi^{[D-2]}}{H^{[D-2]}}
  &\!\!\!\!\leq
  \frac{\frac{1}{1-\chi}{\frac{1-\psidual^{[D-1]}}{\left\| {\bf t}_{\changein}^{[D-1]\backstep} \right\|_\infty^2}}
        -
        \frac{s^{[D-1]2}}{\left( 1-\phidual^{[D-1]} \right) \sqrt{{{\rho}}^{[D-1]}}}}
       {\frac{1}{H^{[D-1]}} \mathop{\max}\limits_{i_D} \left\{ \frac{\left\| {\bf W}_{\origin :i_{D}}^{[D-1]} \right\|_2^2}{
        \frac{1}{1-\wtune} \left\| {\bf W}^{[D-1]\backstep}_{\changein :i_{D}} \right\|_2^2 - \left\| {\bf W}^{[D-1]\backstep}_{\changein :i_{D}} \right\|_\infty^2
        } + H^{[D-2]} \right\} 
        \frac{\frac{\psi^{[D-2]}}{H^{[D-2]}}}{\left( 1-\phidual^{[D-1]} \right) \sqrt{{{\rho}}^{[D-1]}}}}
 \end{array}
\]
or, equivalently:
\begin{equation}
 \begin{array}{rl}
  \bangrad
  &\!\!\!\!=
  \frac{4}{\left\| {\bf t}_{\changein}^{[D-1]\backstep} \right\|_\infty^2}
  \kappa \left( \frac{1 - \psidual^{[D-1]}}{1-\chi} \right) \\

  \psidual^{[D-1]} 
  &\!\!\!\!\in
  \left( 0,
  1-\left(1-\chi\right)\frac{s^{[D-1]2}}{\left( 1-\phidual^{[D-1]} \right) \sqrt{{{\rho}}^{[D-1]}}} \left\| {t}_{\changein}^{[D-1]\backstep} \right\|_\infty^2 
  \right) \\

  \phidual^{[D-2]}
  &\!\!\!\!\in \left(
  1 - \frac{1}{\sqrt{\rho^{[D-2]}}} \overline{\sigma}^{[D-2]-1} \left(
  \frac{\frac{1}{1-\chi} {\frac{1-\psidual^{[D-1]}}{\left\| {\bf t}_{\changein}^{[D-1]\backstep} \right\|_\infty^2}}
        -
        \frac{s^{[D-1]2}}{\left( 1-\phidual^{[D-1]} \right) \sqrt{{{\rho}}^{[D-1]}}}}
       {\frac{1}{H^{[D-1]}} \mathop{\max}\limits_{i_D} \left\{ \frac{\left\| {\bf W}_{\origin :i_{D}}^{[D-1]} \right\|_2^2}{
        \frac{1}{1-\wtune}\left\| {\bf W}^{[D-1]\backstep}_{\changein :i_{D}} \right\|_2^2 - \left\| {\bf W}^{[D-1]\backstep}_{\changein :i_{D}} \right\|_\infty^2
        } + H^{[D-2]} \right\} 
        \frac{\frac{\psi^{[D-2]}}{H^{[D-2]}}}{\left( 1-\phidual^{[D-1]} \right) \sqrt{{{\rho}}^{[D-1]}}}}
   \right),
   1
   \right)
 \end{array}
\end{equation}
and we obtain our second additional constraint:
\[
 \begin{array}{rl}
  \left\| {\bf t}_{\changein}^{[D-1]\backstep} \right\|_\infty^2 
  < 
  \frac{\left(1-\chi\right) \left( 1-\psidual^{[D-1]} \right)}{\left( \frac{s^{[D-1]2}}{\left( 1-\phidual^{[D-1]} \right) \sqrt{{{\rho}}^{[D-1]}}} \right)}
 \end{array}
\]

Next we consider layer $0 < j < D-1$.  We require that, to satisfy (\ref{eq:eqnY}), using (\ref{eq:canshadweightb}):
\[
 \begin{array}{rl}
  &
  \frac{{t}_{\changein i_{j+1}}^{[j]\backstep 2}}{{t}_{\changein i_{j+1}}^{[j]\backstep 2} 
  - 
  \frac{1}{1-\wtune} \left\| {\bf W}_{\changein}^{[j+1]\backstep} \right\|_F^2}
  {{t}_{\changein i_{j+1}}^{[j]\backstep 2}} 
  \geq
                    \frac{1}{1-\psidual^{[j]}} {t}_{\changein i_{j+1}}^{[j]\backstep 2}
  \;\;\;\;\forall i_{j+1} \\

  \Rightarrow &
  {{t}_{\changein i_{j+1}}^{[j]\backstep 2}}
  \geq
  \frac{1}{1-\psidual^{[j]}} \left( {{t}_{\changein i_{j+1}}^{[j]\backstep 2} 
  - 
  \frac{1}{1-\wtune} \left\| {\bf W}_{\changein}^{[j+1]\backstep} \right\|_F^2} \right)
  \;\;\;\;\forall i_{j+1} \\

  \Rightarrow &
  \left( \frac{1}{1-\psidual^{[j]}} - 1 \right) {{t}_{\changein i_{j+1}}^{[j]\backstep 2}}
  \leq
  \frac{1}{1-\psidual^{[j]}} \frac{1}{1-\wtune} \left\| {\bf W}_{\changein}^{[j+1]\backstep} \right\|_F^2
  \;\;\;\;\forall i_{j+1} \\

  \Rightarrow &
  \psidual^{[j]}
  \leq
  \frac{1}{1-\wtune} \frac{\left\| {\bf W}_{\changein}^{[j+1]\backstep} \right\|_F^2}{{{t}_{\changein i_{j+1}}^{[j]\backstep 2}}} 
  \;\;\;\;\forall i_{j+1} \\
 \end{array}
\]
and it suffices that:
\[
 \begin{array}{rl}
  &
  \frac{{t}_{\changein i_{j+1}}^{[j]\backstep 2}}{{t}_{\changein i_{j+1}}^{[j]\backstep 2} 
  - 
  \frac{1}{1-\wtune} \left\| {\bf W}_{\changein}^{[j+1]\backstep} \right\|_F^2}
  {{t}_{\changein i_{j+1}}^{[j]\backstep 2}} 
  \leq
                     \frac{1}{\frac{s^{[j]2}}{\left( 1-\phidual^{[j]} \right) \sqrt{{{\rho}}^{[j]}}}
                    +
                    \frac{1}{H^{[j]}} \left( \frac{\left\| {\bf W}_{\origin :i_{j+1}}^{[j]} \right\|_2^2}{
\frac{1}{1-\wtune} \left\| {\bf W}^{[j]\backstep}_{\changein :i_{j+1}} \right\|_2^2 - \left\| {\bf W}^{[j]\backstep}_{\changein :i_{j+1}} \right\|_\infty^2
} + H^{[j-1]} \right) 
                    \frac{\frac{\psi^{[j-1]}}{H^{[j-1]}} \frac{\phi^{[j-1]}}{H^{[j-1]}}}{\left( 1-\phidual^{[j]} \right) \sqrt{{{\rho}}^{[j]}}}}
  \;\;\;\;\forall i_{j+1} \\

  \Rightarrow &
                    {\frac{s^{[j]2}}{\left( 1-\phidual^{[j]} \right) \sqrt{{{\rho}}^{[j]}}}
                    +
                    \frac{1}{H^{[j]}} \left( \frac{\left\| {\bf W}_{\origin :i_{j+1}}^{[j]} \right\|_2^2}{
\frac{1}{1-\wtune} \left\| {\bf W}^{[j]\backstep}_{\changein :i_{j+1}} \right\|_2^2 - \left\| {\bf W}^{[j]\backstep}_{\changein :i_{j+1}} \right\|_\infty^2
} + H^{[j-1]} \right) 
                    \frac{\frac{\psi^{[j-1]}}{H^{[j-1]}} \frac{\phi^{[j-1]}}{H^{[j-1]}}}{\left( 1-\phidual^{[j]} \right) \sqrt{{{\rho}}^{[j]}}}}
  \leq
  \frac{{{{t}_{\changein i_{j+1}}^{[j]\backstep 2}}}
  - 
  \frac{1}{1-\wtune} {\left\| {\bf W}_{\changein}^{[j+1]\backstep} \right\|_F^2}}
       {{t}_{\changein i_{j+1}}^{[j]\backstep 4}}
  \;\;\;\;\forall i_{j+1} \\
\end{array}
\]
so it suffices to let:\footnote{Note that if $0 < a < x_{\rm min} \leq x$ then 
$\frac{x-a}{x^2}$ is minimised by setting $x = x_{\rm min}$.}
\[
 \begin{array}{rl}
  \psidual^{[j]}
  &\!\!\!\!\leq
  \frac{1}{1-\wtune} \frac{\left\| {\bf W}_{\changein}^{[j+1]\backstep} \right\|_F^2}{\left\| {{\bf t}_{\changein}^{[j]\backstep}} \right\|_\infty^2}  \\

  \frac{\phi^{[j-1]}}{H^{[j-1]}}
  &\!\!\!\!\leq
  \frac{
  \frac{1}{{\left\| {\bf t}_{\changein}^{[j]\backstep} \right\|_{\infty}^2}} {\left( 1
  - 
  \frac{1}{1-\wtune} \frac{\left\| {\bf W}_{\changein}^{[j+1]\backstep} \right\|_F^2}{\left\| {\bf t}_{\changein}^{[j]\backstep} \right\|_{\infty}^2} \right)}
   -
   \frac{s^{[j]2}}{\left( 1-\phidual^{[j]} \right) \sqrt{{{\rho}}^{[j]}}}
   }{
                    {\frac{1}{H^{[j]}} \mathop{\max}\limits_{i_{j+1}} \left\{ \frac{\left\| {\bf W}_{\origin :i_{j+1}}^{[j]} \right\|_2^2}{
\frac{1}{1-\wtune} \left\| {\bf W}^{[j]\backstep}_{\changein :i_{j+1}} \right\|_2^2 - \left\| {\bf W}^{[j]\backstep}_{\changein :i_{j+1}} \right\|_\infty^2
} + H^{[j-1]} \right\} 
                    \frac{\frac{\psi^{[j-1]}}{H^{[j-1]}} }{\left( 1-\phidual^{[j]} \right) \sqrt{{{\rho}}^{[j]}}}}
   }
 \end{array}
\]
or, equivalently:
\[
 \begin{array}{rl}
  \psidual^{[j]}
  &\!\!\!\!\in
  \left( 0, \frac{1}{1-\wtune} \frac{\left\| {\bf W}_{\changein}^{[j+1]\backstep} \right\|_F^2}{\left\| {{\bf t}_{\changein}^{[j]\backstep}} \right\|_\infty^2} \right) \\

  \phidual^{[j-1]}
  &\!\!\!\!\in
  \left(
  1 - \frac{1}{\sqrt{\rho^{[j-1]}}} \overline{\sigma}^{[j-1] -1} \left(
  \frac{
  \frac{1}{{\left\| {\bf t}_{\changein}^{[j]\backstep} \right\|_{\infty}^2}} {\left( 1
  - 
  \frac{1}{1-\wtune} \frac{\left\| {\bf W}_{\changein}^{[j+1]\backstep} \right\|_F^2}{\left\| {\bf t}_{\changein}^{[j]\backstep} \right\|_{\infty}^2} \right)}
   -
   \frac{s^{[j]2}}{\left( 1-\phidual^{[j]} \right) \sqrt{{{\rho}}^{[j]}}}
   }{
                    {\frac{1}{H^{[j]}} \mathop{\max}\limits_{i_{j+1}} \left\{ \frac{\left\| {\bf W}_{\origin :i_{j+1}}^{[j]} \right\|_2^2}{
\frac{1}{1-\wtune} \left\| {\bf W}^{[j]\backstep}_{\changein :i_{j+1}} \right\|_2^2 - \left\| {\bf W}^{[j]\backstep}_{\changein :i_{j+1}} \right\|_\infty^2
} + H^{[j-1]} \right\} 
                    \frac{\frac{\psi^{[j-1]}}{H^{[j-1]}} }{\left( 1-\phidual^{[j]} \right) \sqrt{{{\rho}}^{[j]}}}}
   }
   \right), 
   1
   \right)
 \end{array}
\]
where positivity compels us to further constrain the weight-step to ensure 
this is positive.  Using (\ref{eq:canaddcon}), positivity requires our third 
additional constraint:
\[
 \begin{array}{rl}
  \left\| {\bf t}_{\changein}^{[j]\backstep} \right\|_{\infty}^2
  <
  \frac{\degrowth^{[j+1]} \left( 1-\psidual^{[j]} \right) }{\left( \frac{s^{[j]2}}{\left( 1-\phidual^{[j]} \right) \sqrt{{{\rho}}^{[j]}}} \right)}
 \end{array}
\]

Finally we consider the case $j = 0$.  We require that:
\[
 \begin{array}{rl}
  &
  \frac{{t}_{\changein i_{1}}^{[0]\backstep 2}}{{t}_{\changein i_{1}}^{[0]\backstep 2} 
  - 
  \frac{1}{1-\wtune} \left\| {\bf W}_{\changein}^{[1]\backstep} \right\|_F^2}
  {{t}_{\changein i_{1}}^{[0]\backstep 2}} 
  \geq
                    \frac{1}{1-\psidual^{[0]}} {t}_{\changein i_{1}}^{[0]\backstep 2}
  \;\;\;\;\forall i_{1} \\

  \Rightarrow &
  \psidual^{[0]}
  \leq
  \frac{1}{1-\wtune} \frac{\left\| {\bf W}_{\changein}^{[1]\backstep} \right\|_F^2}{{{t}_{\changein i_{1}}^{[0]\backstep 2}}} 
  \;\;\;\;\forall i_{1} \\
 \end{array}
\]
and:
\[
 \begin{array}{rl}
  &
  \frac{{t}_{\changein i_{1}}^{[0]\backstep 2}}{{t}_{\changein i_{1}}^{[0]\backstep 2} 
  - 
  \frac{1}{1-\wtune} \left\| {\bf W}_{\changein}^{[1]\backstep} \right\|_F^2}
  {{t}_{\changein i_{1}}^{[0]\backstep 2}} 
  \leq
                     \frac{1}{\left( \frac{s^{[0]2}}{\left( 1-\phidual^{[0]} \right) \sqrt{\rho^{[0]}}} \right)}
  \;\;\;\;\forall i_{1} \\

  \Rightarrow &
                    \frac{s^{[0]2}}{\left( 1-\phidual^{[0]} \right) \sqrt{\rho^{[0]}}}
  \leq
  \frac{1
  - 
  \frac{1}{1-\wtune} \frac{\left\| {\bf W}_{\changein}^{[1]\backstep} \right\|_F^2}{{t}_{\changein i_{1}}^{[0]\backstep 2}}}
       {{t}_{\changein i_{1}}^{[0]\backstep 2}}
  \;\;\;\;\forall i_{1} \\

  \Rightarrow &
  \left\| {{\bf t}_{\changein}^{[0]\backstep}} \right\|_\infty^2
  \leq
  \frac{1
  - 
  \frac{1}{1-\wtune} \frac{\left\| {\bf W}_{\changein}^{[1]\backstep} \right\|_F^2}{\left\| {\bf t}_{\changein}^{[0]\backstep} \right\|_\infty^2}}
       {\left( \frac{s^{[0]2}}{\left( 1-\phidual^{[0]} \right) \sqrt{\rho^{[0]}}} \right)}
 \end{array}
\]
so it suffices to let:
\[
 \begin{array}{rl}
  \psidual^{[0]}
  &\!\!\!\!\leq
  \frac{1}{1-\wtune} \frac{\left\| {\bf W}_{\changein}^{[1]\backstep} \right\|_F^2}{\left\| {{\bf t}_{\changein}^{[0]\backstep}} \right\|_\infty^2}  \\
 \end{array}
\]
or, equivalently:
\[
 \begin{array}{rl}
  \psidual^{[0]}
  &\!\!\!\!\in
  \left( 0, \frac{1}{1-\wtune} \frac{\left\| {\bf W}_{\changein}^{[1]\backstep} \right\|_F^2}{\left\| {{\bf t}_{\changein}^{[0]\backstep}} \right\|_\infty^2} \right) \\
 \end{array}
\]
and our fourth additional constraint:
\[
 \begin{array}{rl}
  \left\| {{\bf t}_{\changein}^{[0]\backstep}} \right\|_\infty^2
  \leq
  \frac{\degrowth^{[1]} \left( 1-\psidual^{[0]} \right)}
       {\left( \frac{s^{[0]2}}{\left( 1-\phidual^{[0]} \right) \sqrt{\rho^{[0]}}} \right)} \\
 \end{array}
\]

There is one last technicality.  Selecting
\[
 \begin{array}{rl}
  \psidual^{[D-1]} 
  &\!\!\!\!\uparrow
  1-\left(1-\chi\right)\frac{s^{[D-1]2}}{\left( 1-\phidual^{[D-1]} \right) \sqrt{{{\rho}}^{[D-1]}}} \left\| {t}_{\changein}^{[D-1]\backstep} \right\|_\infty^2 
 \end{array}
\]
will result in:
\[
 \begin{array}{rl}
  \phidual^{[D-2]}
  &\!\!\!\!\uparrow
  1
 \end{array}
\]
which is not allowed as $\phidual^{[D-2]} < 1$ (that is, the range of $\phidual^{[D-2]}$ 
becomes the empty set in the limit.  Thus we tighten our contraints on $\psidual^{[D-1]}$ 
to:
\[
 \begin{array}{rl}
  \psidual^{[D-1]} 
  &\!\!\!\!\in
  \left( 0,
  1-\frac{s^{[D-1]2}}{\left( 1-\phidual^{[D-1]} \right) \sqrt{{{\rho}}^{[D-1]}}} \left\| {t}_{\changein}^{[D-1]\backstep} \right\|_\infty^2 
  \right) \\
 \end{array}
\]
which completes the proof after some minor refactoring.
\end{proof}

From which we arrive at the corollary used in the main body of the paper:
\begin{cor_canexistgen_simple}
 Let $\epsilon, \chi \in (0,1)$ and for a given neural 
 network with initial weights ${\bf W}_\origin$, and let ${\bf 
 W}_\changein^\backstep$ be the weight-step for this derived from 
 back-propagation.  Let:
 \[
  %
 \]
 If the weight-step satisfies:
 \[
  \begin{array}{rl}
   \left\| {\bf t}_{\changein}^{[j]\backstep} \right\|_{\infty}^2
   &\!\!\!\!<
   \frac{\left( 1-\chi \right)^2 \left( 1 - \chi^{D-j-1} {1\!\!\,\;\!\!\!1}_{j<D-1} \psidual \right)}{\left( \frac{s^{[j]2}}{\left( 1-\phidual^{[j]} \right) \sqrt{{{\rho}}^{[j]}}} \right)}
   \;\;\;\; \forall j \in \infset{N}_D \\
  \end{array}
 \]
 then there exists of a canonical scaling:
 \[
  \begin{array}{rl}
   \left. \frac{\partial}{\partial {\bf W}_\changein} \left\| {\bg{\Psi}}_\origin \left( {\bf W}_\changein \right) \right\|_{\mathcal{W}_\origin} \right|_{{\bf W}_\changein = {\bf W}_\changein^\backstep} = \bangrad {\bf W}_\changein^\backstep
  \end{array}
 \]
 where:
 \[
  \begin{array}{l}
   \bangrad
   =
   \frac{4}{\left\| {\bf t}_{\changein}^{[D-1]\backstep} \right\|_\infty^2}
   \kappa \left( \frac{1 - \psidual}{1-\chi} \right) \\
  \end{array}
 \]
 satisfying $\| {\bg{\Phi}}_\origin ({\bf x}) \|_F^2 \leq H^{[D-1]} \overline{\sigma}^{[D-1]} 
 ( \left( 1-\epsilon ) \sqrt{{\rho}^{[D-1]}} \right)$ $\forall {\bf x} \in \mathbb{X}$, $\| {{\bg{\Psi}}}_\origin 
 ({\bf W}_\changein) \|_F^2 \leq H^{[D-1]} \frac{1-\psidual}{\psidual}$.  
 \label{cor:canexistgen_simple}
\end{cor_canexistgen_simple}
\begin{proof}
We begin with theorem \ref{th:canexistgen}.  Let $\degrowth^{[j]} = 1-\chi$ and 
$\phidual^{[D-1]} = \phidual$.  The definitions in theorem \ref{th:canexistgen} become:
\[
 %
}}
\]
and subsequently for some $\alpha^{[j]} \in \infset{R}_+$:
\[
 \begin{array}{rl}
  \left\| {\bf W}_{\changein}^{[j+1]\backstep} \right\|_F^2
  =
  \left( 1 - \wtune \right) \chi \left\| {\bf t}_{\changein}^{[j]\backstep} \right\|_{\infty}^2 \\
 \end{array}
\]
The condition on the weight-step satisfies:
\[
 \begin{array}{rl}
  \left\| {\bf t}_{\changein}^{[j]\backstep} \right\|_{\infty}^2
  &\!\!\!\!<
  \frac{\left( 1 - \chi \right) \left( 1-\psidual^{[j]} \right)}{\left( \frac{s^{[j]2}}{\left( 1-\phidual^{[j]} \right) \sqrt{{{\rho}}^{[j]}}} \right)}
  \;\;\;\; \forall j \in \infset{N}_D
 \end{array}
\]
implying the existence of a canonical scaling:
\[
 \begin{array}{rl}
  \left. \frac{\partial}{\partial {\bf W}_\changein} \left\| {\bg{\Psi}}_\origin \left( {\bf W}_\changein \right) \right\|_{\mathcal{W}_\origin} \right|_{{\bf W}_\changein = {\bf W}_\changein^\backstep} = \bangrad {\bf W}_\changein^\backstep
 \end{array}
\]
where:
\[
 \begin{array}{l}
  \bangrad
  =
  \frac{4}{\left\| {\bf t}_{\changein}^{[D-1]\backstep} \right\|_\infty^2}
  \kappa \left( \frac{1 - \psidual^{[D-1]}}{1-\chi} \right) \\
 \end{array}
\]
satisfying $\| {\bg{\Phi}}_\origin ({\bf x}) \|_F^2 \leq \phi^{[D-1]}$, $\| 
{{\bg{\Psi}}}_\origin ({\bf W}_\changein) \|_F^2 \leq \psi^{[D-1]}$.  To 
further simplify the constraints, note that:
\[
 \begin{array}{l}
  1 - \psidual^{[j]}
  =
  1 - \chi^{D-1-j} + \chi^{D-1-j} \frac{s^{[D-1]2}}{\left( 1-\phidual \right) \sqrt{{{\rho}}^{[D-1]}}} \left\| {\bf t}_{\changein}^{[D-1]\backstep} \right\|_\infty^2 
  >
  1 - \chi^{D-1-j} \\

  \frac{\psi^{[j]}}{H^{[j]}}
  =
  \frac{1-\psidual^{[j]}}{\psidual^{[j]}}
  <
  \frac{1}{\psidual^{[j]}}
  =
  \frac{1}{\chi^{D-1-j}} \frac{1}{1-\frac{s^{[D-1]2}}{\left( 1-\phidual \right) \sqrt{{{\rho}}^{[D-1]}}} \left\| {\bf t}_{\changein}^{[D-1]\backstep} \right\|_\infty^2}
 \end{array}
\]
so we can tighten the constraints on $\phidual^{[j]}$ and $\| {\bf 
t}_{\changein}^{[j]\backstep} \|_{\infty}^2$ to:
\[
 %
 \right. \\

  \left\| {\bf t}_{\changein}^{[j]\backstep} \right\|_{\infty}^2
  &\!\!\!\!<
  \frac{\left( 1 - \chi \right) \left( 1 - \chi^{D-1-j} \left( 1-\frac{s^{[D-1]2}}{\left( 1-\phidual \right) \sqrt{{{\rho}}^{[D-1]}}} \left\| {\bf t}_{\changein}^{[D-1]\backstep} \right\|_\infty^2 \right) \right)}{\left( \frac{s^{[j]2}}{\left( 1-\phidual^{[j]} \right) \sqrt{{{\rho}}^{[j]}}} \right)}
  \;\;\;\; \forall j \in \infset{N}_D \\
 \end{array}
\]

Subsequently the definitions in theorem \ref{th:canexistgen} can be tightened to, 
being careful to ensure that $\left\| {\bf t}_{\changein}^{[j]\backstep} 
\right\|_{\infty}^2$ is not overconstrained:
\[
 \begin{array}{c}
 \begin{array}{rl}
  \psidual 
  &\!\!\!\!= 
  \left( 1-\frac{1}{1-\chi}\frac{s^{[D-1]2}}{\left( 1-\phidual \right) \sqrt{{{\rho}}^{[D-1]}}} \left\| {\bf t}_{\changein}^{[D-1]\backstep} \right\|_\infty^2 \right) \\

  \phidual^{[j]}
  &\!\!\!\!=
  \left\{ \begin{array}{ll}
     \phidual
  & \mbox{if } j = D-1 \\ \\

     1 - \frac{1}{\sqrt{\rho^{[j]}}} \overline{\sigma}^{[j]-1} \left(
     \frac{{\frac{1}{1-\chi} \frac{\chi^{D-1-j} \psidual \left( 1-\phidual^{[j+1]} \right) \sqrt{\rho^{[j+1]}}}{\left\| {\bf t}_{\changein}^{[j+1]\backstep} \right\|_\infty^2}}
           -
           \frac{\chi^{D-1-j} \psidual}{1-\chi^{D-1-j} \psidual} s^{[j+1]2} }
          {\mathop{\max}\limits_{i_{j+2}} \left\{ \frac{\frac{1}{H^{[j+1]}} \left\| {\bf W}_{\origin :i_{j+2}}^{[j+1]} \right\|_2^2}{
           \frac{1}{1-\wtune} \left\| {\bf W}^{[j+1]\backstep}_{\changein :i_{j+2}} \right\|_2^2 - \left\| {\bf W}^{[j+1]\backstep}_{\changein :i_{j+2}} \right\|_\infty^2
           } + \frac{H^{[j]}}{H^{[j+1]}} \right\} }
      \right)

  & \mbox{otherwise} \\
  \end{array} \right. \\
 \end{array} \;\;\;\; \forall j \in \infset{N}_D \\
 \end{array}
\]
and subsequently for some $\alpha^{[j]} \in \infset{R}_+$:
\[
 \begin{array}{rl}
  \left\| {\bf W}_{\changein}^{[j+1]\backstep} \right\|_F^2
  =
  \left( 1 - \wtune \right) \chi \left\| {\bf t}_{\changein}^{[j]\backstep} \right\|_{\infty}^2 \\
 \end{array}
\]
The condition on the weight-step satisfies (the first constraint is required to 
ensure that $\psidual > 0$):
\[
 \begin{array}{rl}
  \left\| {\bf t}_{\changein}^{[j]\backstep} \right\|_{\infty}^2
  &\!\!\!\!<
  \frac{1-\chi}{\left( \frac{s^{[j]2}}{\left( 1-\phidual^{[j]} \right) \sqrt{{{\rho}}^{[j]}}} \right)}
  \left\{ \begin{array}{ll}
  1
  & \mbox{if } j = D-1 \\ \\
  \left( 1 - \chi^{D-1-j} \psidual \right)
  & \mbox{otherwise} \\
  \end{array} \right.
  \;\;\;\; \forall j \in \infset{N}_D \\
 \end{array}
\]
implying the existence of a canonical scaling:
\[
 \begin{array}{rl}
  \left. \frac{\partial}{\partial {\bf W}_\changein} \left\| {\bg{\Psi}}_\origin \left( {\bf W}_\changein \right) \right\|_{\mathcal{W}_\origin} \right|_{{\bf W}_\changein = {\bf W}_\changein^\backstep} = \bangrad {\bf W}_\changein^\backstep
 \end{array}
\]
where:
\[
 \begin{array}{l}
  \bangrad
  =
  \frac{4}{\left\| {\bf t}_{\changein}^{[D-1]\backstep} \right\|_\infty^2}
  \kappa \left( \frac{1 - \psidual}{1-\chi} \right) \\
 \end{array}
\]
satisfying $\| {\bg{\Phi}}_\origin ({\bf x}) \|_F^2 \leq H^{[D-1]} \overline{\sigma}^{[D-1]} 
( \left( 1-\phidual ) \sqrt{{\rho}^{[D-1]}} \right)$, $\| {{\bg{\Psi}}}_\origin 
({\bf W}_\changein) \|_F^2 \leq H^{[D-1]} \frac{1-\psidual}{\psidual}$.

At this point we have enough leeway in our construct to let $\wtune \to 0$, and 
tidy up with $\phidual = \epsilon$, so our constraints become:
\[
 \begin{array}{c}
 \begin{array}{rl}
  \psidual 
  &\!\!\!\!= 
  \left( 1-\frac{1}{1-\chi}\frac{s^{[D-1]2}}{\left( 1-\epsilon \right) \sqrt{{{\rho}}^{[D-1]}}} \left\| {\bf t}_{\changein}^{[D-1]\backstep} \right\|_\infty^2 \right) \\

  \phidual^{[j]}
  &\!\!\!\!=
  \left\{ \begin{array}{ll}
     \epsilon
  & \mbox{if } j = D-1 \\ \\

     1 - \frac{1}{\sqrt{\rho^{[j]}}} \overline{\sigma}^{[j]-1} \left(
     \frac{{\frac{1}{1-\chi} \frac{\chi^{D-1-j} \psidual \left( 1-\phidual^{[j+1]} \right) \sqrt{\rho^{[j+1]}}}{\left\| {\bf t}_{\changein}^{[j+1]\backstep} \right\|_\infty^2}}
           -
           \frac{\chi^{D-1-j} \psidual}{1-\chi^{D-1-j} \psidual} s^{[j+1]2} }
          {\mathop{\max}\limits_{i_{j+2}} \left\{ \frac{\frac{1}{H^{[j+1]}} \left\| {\bf W}_{\origin :i_{j+2}}^{[j+1]} \right\|_2^2}{
           \frac{1}{1-\wtune} \left\| {\bf W}^{[j+1]\backstep}_{\changein :i_{j+2}} \right\|_2^2 - \left\| {\bf W}^{[j+1]\backstep}_{\changein :i_{j+2}} \right\|_\infty^2
           } + \frac{H^{[j]}}{H^{[j+1]}} \right\} }
      \right)

  & \mbox{otherwise} \\
  \end{array} \right. \\
 \end{array} \;\;\;\; \forall j \in \infset{N}_D \\
 \end{array}
\]
and subsequently for some $\alpha^{[j]} \in \infset{R}_+$:
\[
 \begin{array}{rl}
  \left\| {\bf W}_{\changein}^{[j+1]\backstep} \right\|_F^2
  =
  \left( 1 - \wtune \right) \chi \left\| {\bf t}_{\changein}^{[j]\backstep} \right\|_{\infty}^2 \\
 \end{array}
\]
The condition on the weight-step satisfies (the first constraint is required to 
ensure that $\psidual > 0$):
\[
 \begin{array}{rl}
  \left\| {\bf t}_{\changein}^{[j]\backstep} \right\|_{\infty}^2
  &\!\!\!\!<
  \frac{1-\chi}{\left( \frac{s^{[j]2}}{\left( 1-\phidual^{[j]} \right) \sqrt{{{\rho}}^{[j]}}} \right)}
  \left\{ \begin{array}{ll}
  1
  & \mbox{if } j = D-1 \\ \\
  \left( 1 - \chi^{D-1-j} \psidual \right)
  & \mbox{otherwise} \\
  \end{array} \right.
  \;\;\;\; \forall j \in \infset{N}_D \\
 \end{array}
\]
implying the existence of a canonical scaling:
\[
 \begin{array}{rl}
  \left. \frac{\partial}{\partial {\bf W}_\changein} \left\| {\bg{\Psi}}_\origin \left( {\bf W}_\changein \right) \right\|_{\mathcal{W}_\origin} \right|_{{\bf W}_\changein = {\bf W}_\changein^\backstep} = \bangrad {\bf W}_\changein^\backstep
 \end{array}
\]
where:
\[
 \begin{array}{l}
  \bangrad
  =
  \frac{4}{\left\| {\bf t}_{\changein}^{[D-1]\backstep} \right\|_\infty^2}
  \kappa \left( \frac{1 - \psidual}{1-\chi} \right) \\
 \end{array}
\]
satisfying $\| {\bg{\Phi}}_\origin ({\bf x}) \|_F^2 \leq H^{[D-1]} \overline{\sigma}^{[D-1]} 
( \left( 1-\epsilon ) \sqrt{{\rho}^{[D-1]}} \right)$, $\| {{\bg{\Psi}}}_\origin 
({\bf W}_\changein) \|_F^2 \leq H^{[D-1]} \frac{1-\psidual}{\psidual}$.

Finally, we need to ensure that $\bangrad$ is well defined, for which we 
require that $1-\psidual < 1-\chi$ or, equivalently, $\psidual > \chi$.  For 
This is suffices that:
\[
 \begin{array}{rl}
  \left\| {\bf t}_{\changein}^{[D-1]\backstep} \right\|_{\infty}^2
  &\!\!\!\!<
  \frac{1-\chi}{\left( \frac{s^{[D-1]2}}{\left( 1-\phidual^{[D-1]} \right) \sqrt{{{\rho}}^{[D-1]}}} \right)}
 \end{array}
\]
and so, tightening bounds slightly:
\[
 \begin{array}{rl}
  \left\| {\bf t}_{\changein}^{[j]\backstep} \right\|_{\infty}^2
  &\!\!\!\!<
  \frac{1-\chi}{\left( \frac{s^{[j]2}}{\left( 1-\phidual^{[j]} \right) \sqrt{{{\rho}}^{[j]}}} \right)}
  \left\{ \begin{array}{ll}
  \left( 1 - \chi \right)
  & \mbox{if } j = D-1 \\ \\
  \left( 1 - \chi^{D-1-j} \psidual \right)
  & \mbox{otherwise} \\
  \end{array} \right.
  \;\;\;\; \forall j \in \infset{N}_D \\
 \end{array}
\]
which completes the proof.
\end{proof}

Subsequently we obtain a bound on Rademacher complexity:
\begin{th_onerad}
 Let $\epsilon, \chi \in (0,1)$ and for a given neural 
 network with initial weights ${\bf W}_\origin$, and let ${\bf 
 W}_\changein^\backstep$ be the weight-step for this derived from 
 back-propagation satisfying the conditions set out in corollary 
 \ref{cor:canexistgen_simple}.  Then $f_\changein^\backstep \in \mathcal{F}^\regstep$, where the 
 Rademacher complexity of $\mathcal{F}^\regstep$ is bounded as:
\[
 \begin{array}{l}
  \mathcal{R}_N \left( \mathcal{F}^\regstep \right) 
  \leq
  H^{[D-1]} \sqrt{\frac{1}{N} {\frac{\left\| {\bf t}_{\changein}^{[D-1]\backstep} \right\|_\infty^2}
   {\frac{1}{1-\chi} \frac{\left( 1-\chi \right)^2}{\left({\frac{s^{[D-1]2}}{\left( 1-\epsilon \right) \sqrt{{{\rho}}^{[D-1]}}}} \right)} - \left\| {\bf t}_{\changein}^{[D-1]\backstep} \right\|_\infty^2} \bar{\sigma}^{[D-1]} \left( \left( 1-\epsilon \right) \sqrt{\rho^{[D-1]}} \right)}} 
 \end{array}
\]
 \label{th:onerad}
\end{th_onerad}
\begin{proof}
By corollary \ref{cor:canexistgen_simple} we know that, subject to assumptions on 
the size of the weight-step, there must exist scale factors, shadow weighs and 
regularisation parameter $\lambda$ such that the change in neural network 
operation $f_\changein^\backstep$ due to backpropagation and the change 
in neural network operation $f_\changein^\regstep$ resulting from 
minimisation of the regularised risk $R_\lambda$ will coincide, so 
$f_\changein^\backstep = f_\changein^\regstep \in \mathcal{B}_\origin$.  
Fixing these parameters, moreover, we know from corollary \ref{cor:canexistgen_simple} 
that $\| {\bg{\Phi}}_\origin ({\bf x}) \|_F^2 \leq H^{[D-1]} \overline{\sigma}^{[D-1]} 
( \left( 1-\epsilon ) \sqrt{{\rho}^{[D-1]}} \right)$ $\forall {\bf x} \in 
\mathbb{X}$, $\| {{\bg{\Psi}}}_\origin ({\bf W}_\changein) \|_F^2 \leq 
H^{[D-1]} \frac{1-\psidual}{\psidual}$, so: 
\[
 \begin{array}{rl}
  f_\changein^\regstep \in \mathcal{F}^\regstep = \left\{ \left. \left< {\bg{\Phi}}_\origin \left( \cdot \right), {\bg{\Omega}} \right> \right| \left\| {\bg{\Omega}} \right\|_F^2 \leq H^{[D-1]} \frac{1-\psidual}{\psidual} \right\}
 \end{array}
\]
Hence for a Rademacher random variable $\epsilon$ the Rademacher complexity is 
bounded as follows:
\[
 \begin{array}{l}
  \mathcal{R}_N \left( \mathcal{F}^\regstep \right) 
  =
  \expect_\nu \expect_\epsilon \left[ \mathop{\sup}\limits_{f_\changein \in \mathcal{F}^\regstep} \left| \frac{1}{N} \sum_i \epsilon_i f \left( {\bf x}_i \right) \right| \right] \\
  = 
  \expect_\nu \expect_\epsilon \left[ \mathop{\sup}\limits_{f_\changein \in \mathcal{F}^\regstep} \left| \frac{1}{N} \sum_i \epsilon_i \left< {\bg{\Phi}}_\origin \left( {\bf x}_i \right), {\bg{\Psi}} \left( {\bf W}_\changein^\backstep \right) \right> \right| \right] \\
  \leq^{C.S.}
  \expect_\nu \left[ \frac{1}{N} \expect_\epsilon \left[ \mathop{\sup}\limits_{f_\changein \in \mathcal{F}^\regstep} \left\| \sum_i \epsilon_i {\bg{\Phi}}_\origin \left( {\bf x}_i \right) \right\|_F \left\| {\bg{\Psi}} \left( {\bf W}_\changein^\backstep \right) \right\|_F \right] \right] \\
  \leq
  \expect_\nu \left[ \sqrt{\frac{1}{N}} \sqrt{ H^{[D-1]} \frac{1-\psidual}{\psidual} \expect_\epsilon \left[ \left\| \frac{1}{N} \sum_i \epsilon_i {\bg{\Phi}}_\origin \left( {\bf x}_i \right) \right\|_F^2 \right]} \right] \\
  \leq^{\rm Jensen}
  \expect_\nu \left[ \sqrt{\frac{1}{N} H^{[D-1]} \frac{1-\psidual}{\psidual}} \sqrt{ \frac{1}{N} \sum_i \left\| {\bg{\Phi}}_\origin \left( {\bf x}_i \right) \right\|_F^2} \right] \\
  \leq
  H^{[D-1]} \sqrt{\frac{1}{N} {\frac{1-\psidual}{\psidual} \bar{\sigma}^{[j]} \left( \left( 1-\epsilon \right) \sqrt{\rho^{[D-1]}} \right)}} \\
 \end{array}
\]
independent of the data distribution $\nu$.

Next recall the definitions from corollary \ref{cor:canexistgen_simple}:
\[
 \begin{array}{rl}
   \psidual 
   &\!\!\!\!= 
   1-\frac{1}{1-\chi}\frac{s^{[D-1]2}}{\left( 1-\epsilon \right) \sqrt{{{\rho}}^{[D-1]}}} \left\| {\bf t}_{\changein}^{[D-1]\backstep} \right\|_\infty^2 \\
 \end{array}
\]
so that:
\[
 \begin{array}{rl}
   \frac{1-\psidual}{\psidual} 
   &\!\!\!\!= 
   \frac{\left\| {\bf t}_{\changein}^{[D-1]\backstep} \right\|_\infty^2}
   {\frac{1}{1-\chi} \frac{\left( 1-\chi \right)^2}{\left({\frac{s^{[D-1]2}}{\left( 1-\epsilon \right) \sqrt{{{\rho}}^{[D-1]}}}} \right)} - \left\| {\bf t}_{\changein}^{[D-1]\backstep} \right\|_\infty^2}
 \end{array}
\]
and hence:
\[
 \begin{array}{l}
  \mathcal{R}_N \left( \mathcal{F}^\regstep \right) 
  \leq
  H^{[D-1]} \sqrt{\frac{1}{N} {\frac{\left\| {\bf t}_{\changein}^{[D-1]\backstep} \right\|_\infty^2}
   {\frac{1}{1-\chi} \frac{\left( 1-\chi \right)^2}{\left({\frac{s^{[D-1]2}}{\left( 1-\epsilon \right) \sqrt{{{\rho}}^{[D-1]}}}} \right)} - \left\| {\bf t}_{\changein}^{[D-1]\backstep} \right\|_\infty^2} \bar{\sigma}^{[j]} \left( \left( 1-\epsilon \right) \sqrt{\rho^{[D-1]}} \right)}} 
 \end{array}
\]
\end{proof}

\section{Neural Networks and Reproducing Kernel Banach Spaces}

A reproducing kernel Hilbert space (RKHS) \citep{Aro1} is a Hilbert spaces 
$\mathcal{H}$ of functions $f : \infset{X} \to \infset{Y}$ for which the 
point evaluation functions are continuous.  Thus, applying the Riesz 
representor theorem, there exists a kernel $K$ such that:
\[
 \begin{array}{l}
  f \left( {\bf x} \right) = \left< f \left( \cdot \right), K \left( {\bf x}, \cdot \right) \right>_\mathcal{H}
 \end{array}
\]
for all $f \in \mathcal{H}$.  Subsequently $K({\bf x}, {\bf x}') = \left< 
K({\bf x},\cdot), K({\bf x}',\cdot) \right>$ and, by the Moore-Aronszajn 
theorem, $K$ is uniquely defined by $\mathcal{H}$ and vice-versa.  $K$ is 
called the reproducing kernel, and the corresponding RKHS is denoted 
$\mathcal{H}_K$.  RKHSs have gained popularity in machine learning because 
they are well suited to many aspects of machine learning 
\citep{Ste3,Sha3,Cor1,Cho7,Cri4,Gen2,Gon7,Her2,Li7,Mul2,Smo12}.  The inner 
product structure enables the kernel trick, which is of great practical use, 
and the kernel itself is readily understood as a similarity measure.  More 
importantly here, the structure of RKHSs has led to a rich framework of 
complexity analysis and generalisation bounds \citep{Ste3,Sha3} that form a 
foundation for this branch of machine learning, which more recently has been 
extended to neural networks through the theory of neural tangent kernels 
\citep{Jac2}.

Reproducing kernel Banach spaces (RKBSs) are a generalisationn of RKHSs 
which start with a Banach space of functions rather than a Hilbert space 
\citep{Der1,Lin10,Zha11,Zha14,Son1,Sri3,Xu4}.  Precisely:
\begin{def_rkbs}[Reproducing kernel Banach space (RKBS, \citep{Lin10})]
 A reproducing kernel Banach space $\mathcal{B}$ on a set $\infset{X}$ is a 
 Banach space of functions $f : \infset{X} \to \infset{Y}$ such that every 
 point evaluation $\delta_{\bf x} : \mathcal{B} \to \infset{Y}$, ${\bf x} \in 
 \infset{X}$, on $\mathcal{B}$ is continuous.  That is, $\exists C_{\bf x} \in 
 \infset{R}_+$ such that:
 \[
  \begin{array}{l}
   \left| \delta_{\bf x} \left( f \right) \right| = \left| f \left( {\bf x} \right) \right| \leq C_{\bf x} \left\| f \right\|_{\mathcal{B}}
  \end{array}
 \]
 for all $f \in \mathcal{B}$.
 \label{def:rkbs}
\end{def_rkbs}

\begin{figure}
\begin{center}
\begin{tabular}{|l||l|l|}
 \hline 
 & Notation in present Paper & Notation used in \citep{Lin10} \\
 \hline 
 \hline 
 Data space:        & $\infset{X}         \subset \infset{R}^n$                                                                         & $\Omega_1$ (input space)  \\
 \hline 
 Weight-step space: & $\infset{W}_\origin \subset \prod_{j \in \infset{N}_D} \infset{R}^{H^{[j-1]} \times H^{[j]}} \times \infset{R}^{H^{[j]}}$ & $\Omega_2$ (weight space) \\
 \hline 
 Data Feature map:        & ${\bg{\Phi}}_\origin : \infset{X}         \to \mathcal{X}_\origin \subset \infset{R}^{\infty \times m}$ & $\Phi_1 : \Omega_1 \to \mathcal{W}_1$ \\
 \hline
 Weight-step feature map: & ${\bg{\Psi}}_\origin : \infset{W}_\origin \to \mathcal{W}_\origin \subset \infset{R}^{\infty \times m}$ & $\Phi_2 : \Omega_2 \to \mathcal{W}_2$ \\
 \hline
 Data Banach space:        & $\mathcal{X}_\origin = {{\rm span}} ({\bg{\Phi}}_{\origin} (\infset{X}))         \subset \infset{R}^{\infty \times m}$, where & $\mathcal{W}_1$ with norm $\| \cdot \|_{\mathcal{W}_1}$ \\
                           & $\| \cdot \|_{\mathcal{X}_\origin} = \| \cdot \|_{F}$                                                                       &                                                         \\
 \hline
 Weight-step Banach space: & $\mathcal{W}_\origin = {{\rm span}} ({\bg{\Psi}}_{\origin} (\infset{W}_\origin)) \subset \infset{R}^{\infty \times m}$, where & $\mathcal{W}_2$ with norm $\| \cdot \|_{\mathcal{W}_2}$ \\
                           & $\| \cdot \|_{\mathcal{W}_\origin} = \| \cdot \|_{F}$                                                                       &                                                         \\
 \hline
 Bilinear form: & $\left< \cdot, \cdot \right>_{\mathcal{X}_\origin \times \mathcal{W}_\origin} : \infset{R}^{\infty \times m} \times \infset{R}^{\infty \times m} \to \infset{R}^m$,   & $\left< \cdot, \cdot \right>_{{\mathcal{W}}_1 \times {\mathcal{W}}_2} : \mathcal{W}_1 \times \mathcal{W}_2 \to \infset{Y}$ \\
                & $\left< {{\bg{\Omega}}}, {{\bg{\Xi}}} \right>_{\mathcal{X}_\origin \times \mathcal{W}_\origin} = {\rm diag} \left( {{\bg{\Omega}}}^\tsp {{\bg{\Xi}}} \right)$ & \\
 \hline
\end{tabular}
\end{center}
\caption{Summary of the construction of reproducing kernel Banach space as per 
         \citep{Lin10}.}
\label{tab:banconstruct}
\end{figure}

This introduces a richer set of geometrical structures and allows for new 
and exciting extensions to the usual kernel framework, such as asymmetric 
kernels, sparse learning in Feature space, lasso in statistics and 
$m$-kernels.  In the present context it allows us to extend NTKs from a 
first-order approximation in the overparametrised regime to an exact 
representation without the need for infinite width approximations.  
There are many approaches to RKBS theory in the literature, but in the 
present context we find the method of \citep{Lin10} most helpful.

In this formulation, reproducing kernels may be constructed from a set of 
basic ingredients that closely mirror our construction of ${\bf f}_\changein$ 
in the preceeding sections.  In particular, as per \citep{Lin10}, and in light 
of lemma \ref{lem:densekey}, we can construct RKBS ${\mathcal{B}}_\origin$ 
containing ${\bf f}_\changein$ using the ingredients defined in Figure 
\ref{tab:banconstruct}.  This gives us the reproducing kernel Banach space 
${{\mathcal{B}}}_\origin$ on $\infset{X}$:
\begin{equation}
 \begin{array}{rl}
  {{\mathcal{B}}}_\origin = \left\{ \left. \left< {{\bg{\Phi}}}_\origin \left( \cdot \right), {\bg{\Omega}} \right>_{\mathcal{X}_\origin \times \mathcal{W}_\origin} \right| {\bg{\Omega}} \in {\mathcal{W}_\origin} \right\}\!, 
  \mbox{ where }
  \left\| \left< {{\bg{\Phi}}}_\origin \left( \cdot \right), {\bg{\Omega}} \right>_{\mathcal{X}_\origin \times \mathcal{W}_\origin} \right\|_{{{\mathcal{B}}}_\origin} = \left\| {\bg{\Omega}} \right\|_{\mathcal{W}_\origin} \\
 \end{array}
 \label{eq:rkbs_main}
\end{equation}
the elements of ${{\mathcal{B}}}_\origin$ include the functions ${\bf 
f}_\changein : \infset{R}^n \to \infset{R}^m$ - that is, the change in the 
neural network behaviour due to a change ${\bf W}_\changein \in 
\infset{W}_\origin$ in weights and biases during one iteration, subject to 
convergence conditions - as well all linear combinations thereof (recalling 
that $\mathcal{W}_\origin = {{\rm span}} ({{\bg{\Psi}}}_{\origin} 
(\infset{W}_\origin))$).  This has the reproducing kernel ${\bf K}_\origin : 
\infset{X} \times \infset{W}_\origin \to \infset{R}^{m \times m}$:
\begin{equation}
\begin{array}{l}
  {\bf K}_\origin \left( {\bf x}, {\bf W}_\changein \right) 
  = {\rm diag} \left( \left< {{\bg{\Phi}}}_\origin \left( {\bf x} \right), {{\bg{\Psi}}}_{\origin} \left( {\bf W}_\changein \right) \right>_{\mathcal{X}_\origin \times \mathcal{W}_\origin}  \right)
  = {\rm diag} \left( {\bf f}_\changein \left( {\bf x} \right) \right)
 \end{array}
 \label{eq:define_k}
\end{equation}
We also obtain the corresponding reproducing kernel Banach space 
${{\mathcal{B}}}_\origin^*$ on $\infset{W}_\origin$:
\begin{equation}
 \begin{array}{rl}
  {{\mathcal{B}}}_\origin^* = \left\{ \left. \left< {\bg{\Xi}}, {{\bg{\Psi}}}_\origin \left( \cdot \right) \right>_{\mathcal{X}_\origin \times \mathcal{W}_\origin} \right| {\bg{\Xi}} \in {\mathcal{X}_\origin} \right\}\!, 
  \mbox{ where }
  \left\| \left< {\bg{\Xi}}, {{\bg{\Psi}}}_\origin \left( \cdot \right) \right>_{\mathcal{X}_\origin \times \mathcal{W}_\origin} \right\|_{{{\mathcal{B}}}_\origin^*} = \left\| {\bg{\Xi}} \right\|_{\mathcal{X}_\origin} \\
 \end{array}
 \label{eq:rkbs_dual}
\end{equation}
whose members are the evaluation functionals for $\mathcal{B}_\origin$, which 
has the reproducing kernel ${\bf K}_\origin^* : \infset{W}_\origin \times 
\infset{X} \to \infset{R}^{m \times m}$ given by:
\[
 \begin{array}{l}
  {\bf K}_\origin^* \left( {\bf W}_\changein, {\bf x} \right)
  = {\bf K}_\origin \left( {\bf x}, {\bf W}_\changein \right)
  = {\rm diag} \left( {\bf f}_\changein \left( {\bf x} \right) \right)
 \end{array}
\]

\section{Case study - The tanh Activation Function} \label{sec:tanhcase}

Let us consider the activation function:
\begin{equation}
 \begin{array}{l}
  \tau^{[j]} \left( \zeta \right) = {\rm tanh} \left( \zeta \right)
 \end{array}
\end{equation}
We wish to construct the corresponding $\sigma_{z,z'}^{[j]}$ for 
arbitrary $z,z' \in \infset{R}$.  As we shown in section 
\ref{sec:taylortanh}, the Taylor expansion of ${\rm tanh}$ about 
an arbitrary $z \in \infset{R}$ is:
\[
 \begin{array}{rl}
  {\rm tanh} \left( z + \zeta \right) 
  &\!\!\!\!= 
  {\rm tanh} \left( z \right)
  +
  \mathop{\sum}\limits_{m=1}^\infty a_{\left(z\right)m}^{\rm tanh} \zeta^m
 \end{array}
\]
where:
\[
 \begin{array}{rl}
  a_{\left(z\right)m}^{\rm tanh}

  &\!\!\!\!= 
  2
  \left( -\sgn \left( z \right) \right)^{m} 
  \mathop{\sum}\limits_{n=0}^\infty 
  \left( \frac{1}{\sqrt{z^2 + \left( \left( n + \frac{1}{2} \right) \pi \right)^2}} \right)^{m+1}
  \cos \left( {\left( m+1 \right) {\rm atan} \left( \frac{\left( n+\frac{1}{2} \right) \pi}{z} \right)} \right) \\
 \end{array}
\]
which is valid for all $z,\zeta \in \infset{R}$ such that $|\zeta| \leq 
\frac{\pi}{2}$.  Hence, by our definition (\ref{eq:euceuc_lambda}), using this 
expansion:
\begin{equation}
 \begin{array}{l}
  \sigma_{\left( z,z' \right)}^{[j]} \left( \zeta \right) 
  = 
  \mathop{\sum}\limits_{m=1}^\infty
  a_{\left(z\right)m}^{\rm tanh} a_{\left(z'\right)m}^{\rm tanh}
  \zeta^m \\
 \end{array}
\label{eq:lambda_gen_series}
\end{equation}
In the case $z = z' = 0$ this simplifies to:
\[
 \begin{array}{l}
  a_{\left(0\right)m}^{\rm tanh} 
  = 
  \frac{2^{2m} \left( 2^{2m} - 1 \right) B_{2m}}{\left( 2m \right)!}
 \end{array}
\]
where $B_{2m}$ are the Bernoulli numbers.  So:
\[
 \begin{array}{l}
  \sigma_{\left( 0,0 \right)}^{[j]} \left( \zeta \right) 
  = 
  \mathop{\sum}\limits_{m=1}^\infty
  \left( \frac{2^{2m} \left( 2^{2m} - 1 \right) B_{2m}}{\left( 2m \right)!} \right)^2
  \zeta^m 
 \end{array}
\]

\subsection{The Taylor Expansions of ${\rm tanh}$ About an Arbitrary Point} \label{sec:taylortanh}

The Taylor expansion of ${\rm tanh}$ about $0$ is well-known, specifically:
\[
 \begin{array}{l}
  {\rm tanh} \left( \xi \right) = \sum_{m=1}^\infty \frac{2^{2m} \left( 2^{2m} - 1 \right) B_{2m}}{\left( 2m \right)!} \xi^{2m-1}
 \end{array}
\]
where $B_{2n}$ is the Bernoulli number, which is valid for $|\xi| \leq 
\frac{\pi}{2}$.  However in this paper we require the Taylor expansion of 
${\rm tanh}$ about an {\em arbitrary} point $z$ - however we have been unable 
to find such an expansion in the literature.  In this section we obtain such 
an expansion using a method inspired by the approach given in the post 
\citep{TanhExpand}.

We know that the Weierstrass product expansion of ${\rm cosh}$ is:
\[
 \begin{array}{rl}
  {\rm cosh} \left( z + \xi \right) 
  &\!\!\!\!= \mathop{\prod}\limits_{n=0}^\infty \left( 1 + \frac{\left( z + \xi \right)^2}{\left( n + \frac{1}{2} \right)^2 \pi^2} \right) \\
  &\!\!\!\!= \mathop{\prod}\limits_{n=0}^\infty \left( 1 + \frac{\left( \left| z \right| + \sgn \left( z \right) \xi \right)^2}{\left( n + \frac{1}{2} \right)^2 \pi^2} \right) \\
 \end{array}
\]
for all $z,\xi \in \infset{R}$ (the ``sign absorption'' here will be 
important later in our derivation).  Hence:
\[
 \begin{array}{l}
  \log \left( {\rm cosh} \left( z + \xi \right) \right) = \mathop{\sum}\limits_{n=0}^\infty \log \left( 1 + \frac{\left( \left| z \right| + {\rm sgn} \left( z \right) \xi \right)^2}{\left( n + \frac{1}{2} \right)^2 \pi^2} \right)
 \end{array}
\]
and so, taking the derivative $\frac{\partial}{\partial \xi}$ on both sides:
\[
 \begin{array}{rl}
  {\rm sgn} \left( z \right) {\rm tanh} \left( z + \xi \right) 
  &\!\!\!\!= 
  2 \left( \left| z \right| + \sgn \left( z \right) \xi \right) \mathop{\sum}\limits_{n=0}^\infty \frac{\frac{1}{\left( n + \frac{1}{2} \right)^2 \pi^2}}{1+\frac{\left( \left| z \right| + \sgn \left( z \right) \xi \right)^2}{\left( n + \frac{1}{2} \right)^2 \pi^2}} \\
  &\!\!\!\!= 
  2 \left( \left| z \right| + \sgn \left( z \right) \xi \right) \mathop{\sum}\limits_{n=0}^\infty \frac{1}{\left( n + \frac{1}{2} \right)^2 \pi^2 + z^2 + 2 \left| z \right| \sgn \left( z \right) \xi + \xi^2} \\
  &\!\!\!\!= 
  2 \left( \left| z \right| + \sgn \left( z \right) \xi \right) \mathop{\sum}\limits_{n=0}^\infty \frac{1}{\left| z \right| + {\tt i} \left( n + \frac{1}{2} \right) \pi + \sgn \left( z \right) \xi} 
                                                                                                  \frac{1}{\left| z \right| - {\tt i} \left( n + \frac{1}{2} \right) \pi + \sgn \left( z \right) \xi} \\
  &\!\!\!\!= 
  2 \left( \left| z \right| + \sgn \left( z \right) \xi \right) \mathop{\sum}\limits_{n=0}^\infty \frac{1}{\left| z \right| + {\tt i} \left( n + \frac{1}{2} \right) \pi} 
                                                                                                  \frac{1}{\left| z \right| - {\tt i} \left( n + \frac{1}{2} \right) \pi} 
  \frac{1}{1 + \frac{1}{\left| z \right| + {\tt i} \left( n + \frac{1}{2} \right) \pi} \sgn \left( z \right) \xi}
  \frac{1}{1 + \frac{1}{\left| z \right| - {\tt i} \left( n + \frac{1}{2} \right) \pi} \sgn \left( z \right) \xi}
 \end{array}
\]
where ${\tt i} = \sqrt{-1}$.  Thus, for all $\xi^2 \leq z^2 + (\frac{\pi}{2})^2$:
\[
 \begin{array}{l}
  {\rm sgn} \left( z \right) {\rm tanh} \left( z + \xi \right) 
  =\ldots \\ \;\;\;\;\;\ldots =
  2 \left( \left| z \right| + \sgn \left( z \right) \xi \right) \mathop{\sum}\limits_{n=0}^\infty \frac{1}{\left| z \right| + {\tt i} \left( n + \frac{1}{2} \right) \pi} 
                                                             \frac{1}{\left| z \right| - {\tt i} \left( n + \frac{1}{2} \right) \pi} 
  \mathop{\sum}\limits_{p,q=0}^\infty \left( -\sgn \left( z \right) \right)^{p+q} \left( \frac{1}{\left| z \right| + {\tt i} \left( n + \frac{1}{2} \right) \pi} \right)^p 
                                                                                  \left( \frac{1}{\left| z \right| - {\tt i} \left( n + \frac{1}{2} \right) \pi} \right)^q \xi^{p+q} \\
  \;\;\;\;\;=
  2 \left( \left| z \right| + \sgn \left( z \right) \xi \right) \mathop{\sum}\limits_{n=0}^\infty \frac{1}{z^2 + \left( n + \frac{1}{2} \right)^2 \pi^2}
  \mathop{\sum}\limits_{p,q=0}^\infty \left( -\sgn \left( z \right) \right)^{p+q} \left( \frac{1}{z^2 + \left( n + \frac{1}{2} \right)^2 \pi^2} \right)^{\frac{p+q}{2}} 
  e^{{\tt i} \left( p-q \right) {\rm atan} \left( \frac{\left( n+\frac{1}{2} \right) \pi}{\left| z \right|} \right)} \xi^{p+q} \\
  \;\;\;\;\;=
  2 \left( \left| z \right| + \sgn \left( z \right) \xi \right) \mathop{\sum}\limits_{n=0}^\infty \frac{1}{z^2 + \left( n + \frac{1}{2} \right)^2 \pi^2}
  \mathop{\sum}\limits_{p,q=0}^\infty \left( -{\rm sgn} \left( z \right) \right)^{p+q} \left( \frac{1}{z^2 + \left( n + \frac{1}{2} \right)^2 \pi^2} \right)^{\frac{p+q}{2}} 
  \cos \left( {\left( p-q \right) {\rm atan} \left( \frac{\left( n+\frac{1}{2} \right) \pi}{\left| z \right|} \right)} \right) \xi^{p+q} \\
 \end{array}
\]
and so, re-arranging our indexing:
\[
 {{
 %
 }}
\]
Using the result \citep{Kna1}:
\[
 \begin{array}{l}
  \mathop{\sum}\limits_{j=0}^{N-1} \cos \left( a + jd \right) = \frac{\sin \left( \frac{Nd}{2} \right)}{\sin \left( \frac{d}{2} \right)} \cos \left( a + \frac{\left( N-1 \right)d}{2} \right)
 \end{array}
\]
and the usual results for (co)sines of sums and differences, we have that:
\[
 %
\]
(As a quick sanity check, let $z \to 0^\pm$, so ${\rm atan} (\frac{(n+\frac{1}{2})\pi}{|z|}) 
\to \frac{\pi}{2}$ and $\sgn (z) = \pm 1$:
\[
 %
\]
where $\zeta$ is the Reimann-zeta function and $B_{2n}$ the Bernoulli 
number, which is the usual Taylor expansion of ${\rm tanh}$ about $0$).

Continuing, using \citep[(1.421.2)]{Gra1}:
\[
 %
\]
and hence, using the fact that ${\rm tanh}$ is an odd function:
\begin{equation}
 \begin{array}{l}
  {\rm tanh} \left( z + \xi \right) 
  =
  {\rm tanh} \left( z \right) + \sum_{m=1}^{\infty} a^{\rm tanh}_{\left(z\right)m} \xi^m

 \end{array}
 \label{eq:tanh_taylor_general}
\end{equation}
where, using that $\cos$ and ${\rm atan}$ are odd:
\[
 \begin{array}{rl}
  a_{\left(z\right)m}^{\rm tanh}

  &\!\!\!\!= 
  2
  \left( 
  \left( -\sgn \left( z \right) \right)^{m} 
  \mathop{\sum}\limits_{n=0}^\infty 
  \left( \frac{1}{\sqrt{z^2 + \left( \left( n + \frac{1}{2} \right) \pi \right)^2}} \right)^{m+1}
  \cos \left( {\left( m+1 \right) {\rm atan} \left( \frac{\left( n+\frac{1}{2} \right) \pi}{z} \right)} \right)
  \right) \\
 \end{array}
\]

%%%%%%%%%%%%%%%%%%%%%%%%%%%%%%%%%%%%%%%%%%%%%%%%%%%%%%%%%%%%%%%%%%%%%%%%%%%%%%%
%%%%%%%%%%%%%%%%%%%%%%%%%%%%%%%%%%%%%%%%%%%%%%%%%%%%%%%%%%%%%%%%%%%%%%%%%%%%%%%

\end{document}